\documentclass{article}
\usepackage{graphicx} 
\usepackage{acro}
\usepackage[most]{tcolorbox}
\usepackage{amssymb}
\usepackage[dvipsnames]{xcolor}

\usepackage{color, colortbl}
\usepackage[first=0,last=9]{lcg}

\usepackage{algorithm}
\usepackage{algorithmic}
\usepackage{subcaption}

\usepackage{booktabs}
\usepackage{siunitx}
\usepackage{multirow}

\usepackage{geometry}
\definecolor{Gray}{gray}{0.9}

\usepackage[
backend=biber,
style=numeric,
sorting=ynt
]{biblatex}

\newcommand{\cTopology}[1]{\textcolor{NavyBlue}{\textbf{#1}}} 
\newcommand{\cRep}[1]{\textcolor{Purple}{\textbf{#1}}} 
\newcommand{\cNumerical}[1]{\textcolor{OliveGreen}{\textbf{#1}}} 

\newtcbtheorem[number within=section]{defbox}{Definition}%
{enhanced,
 colback=green!6,
 colframe=green!60!black,
 arc=3mm,
 boxrule=0.8pt,
 left=6pt, right=6pt, top=6pt, bottom=6pt,
 fonttitle=\itshape\bfseries
}{def}

\newtcbtheorem[number within=section]{propbox}{Proposition}%
{enhanced,
 colback=yellow!6,
 colframe=yellow!60!black,
 arc=3mm,
 boxrule=0.8pt,
 left=6pt, right=6pt, top=6pt, bottom=6pt,
 fonttitle=\itshape\bfseries
}{def}

\title{Language Models Refine Mechanical Linkage Designs Through Symbolic Reflection and Modular Optimisation}
\date{}

\author{
Jo\~ao Pedro Gandarela \\
Idiap Research Institute, Martigny, Switzerland \\
École Polytechnique Fédérale de Lausanne (EPFL), Switzerland \\
\texttt{firstname.lastname@idiap.ch}, \texttt{joao.gandareladesouza@epfl.ch}
\AND
Thiago Rios \and Stefan Menzel \\
Honda Research Institute Europe, Offenbach, Germany \\
\texttt{thiago.rios@honda-ri.de}, \texttt{stefan.menzel@honda-ri.de}
\AND
Andr\'e Freitas \\
Idiap Research Institute, Martigny, Switzerland \\
Department of Computer Science, University of Manchester, UK \\
National Biomarker Centre, CRUK-MI, University of Manchester, UK \\
\texttt{andre.freitas@manchester.ac.uk}
}
\author{
Jo\~ao Pedro Gandarela\textsuperscript{1,2} \quad
Thiago Rios\textsuperscript{3} \quad
Stefan Menzel\textsuperscript{3} \quad
Andr\'e Freitas\textsuperscript{1,4,5} \\
\textsuperscript{1}Idiap Research Institute, Switzerland \\
\textsuperscript{2}École Polytechnique Fédérale de Lausanne (EPFL), Switzerland \\
\textsuperscript{3}Honda Research Institute Europe, Germany \\
\textsuperscript{4}Department of Computer Science, University of Manchester, UK \\
\textsuperscript{5}National Biomarker Centre, CRUK-MI, University of Manchester, UK \\
\texttt{firstname.lastname@idiap.ch}, \texttt{joao.gandareladesouza@epfl.ch} \\
\texttt{firstname.lastname@honda-ri.de}, \texttt{andre.freitas@manchester.ac.uk} \\
}

\DeclareAcronym{mtl}{short=MTL,long=Metric Temporal Logic}
\DeclareAcronym{nesy}{short=Nesy,long=Neurosymbolic}
\DeclareAcronym{llm}{short=LLM,long=Large Language Model}
\DeclareAcronym{sr}{short=SR,long=Symbolic Regression}
\DeclareAcronym{icp}{short=ICP,long=Iterative Closest Point}
\DeclareAcronym{nlp}{short=NLP,long=Natural Language Processing}
\DeclareAcronym{lb}{short=LB,long=Lemniscates of Bernoulli ($\infty$)}
\DeclareAcronym{da}{short=$\mathbb{D}a$,long=Designer Agent}
\DeclareAcronym{ta}{short=$\mathbb{T}a$,long=Topology Agent}
\DeclareAcronym{ca}{short=$\mathbb{C}a$,long=Critique Agent}
\DeclareAcronym{llama}{short=Llama,long=Llama3.3 (70B)}
\DeclareAcronym{gemma}{short=Gemma,long=Gemma3 (12B)}
\DeclareAcronym{qwen}{short=Qwen,long=Qwen3 (4B)}
\DeclareAcronym{qwen-moe}{short=Qwen3 MoE,long=Qwen3 (30B-A3B)}
\DeclareAcronym{lrm}{short=LRM,long=Large Reasoning Model}
\DeclareAcronym{sl}{short=SL,long=Symbolic Lifting}
\DeclareAcronym{dr}{short=DR,long=Discrete Representaion}
\DeclareAcronym{cl}{short=QG,long=Qualitative Geometry}
\DeclareAcronym{pso}{short=PSO,long=Particle Swarm Optimisation}

\begin{document}

\maketitle

\begin{abstract}
Designing mechanical linkages involves combinatorial topology selection and continuous parameter fitting. We show that language models can systematically improve linkage designs through symbolic representations. Language model agents explore discrete topologies while numerical optimisers fit continuous parameters. A symbolic lifting operator translates simulator trajectories into qualitative descriptors, motion labels, temporal predicates, and structural diagnostics that models interpret across iterative design cycles. Across six engineering-relevant motion targets and three open-source models (Llama 3.3 70B, Qwen3 4B, Qwen3 MoE 30B-A3B), the modular architecture reduces geometric error by up to 68\% and improves structural validity by up to 134\% over monolithic baselines. Critically, 78.6\% of iterative refinement trajectories show measurable improvement, with the system correctly diagnosing overconstraint (56.3\%) and underconstraint (35.6\%) failure modes and proposing grounded corrections. Models across all three families acquire interpretable mechanical reasoning strategies without fine-tuning, demonstrating that principled symbolic abstraction bridges generative AI and the numerical precision required for engineering design.
\end{abstract}


\section{Introduction}
\label{sec:introduction}

Mechanical linkages are systems of rigid bodies connected by joints that convert input motion into a desired output path, and are ubiquitous in engineering systems: from automotive suspensions and surgical robots to deployable aerospace structures~\cite{uicker1964iterative,sandor1984advanced}. Designing linkages to trace prescribed trajectories is a long-standing challenge that involves two tightly coupled subproblems that must be solved jointly~\cite{grossmann2002review}. The first is a \cTopology{\textbf{combinatorial}} decision: choosing how many links and joints to use and how to connect them (the \emph{topology}). The second is a \cNumerical{\textbf{continuous}} fitting task: selecting link lengths, joint offsets, and crank angles so that the end-effector path matches a target curve. Conventional methods address both subproblems with mathematical optimisation~\cite{mabie1991mechanisms}, but the combinatorial explosion of possible topologies makes exhaustive search impractical, and gradient-based fitting can stall in local minima.

Recent progress in generative AI, particularly \ac{llm}s, has opened a new avenue for tackling the combinatorial dimension of linkage design problems~\cite{lin2025creative}. \ac{llm}s excel at analogical reasoning, pattern recognition, and rapid proposal generation across large design spaces, which are capabilities needed to navigate the discrete topology landscape. In industrial practice, such exploration has traditionally relied on catalogue lookup and the reuse of validated mechanisms~\cite{aamodt1994case, arthur2001mechanism}.

However, applying \ac{llm}s directly to linkage synthesis exposes a fundamental \cRep{\textbf{representational mismatch}}. Language models process tokenised text and lack an intrinsic notion of coordinate frames, units, or kinematic constraints. As a result, \ac{llm}-generated designs are often linguistically coherent, describing plausible-sounding mechanism configurations, but numerically unsound when the proposed link sizes are considered in numerical simulations of the mechanisms~\cite{mouselinos-etal-2024-beyond}. The kinematic equations that govern linkages (loop-closure conditions, screw coordinates) are \cNumerical{hard, verifiable constraints}~\cite{vinogradov2000fundamentals}, that defined the feasibility of the mechanism.

Therefore, \emph{how can we let language models contribute through combinatorial reasoning while ensuring engineering precision in the continuous domain?} We answer this question based on three interlocked ideas:

\begin{enumerate}
    \item \cTopology{\textbf{Factorised optimisation.}} We separate topology selection (handled by \ac{llm} agents) from dimensional fitting (handled by numerical optimisers). This separation of concerns allows each component to play to its strengths: language models handle combinatorial structure exploration, while dedicated solvers ensure numerical precision.
    \item \cRep{\textbf{Symbolic lifting as a representational bridge.}} We introduce an operator $\mathfrak{L}$ that converts the dense numerical output of a kinematic simulator, hundreds of sampled trajectory points, degrees-of-freedom reports, and error residuals, into a compact set of \emph{qualitative descriptors} that a language model can interpret: motion labels (e.g., ``straight segment'', ``sharp turn''), temporal predicates (e.g., ``the path crosses the $x$-axis between $t{=}0.1$ and $t{=}0.2$''), and structural diagnostics (e.g., ``mechanism is over-constrained''). These descriptors form a \cRep{representation bundle $\mathcal{R}$} that bridges the gap between geometric state and language-level qualitative reasoning.
    \item \cNumerical{\textbf{Closed-loop reflective refinement.}} Using $\mathcal{R}$, a multi-agent pipeline iterates: a \emph{topology agent} proposes a design; a \emph{critic agent} diagnoses errors using the symbolic feedback; a \emph{planner agent} maps the diagnosis to a targeted solution (e.g., ``overconstraint $\to$ remove a redundant link''); and a \emph{refiner agent} implements the correction. This cycle enables the language model to \textbf{reflect on symbolic feedback within a single design session}, progressively improving candidates without parameter updates.
\end{enumerate}

\begin{figure}
    \centering
    \includegraphics[width=1\linewidth]{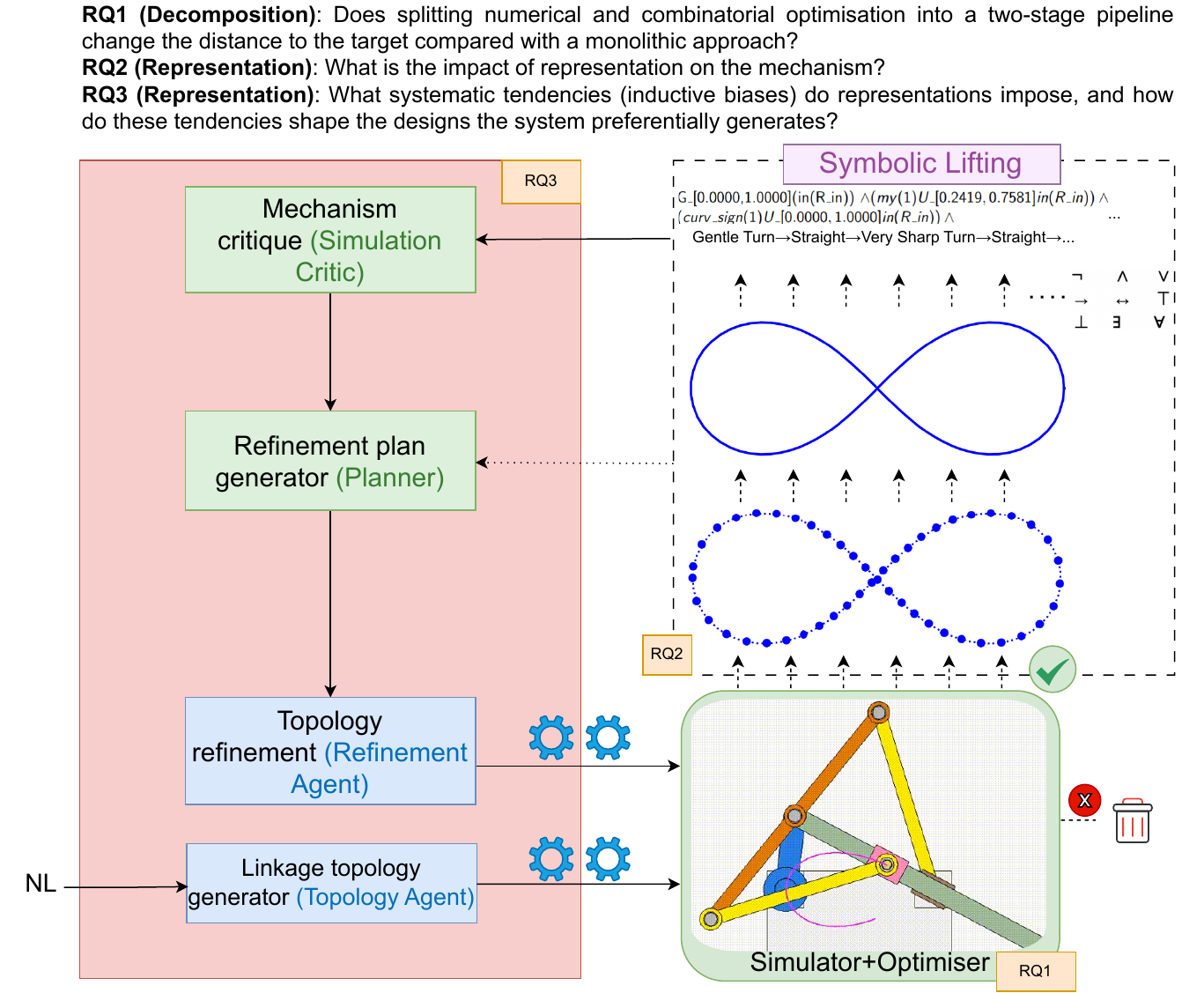}
    \caption{Overview of the symbolic lifting and closed-loop synthesis pipeline. Candidate linkage topologies proposed by language model agents are simulated and optimised to produce continuous end-effector trajectories. The symbolic lifting operator $\mathfrak{L}$ compresses these numerical outputs into a compact representation bundle $\mathcal{R}$, encoding qualitative motion patterns (straight segments, gentle and sharp turns), temporal predicates, and structural diagnostics. This symbolic representation bridges the gap between continuous kinematics and language-level reasoning, enabling iterative, verifiable refinement of mechanism designs.}
    \label{fig:symb_lift}
\end{figure}

In this paper, we show that combining these ideas into a coherent workflow, where \ac{llm}s propose structural variants, numerical optimisers supply validated parameters, and $\mathcal{R}$ mediates verifiable refinements, yields a robust synthesis method with substantially improved fidelity and reduced failure modes compared to purely generative approaches. We validate the approach across experiments on six benchmark motion targets, using three open-source language models of different architectures and scales.  Our results demonstrate that 78.6\% of iterative refinement trajectories show measurable improvement, geometric error is reduced by up to 68\%, and structural validity improves by up to 134\%, relative to monolithic baselines that do not separate topology search from continuous fitting. Crucially, inter-model variance decreases under symbolic feedback, confirming that the representational interface, not model scale, is the primary driver of design quality. These results position symbolic lifting as a principled bridge between language-model reasoning and physics-grounded engineering design, with broader implications for domains where combinatorial structure selection must be coupled with continuous parameter optimisation.

\section{Results}
\label{sec:results}

Our method (Fig.~\ref{fig:symb_lift}) decomposes linkage design into two complementary tasks: language model agents explore the discrete space of linkage topologies, while dedicated numerical optimisers fit continuous parameters (link lengths, joint positions) to minimise the geometric mismatch between the generated end-effector trajectory and a target curve. A symbolic lifting operator $\mathfrak{L}$ translates dense simulator output into compact qualitative descriptors, motion labels, temporal predicates, and structural diagnostics, that the language models use as feedback for subsequent design iterations. We evaluate performance using Chamfer distance between generated and target trajectories (lower is better) and semantic success rate (the fraction of designs that parse and simulate without error); full metric definitions and experimental protocols are given in Methods (Section~\ref{sec:method}).

We evaluated the modular framework (Fig.~\ref{fig:symb_lift}) using three open-source language models, \ac{llama}, \ac{qwen}, and \ac{qwen-moe}, on six engineering-relevant motion targets (Parabola, NACA airfoil, Line, Ellipse, Circle, and Lemniscate of Bernoulli). Full experimental details including hyperparameters, simulator configuration, and evaluation protocols are given in Methods.

\subsection{Language models reason through symbolic feedback}
\label{sec:learning_evidence}

\begin{figure}[!htp]
    \centering
    \includegraphics[width=\linewidth]{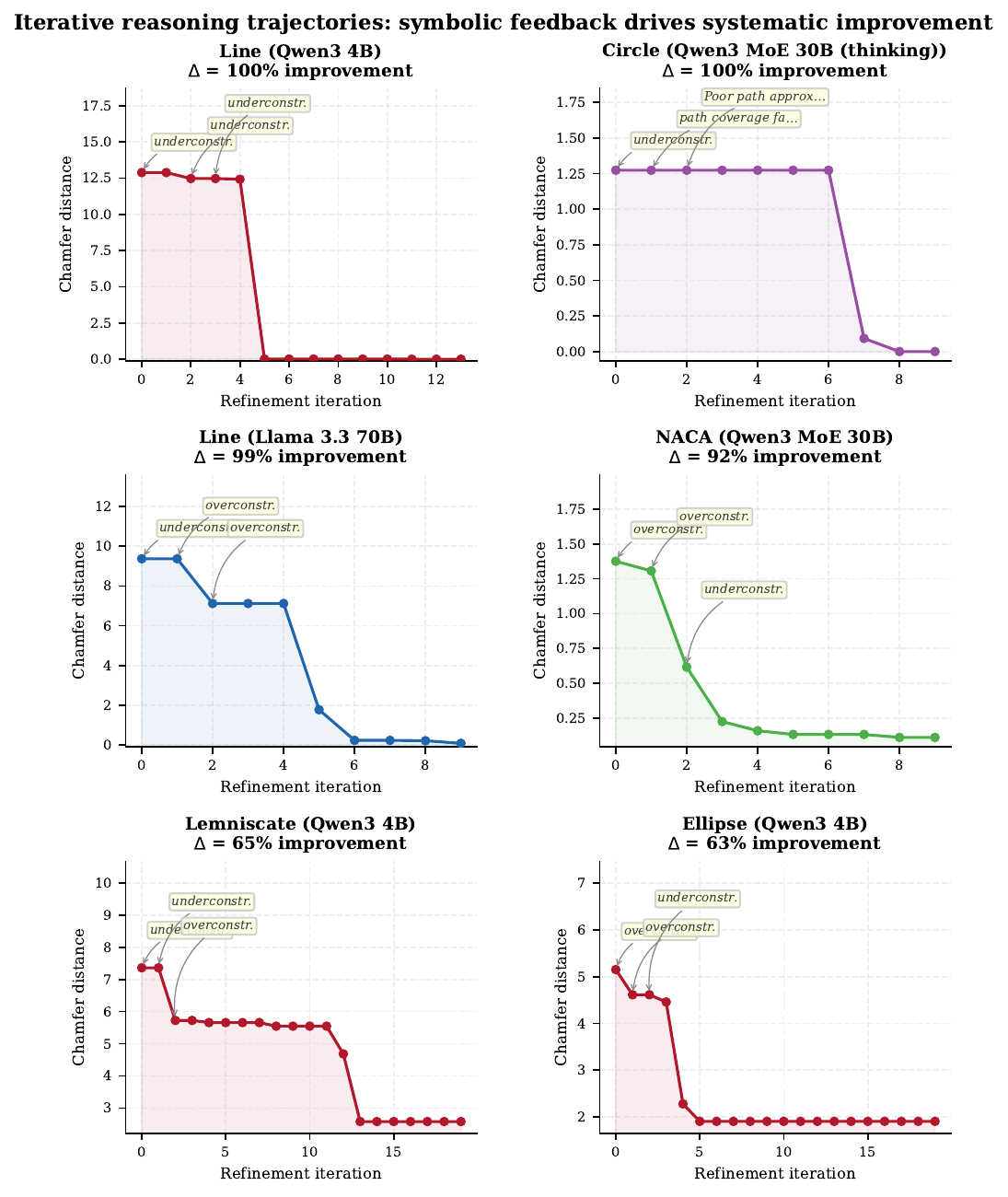}
    \caption{Representative reasoning-driven improvement trajectories showing how language model agents refine linkage designs across iterative cycles. Each panel shows one model-shape combination; the $y$-axis tracks Chamfer distance (lower is better). The consistent downward trend across heterogeneous models and shapes demonstrates that symbolic feedback enables genuine mechanistic reflection, not random search.}
    \label{fig:learning_trajectories}
\end{figure}

\begin{figure}[!htp]
    \centering
    \includegraphics[width=\linewidth]{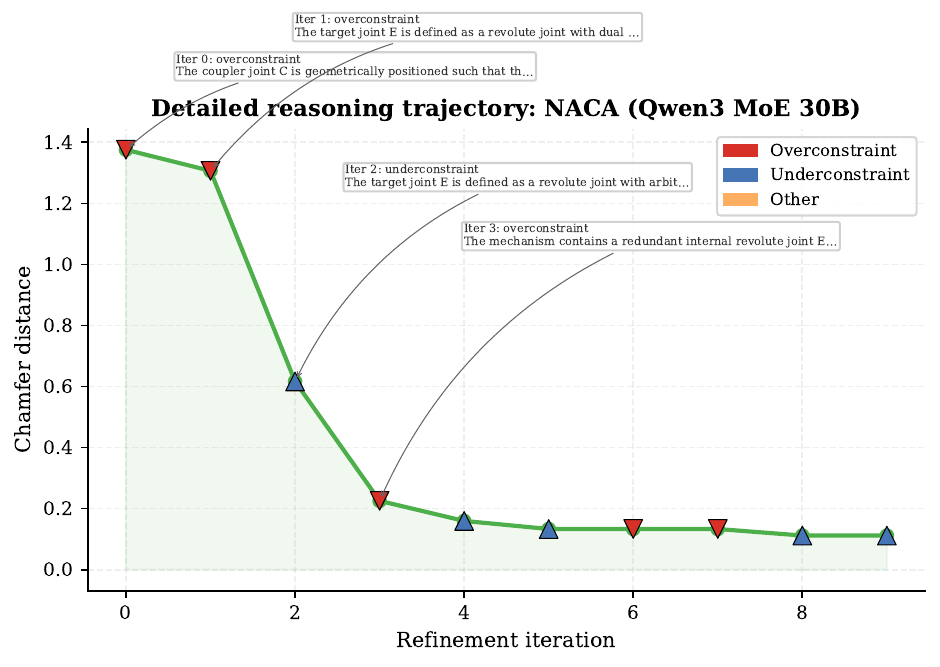}
    \caption{Detailed reasoning trajectory for a single experiment showing how the symbolic lifting operator $\mathfrak{L}$ enables structured mechanical reasoning. The annotations trace the full diagnostic cycle: raw simulation output $\to$ symbolic abstraction $\to$ failure-mode diagnosis $\to$ structural correction. This interpretable reasoning chain, produced without any fine-tuning, demonstrates that symbolic representations provide language models with the domain-specific vocabulary needed for engineering design reasoning. Unlike tool-augmented approaches that expose raw API outputs, our method presents symbolically lifted representations that encode mechanistically relevant information (DOF counts, motion labels, temporal predicates), enabling the planner to pinpoint specific structural deficiencies.}
    \label{fig:reasoning_deepdive}
\end{figure}

The central question of this work is whether language models can engage in reflective reasoning about mechanism structure and improve designs through structured symbolic feedback. Figures~\ref{fig:learning_trajectories} and~\ref{fig:reasoning_deepdive} present direct evidence that they can.

\paragraph{Iterative refinement produces systematic improvement.}
Across all experiments, \textbf{78.6\% of refinement trajectories, sequences of iterative design cycles in which the method proposes, evaluates, and refines a linkage for a single model-shape configuration, show monotonic improvement} in Chamfer distance, with an average relative improvement of 23.8\% per trajectory (measured as $(CD_{\text{initial}} - CD_{\text{final}}) / CD_{\text{initial}}$, i.e.\ the relative reduction from the first to the last iteration within each trajectory). This indicates that the critic$\to$planner$\to$refiner chain systematically translates symbolic diagnoses into effective structural corrections, rather than producing random perturbations. Figure~\ref{fig:learning_trajectories} visualises representative improvement trajectories across three model families and multiple target shapes, confirming that the reflexive feedback signal is robust across architectures.

The monotonic improvement visible in Figure~\ref{fig:learning_trajectories} confirms that symbolic feedback yields a directed refinement signal rather than stochastic perturbation. This is significant because it demonstrates that the propose $\to$ critique $\to$ correct cycle produces cumulative gains in geometric fidelity, a behaviour consistent across all three model families despite their architectural differences.

\paragraph{Failure-mode diagnosis reveals mechanistic reasoning.}
The Refinement Planning Agent consistently identifies structurally meaningful failure modes from the symbolic feedback (Table~\ref{tab:failure_modes}). The two dominant modes, \emph{overconstraint} (56.3\% of diagnosed iterations) and \emph{underconstraint} (35.6\%), account for 91.9\% of all diagnosed failures. Less frequent modes include kinematic inaccuracy (1.6\%), path deviation (0.3\%), and path misalignment (0.2\%). This distribution confirms that the symbolic lifting layer surfaces the mechanically relevant information, primarily degrees-of-freedom and connectivity diagnostics, that enables the planner to reason about \emph{structural} rather than purely \emph{numerical} causes of error.

\paragraph{Reasoning chains demonstrate mechanistically grounded corrections.}
Figure~\ref{fig:reasoning_deepdive} presents a detailed annotated example showing how the symbolic lifting operator $\mathfrak{L}$ transforms raw simulation output into interpretable qualitative descriptors that the planner uses to formulate structural corrections. The examples below show additional representative reasoning chains from the pipeline across diverse models and target shapes, illustrating how the planner diagnoses structural causes and proposes targeted mechanical edits based on the symbolic representations.

The diagnostic chain illustrated in Figure~\ref{fig:reasoning_deepdive} reveals a structured reasoning pattern: the planner does not simply relay numerical error magnitudes but instead identifies \emph{categorical} structural causes (e.g., ``overconstrained substructure at joint~C'') and prescribes mechanistically appropriate corrections (e.g., ``remove the redundant link to restore single-DOF mobility''). This interpretability distinguishes our approach from end-to-end generative methods where intermediate reasoning steps remain opaque.

\begin{table}[h]
\centering
\begin{tabular}{lrr}
  \toprule
  Failure Mode & Proportion (\%) \\
  \midrule
  Overconstraint & 56.3 \\
  Underconstraint & 35.6 \\
  Kinematic Inaccuracy & 1.6 \\
  None & 0.6 \\
  <One Of: Underconstraint, Overconstraint> & 0.4 \\
  Path Deviation & 0.3 \\
  Path Misalignment & 0.2 \\
  \bottomrule
\end{tabular}
\caption{Distribution of failure modes identified by the planner across all experiments. The planner diagnoses structural issues in candidate mechanisms and proposes targeted corrections.}
\label{tab:failure_modes}
\end{table}

\begin{tcolorbox}[enhanced, colback=blue!3, colframe=blue!50!black,
  arc=2mm, boxrule=0.6pt, title={Reasoning Chain: Lemniscate (Llama~3.3 70B), Iteration 5 ,  \textcolor{green!60!black}{$\downarrow$ improved}}]
\textbf{Failure Mode:} underconstraint \\
\textbf{Structural Cause:} The current five-bar linkage mechanism has low kinematic accuracy due to the interaction between the crank link, intermediate links, and target joint, which may not perfectly replicate the complex motion required for the Lemniscate shape. \\
\textbf{Suggested Action:} Add an additional link or loop closure to improve the kinematic accuracy. Specifically, add a new revolute joint between the target joint and one of the intermediate links to increase the degree of freedom and allow for more precise control over the motion of the target.
\\[4pt]
\textbf{Temporal Specification (excerpt):} \\
\texttt{G\_[0.00,1.00](in(R\_in)) \(\wedge\) F\_[0.015,0.035](INF | EX\_x | EX\_y) \(\wedge\) F\_[0.050,0.060](INF | EX\_x | EX\_y) \(\wedge\) F\_[0.065,0.096](INF | EX\_x | EX\_y) \(\wedge\) \ldots}
\\[4pt]
\textit{\textbf{Diagnostic link:}} The target temporal specification requires dense \(\mathsf{F}\)-bounded inflection and extremum events (\texttt{INF~|~EX\_x~|~EX\_y}) at many closely spaced intervals throughout the motion cycle, encoding the Lemniscate's multiple curvature reversals and direction changes. The current five-bar topology cannot generate this event density: with only one coupler loop, the end-effector's trajectory lacks sufficient kinematic complexity to produce the required inflection points. This mismatch between the temporal event density demanded by the specification and the structural capacity of the mechanism is what the planner identifies as \emph{underconstraint}, the topology is too simple, motivating the addition of a sixth link and new loop closure to increase the mechanism's curve-tracing expressiveness.
\end{tcolorbox}

\begin{tcolorbox}[enhanced, colback=green!3, colframe=green!50!black,
  arc=2mm, boxrule=0.6pt, title={Reasoning Chain: NACA Airfoil (Qwen3 MoE 30B), Iteration 7}]
\textbf{Failure Mode:} underconstraint \\
\textbf{Structural Cause:} The coupler link (B--C) and target joint (E) form a substructure with insufficient geometric flexibility to accurately track the desired path, particularly in high-curvature regions. The fixed distance constraints on E restrict its ability to conform to local trajectory curvature. \\
\textbf{Suggested Action:} Add a new revolute joint F on the coupler link C, positioned at an optimised location along the segment B--C, and reposition the target joint E as a dependent revolute joint connected to F instead of directly to B and C. This introduces a new degree of freedom in the coupler substructure via a ternary link with a floating joint.
\\[4pt]
\textbf{Temporal Specification (excerpt):} \\
\texttt{G\_[0.00,0.02](in(R\_in)) \(\wedge\) G\_[0.05,0.09](in(R\_in)) \(\wedge\) G\_[0.11,0.15](in(R\_in)) \(\wedge\) G\_[0.18,0.21](in(R\_in)) \(\wedge\) \ldots}
\\[4pt]
\textit{\textbf{Diagnostic link:}} The temporal specification is dominated by \(\mathsf{G}_{[a,b]}\)(\texttt{in(R\_in)}) predicates, \emph{always-in-region} constraints over many narrow, contiguous time windows. These encode that the NACA airfoil path requires the end-effector to \emph{continuously} remain within the target bounding region across closely spaced sub-intervals, demanding smooth, curvature-faithful path tracking. The current coupler substructure, with fixed distance constraints on E to both B and C, restricts E to a single coupler curve that cannot conform to the local curvature variations revealed by the dense \(\mathsf{G}\) predicates. The planner thus diagnoses \emph{effective underconstraint} (the topology is geometrically inflexible, not functionally sufficient) and prescribes a floating intermediate joint F to introduce local geometric freedom, enabling the mechanism to satisfy the \(\mathsf{G}\)-containment requirements in high-curvature regions.
\end{tcolorbox}

\begin{tcolorbox}[enhanced, colback=red!3, colframe=red!50!black,
  arc=2mm, boxrule=0.6pt, title={Reasoning Chain: Straight Line (Qwen3 MoE 30B), Iteration 8}]
\textbf{Failure Mode:} overconstraint \\
\textbf{Structural Cause:} Joint C is defined with incompatible geometric constraints: the distance between its anchor points (B and D) exceeds the sum of its specified radii, making the constraint system geometrically infeasible. The structural fault is a mismatch between joint coordinates and assigned link distances. \\
\textbf{Suggested Action:} Remove joint C and replace it with a coupler link between B and D directly, forming a standard four-bar linkage (A--B--C--D), where C is a new revolute joint defined only by the intersection of circles from B and D. Then re-establish the path-tracking chain via a new coupler link from C to E.
\\[4pt]
\textit{\textbf{Diagnostic link:}} No temporal specification is produced because the mechanism is \emph{unbuildable}, the symbolic lifting operator $\mathfrak{L}$ receives no trajectory from the simulator and therefore emits no temporal predicates. The \emph{absence} of the temporal specification is itself the diagnostic signal: a buildable mechanism would produce at minimum a \(\mathsf{G}_{[0,1]}\)(\texttt{in(R\_in)}) containment predicate. The planner interprets this null output alongside the simulator's constraint-violation report to identify the geometric infeasibility at joint C, diagnosing \emph{overconstraint} and prescribing the removal of the conflicting joint definition to restore a buildable four-bar foundation.
\end{tcolorbox}

\begin{tcolorbox}[enhanced, colback=orange!3, colframe=orange!50!black,
  arc=2mm, boxrule=0.6pt, title={Reasoning Chain: Ellipse (Qwen3 MoE 30B-Thinking), Iteration 8}]
\textbf{Failure Mode:} overconstraint \\
\textbf{Structural Cause:} The mechanism contains a redundant triangle (crank--target--D) that creates a 4-joint loop with inconsistent mobility. The current design has 6 joints for 6 links, but 7 joints are required for 1-DOF ($F = 3(6-1) - 2\times 6 = 3$ DOF). The direct crank--D link creates an unnecessary constraint that makes the system immobile. \\
\textbf{Suggested Action:} Replace the D joint definition to remove the direct crank--D link. Change D to connect only the target to B (not crank to D), and create a new E--D joint, restructuring the topology into a single well-constrained kinematic chain.
\\[4pt]
\textbf{Temporal Specification (excerpt):} \\
\texttt{G\_[0.00,1.00](in(R\_in)) \(\wedge\) F\_[0.00,1.00](INF | EX\_x | EX\_y) \(\wedge\) (\(\neg\)cross(L\_0) U\_[0.00,1.00] in(R\_in)) \(\wedge\) \ldots}
\\[4pt]
\textit{\textbf{Diagnostic link:}} This specification combines three distinct temporal operator types, each encoding a different kinematic requirement for the ellipse. The \(\mathsf{G}_{[0,1]}\)(\texttt{in(R\_in)}) demands global containment; the \(\mathsf{F}_{[0,1]}\)(\texttt{INF~|~EX\_x~|~EX\_y}) requires that inflection or extremum events eventually occur; and the \(\mathsf{U}\) (until) predicate \(\neg\)\texttt{cross(L\_0)} $\mathsf{U}_{[0,1]}$ \texttt{in(R\_in)} imposes a \emph{temporal ordering} constraint, the trajectory must not cross the reference boundary until it re-enters the target region. The redundant crank--target--D triangle identified by the planner yields $F = 3$ DOF instead of the required $F = 1$, creating structural immobility that prevents the mechanism from satisfying the \(\mathsf{U}\)-ordering: without smooth, single-DOF motion, the trajectory cannot maintain the sequenced enter--stay--exit pattern the \(\mathsf{U}\) predicate encodes. The planner's topological edit removes the redundant loop to restore 1-DOF mobility, aligning the mechanism's kinematic capacity with the temporal ordering demanded by the specification.
\end{tcolorbox}

These chains reveal that the language model agents are not merely pattern-matching: they identify specific structural deficiencies (e.g., ``rigid chain of revolute joints creates over-constrained substructure'', ``insufficient geometric flexibility to track high-curvature regions'') and propose mechanistically appropriate corrections (e.g., ``insert a parallel link to create a secondary loop'', ``reposition the coupler joint to satisfy distance constraints''). Crucially, this reasoning behaviour is consistent across model families, from the 4B-parameter Qwen3 to the 70B-parameter Llama~3.3, indicating that the \emph{symbolic interface} enables mechanical reasoning and not model scale alone.

\subsection{Modular design achieves superior geometric accuracy}
\label{sec:geometric_accuracy}

The modular approach, separating topology search from continuous fitting, consistently reduces the mismatch to the targeted trajectories across all model and shape combinations (Table~\ref{tab:combined_all}). Across the six shapes, the modular method reduces the Chamfer distance by approximately 44\% for \ac{llama} and 48\% for \ac{qwen} under Grid search. Quantitative outcomes per-shape show substantial Chamfer distance reductions: approximately 49.5\% for Circle, 67.7\% for Line, and 36.2\% for Parabola.

To benchmark against a classical search baseline, we compare our symbolic pipeline with Enum+GA while preserving each method's natural grouping: Enum+GA is reported by budget and bar family, and the symbolic method is reported by model family. Table~\ref{tab:combined_all} reports per-shape Chamfer values and aggregated normalized indices, while Table~\ref{tab:combined_all} provides the standalone normalized x/Base view.

Panel C of Table~\ref{tab:combined_all} compares the symbolic method against Enum+GA across matched evaluation budgets. At the same budget as our method (3$\times$20 and 6$\times$20 population$\times$generations), the symbolic method consistently outperforms Enum+GA by 30-57\%, a gap we attribute to the reasoning capabilities of the underlying language model: rather than blindly exploring the search space, the \ac{llm} leverages structural priors to propose topologically meaningful configurations from the outset.

{\small
\begin{table}[!htp]
\centering
\begin{tabular}{ll l r r r}
\toprule

\multicolumn{6}{l}{\textit{A) Effect of Representation}} \\
\cmidrule(r){1-6}
\textbf{Model} & \textbf{Type} & & \textbf{x/Base} & \textbf{$\Delta$ (On$-$Off)} & \\
\midrule
\multirow{2}{*}{\ac{llama}}    & Off & & 1.000                              & \multirow{2}{*}{\textcolor{teal}{$\uparrow$13.8\%}} & \\
                               & On  & & \textbf{\textcolor{teal}{$\uparrow$0.862}} & & \\
\midrule
\multirow{2}{*}{\ac{qwen}}     & Off & & \textbf{1.000}                     & \multirow{2}{*}{\textcolor{red}{$\downarrow$0.2\%}} & \\
                               & On  & & \textcolor{red}{$\downarrow$1.002} & & \\
\midrule
\multirow{2}{*}{\ac{qwen-moe}} & Off & & 1.000                              & \multirow{2}{*}{\textcolor{teal}{$\uparrow$45.6\%}} & \\
                               & On  & & \textbf{\textcolor{teal}{$\uparrow$0.544}} & & \\

\specialrule{1.5pt}{3pt}{3pt}

\multicolumn{6}{l}{\textit{B) Optimizer Comparison}} \\
\cmidrule(r){1-6}
\textbf{Model} & \textbf{Type} & & \textbf{x/Base} & \textbf{$\Delta$ (x$-$Base)} & \\
\midrule
                  & Base & & 1.000 & & \\
\rowcolor{Gray} \textbf{\ac{llama}} & Grid & & \textbf{\textcolor{teal}{$\uparrow$0.588}} & \textcolor{teal}{$\uparrow$41.2\%} & \\
\rowcolor{Gray}   & PSO  & & \textcolor{red}{$\downarrow$1.185}              & \textcolor{red}{$\downarrow$18.5\%} & \\
\midrule
                  & Base & & 1.000 & & \\
\rowcolor{Gray} \textbf{\ac{qwen}} & Grid & & \textbf{\textcolor{teal}{$\uparrow$0.544}} & \textcolor{teal}{$\uparrow$45.6\%} & \\
\rowcolor{Gray}   & PSO  & & \textcolor{teal}{$\uparrow$0.810}               & \textcolor{teal}{$\uparrow$19.0\%} & \\

\specialrule{1.5pt}{3pt}{3pt}

\multicolumn{6}{l}{\textit{C) Enum+GA vs.\ Symbolic by Budget and Bars}} \\
\cmidrule(r){1-6}
\textbf{Pop.\ \texttimes\ Gen.} & \textbf{Bars} & \textbf{Model} & \textbf{x/Base} & \textbf{$\Delta$ (x$-$Base)} & \textbf{Imp.\%} \\
\midrule
3\texttimes20  & 4 & \ac{llama}    & \textbf{$0.437 \pm 0.142$} & \textcolor{teal}{${\downarrow}0.563$}  & \textcolor{teal}{${\uparrow}56.3\%$}  \\
3\texttimes20  & 4 & \ac{qwen}     & \textbf{$0.434 \pm 0.132$} & \textcolor{teal}{${\downarrow}0.566$}  & \textcolor{teal}{${\uparrow}56.6\%$}  \\
3\texttimes20  & 4 & \ac{qwen-moe} & \textbf{$0.508 \pm 0.130$} & \textcolor{teal}{${\downarrow}0.492$}  & \textcolor{teal}{${\uparrow}49.2\%$}  \\
3\texttimes20  & 6 & \ac{llama}    & \textbf{$0.516 \pm 0.157$} & \textcolor{teal}{${\downarrow}0.484$}  & \textcolor{teal}{${\uparrow}48.4\%$}  \\
3\texttimes20  & 6 & \ac{qwen}     & \textbf{$0.512 \pm 0.147$} & \textcolor{teal}{${\downarrow}0.488$}  & \textcolor{teal}{${\uparrow}48.8\%$}  \\
3\texttimes20  & 6 & \ac{qwen-moe} & \textbf{$0.604 \pm 0.152$} & \textcolor{teal}{${\downarrow}0.396$}  & \textcolor{teal}{${\uparrow}39.6\%$}  \\
6\texttimes20  & 4 & \ac{llama}    & \textbf{$0.543 \pm 0.148$} & \textcolor{teal}{${\downarrow}0.457$}  & \textcolor{teal}{${\uparrow}45.7\%$}  \\
6\texttimes20  & 4 & \ac{qwen}     & \textbf{$0.543 \pm 0.137$} & \textcolor{teal}{${\downarrow}0.457$}  & \textcolor{teal}{${\uparrow}45.7\%$}  \\
6\texttimes20  & 4 & \ac{qwen-moe} & \textbf{$0.638 \pm 0.133$} & \textcolor{teal}{${\downarrow}0.362$}  & \textcolor{teal}{${\uparrow}36.2\%$}  \\
6\texttimes20  & 6 & \ac{llama}    & \textbf{$0.599 \pm 0.179$} & \textcolor{teal}{${\downarrow}0.401$}  & \textcolor{teal}{${\uparrow}40.1\%$}  \\
6\texttimes20  & 6 & \ac{qwen}     & \textbf{$0.595 \pm 0.170$} & \textcolor{teal}{${\downarrow}0.405$}  & \textcolor{teal}{${\uparrow}40.5\%$}  \\
6\texttimes20  & 6 & \ac{qwen-moe} & \textbf{$0.703 \pm 0.179$} & \textcolor{teal}{${\downarrow}0.297$}  & \textcolor{teal}{${\uparrow}29.7\%$}  \\
\bottomrule
\end{tabular}
\caption{Normalized Index (x/Base) results across three experimental axes.
  \textbf{Panel A} shows the effect of structured representation (On vs.\ Off)
  on mean Chamfer distance; values below 1.0 indicate improvement over the
  unstructured baseline.
  \textbf{Panel B} compares optimizer strategies (Grid search vs.\ PSO) against
  the monolithic baseline for \ac{llama} and \ac{qwen}.
  \textbf{Panel C} compares Enum+GA (baseline $= 1.0$) against the symbolic
  method grouped by population$\times$generation budget and bar count; values
  are mean $\pm$ standard error aggregated across shared target shapes.
  Bold entries denote the best result per model and panel.
  Panels A and B support \textbf{RQ1} and \textbf{RQ2}; Panel C supports
  \textbf{RQ3}.}
\label{tab:combined_all}
\end{table}}

\begin{figure}[!htp]
    \begin{subfigure}[t]{0.50\textwidth}
        \includegraphics[width=\linewidth]{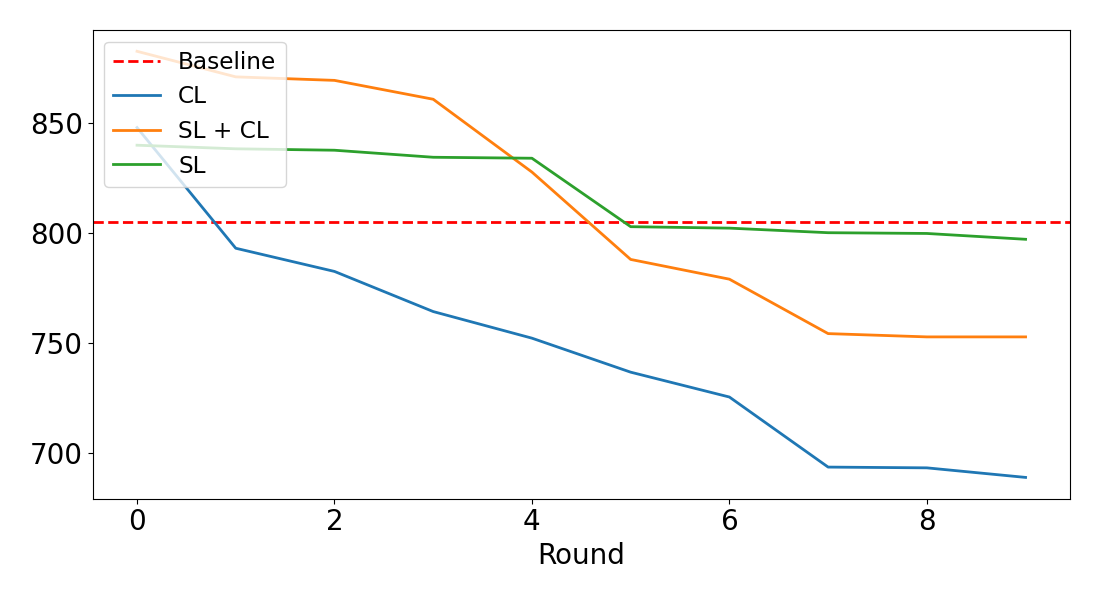}
        \caption{Parabola}
        \label{fig:opt_1}
    \end{subfigure}
    \begin{subfigure}[t]{0.50\textwidth}
        \includegraphics[width=\linewidth]{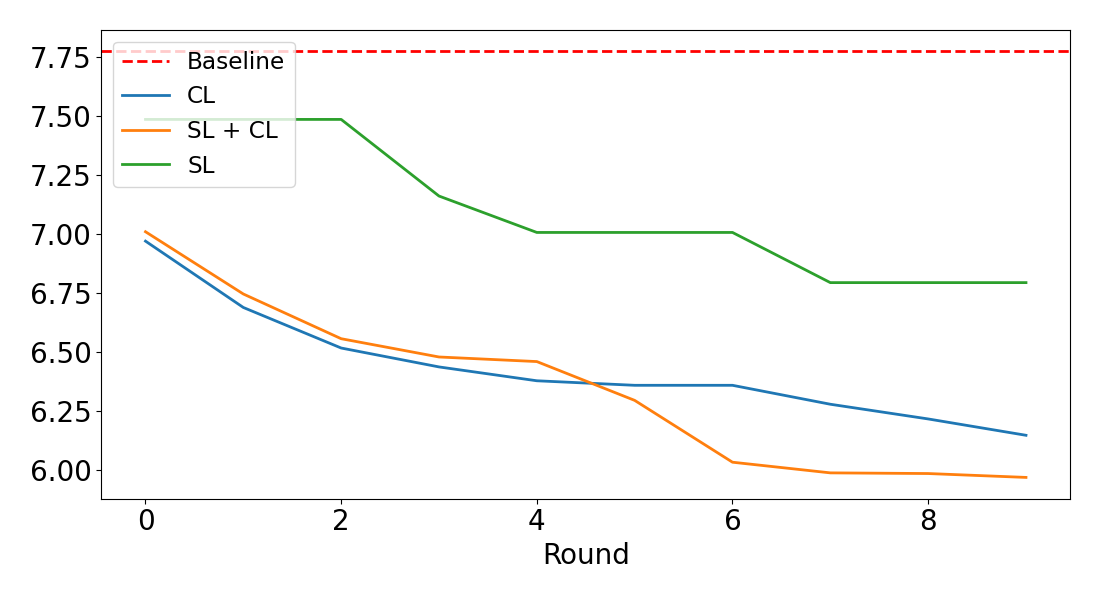}
        \caption{Lemniscate of Bernoulli}
        \label{fig:opt_2}
    \end{subfigure}
    \begin{subfigure}[t]{0.50\textwidth}
        \includegraphics[width=\linewidth]{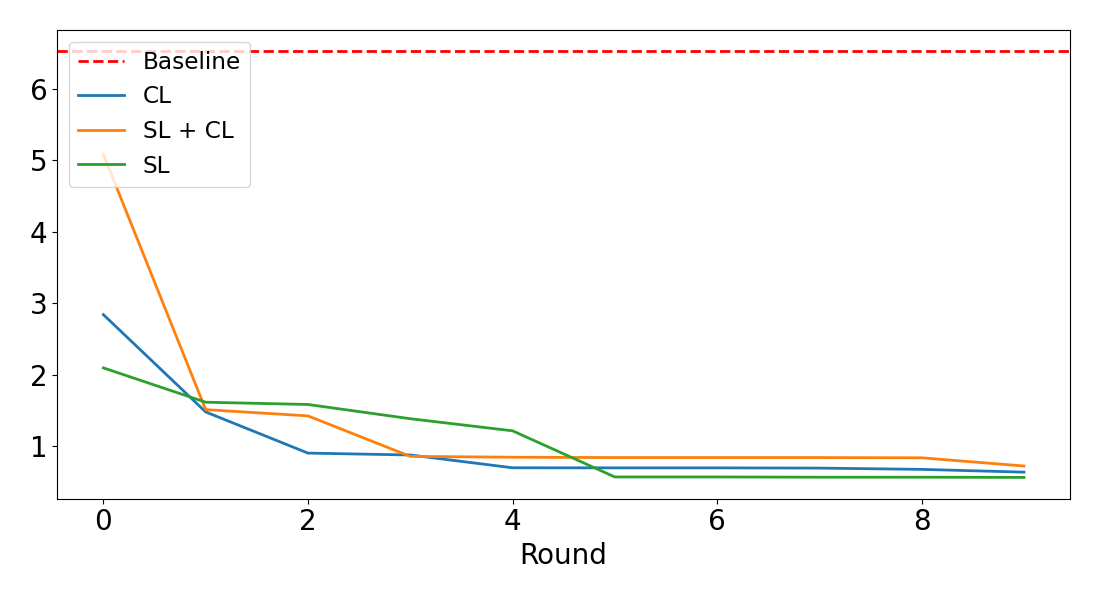}
        \label{fig:opt_3}
        \caption{Line}
    \end{subfigure}
    \begin{subfigure}[t]{0.50\textwidth}
        \includegraphics[width=\linewidth]{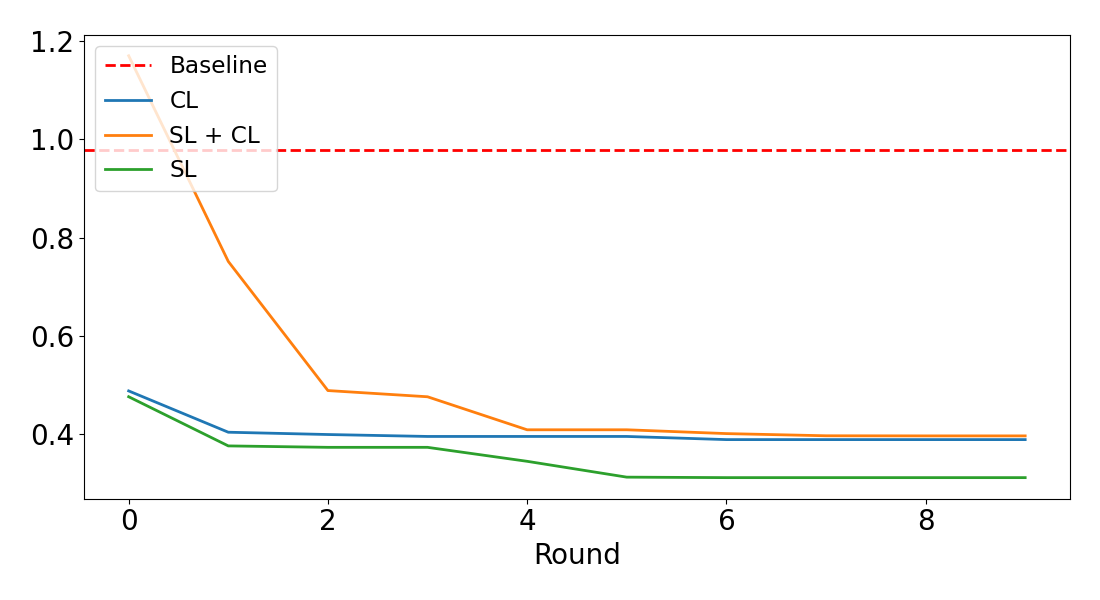}
        \label{fig:opt_4}
        \caption{NACA airfoil}
    \end{subfigure}
    \caption{Mean optimisation trajectories for representative tasks using Grid search with \ac{llama}. Modular and representation-aware configurations (coloured curves) consistently converge faster and reach lower final objective values than the monolithic baseline (black dashed), across all geometries. The shared convergence pattern, independent of task complexity, indicates that gains arise from the structured separation of combinatorial and continuous concerns~\cite{grossmann2002review} rather than being task-specific. In all four panels, modular configurations reach lower plateaux within fewer iterations, confirming that symbolic lifting provides a more informative starting point for the numerical optimiser.}
    \label{fig:opts}
\end{figure}

Figure~\ref{fig:opts} shows mean optimisation trajectories for representative tasks. Modular and representation-aware configurations consistently outperform the monolithic baseline, converging faster and reaching lower final objective values across all tested geometries.

The convergence profiles in Figure~\ref{fig:opts} have two notable characteristics. First, the modular configurations (coloured curves) reach lower final objective values than the monolithic baseline across all four target shapes, confirming the quantitative findings in Table~\ref{tab:combined_all}. Second, representation-enriched configurations converge earlier, suggesting that the symbolic descriptors provided by $\mathfrak{L}$ guide the topology agent toward structurally sound initial proposals that require less continuous refinement. This is consistent with the general principle that factorising mixed discrete-continuous problems into specialised subproblems improves both solution quality and convergence speed~\cite{mabie1991mechanisms}.

\subsection{Improved structural validity and semantic correctness}
\label{sec:semantic_validity}

The geometric improvements translate directly into better structural validity. Semantic correctness, measured as the fraction of generated mechanisms that both parse and simulate without error, increases substantially (Table~\ref{tab:semantic}). On average, semantic scores improve by approximately 56.5\% relative to the baselines, with extreme cases showing 134\% improvement (e.g., \ac{qwen} on Circle: $0.057 \rightarrow 0.835$), converting near-failure cases into semantically correct reconstructions.

{\small
\begin{table}[!htp]
    \centering

\begin{tabular}{lll rr}
\toprule
\textbf{Model} & \textbf{Type} & \textbf{Value} & \textbf{$\Delta$ (x-Base)} & \textbf{Imp. \%} \\
\midrule
 & Base & 0.624 &  &   \\
\rowcolor{Gray} \textbf{\ac{llama}}& Grid & \textbf{0.967} & \textcolor{teal}{$\uparrow$0.343} & \textcolor{teal}{$\uparrow$55.0\%} \\
\rowcolor{Gray} & PSO & {0.924} & \textcolor{teal}{$\uparrow$0.300} & \textcolor{teal}{$\uparrow$48.1\%} \\
\midrule
 & Base & 0.427 &  & \\
\rowcolor{Gray} \textbf{\ac{qwen}}& Grid & {0.824} & \textcolor{teal}{$\uparrow$0.397} & \textcolor{teal}{$\uparrow$93.0\%} \\
\rowcolor{Gray} & PSO & \textbf{1.000} & \textcolor{teal}{$\uparrow$0.573} & \textcolor{teal}{$\uparrow$134.1\%} \\
\bottomrule
\end{tabular}

\caption{Semantic correctness (fraction of semantically valid outputs) by model and optimiser; bold entries mark best result. The modular procedure substantially increases semantic validity across models (e.g. \ac{qwen} $100\%$ vs. baseline 0.42). These results show the modular method improves not only numerical objectives but also semantic success, and, together with representation analyses in Table~\ref{tab:combined_all}, illustrate how representation choices influence output correctness (supports \textbf{RQ1} and \textbf{RQ2}).}

\label{tab:semantic}
\end{table}}

\subsection{Symbolic lifting reduces inter-model variance}
\label{sec:inter_model}

\begin{figure}[!htp]
    \centering
    \includegraphics[width=0.6\linewidth]{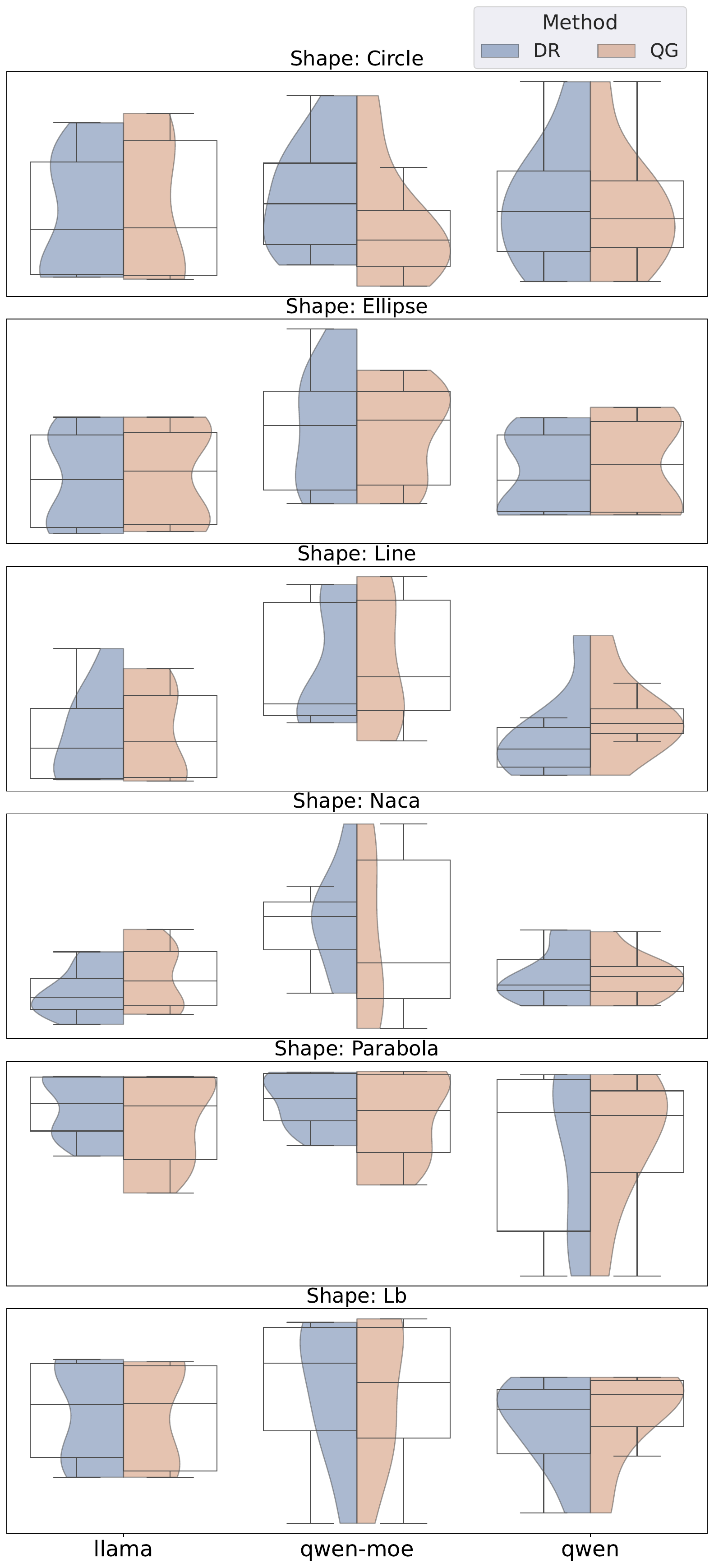}
    \caption{Distribution of best Chamfer distance by model family and representation method. Each raincloud combines a half-violin density, boxplot, and whiskers. Rows correspond to the six target shapes; columns compare discrete-segmental (blue) and compositional (orange) representations across \ac{llama}, \ac{qwen-moe}, and \ac{qwen}. When symbolic representations are active, inter-model variance decreases substantially, indicating that the symbolic interface, rather than model architecture, is the primary driver of design quality. This variance-reduction effect is analogous to how structured prompting stabilises outputs across model scales.}
    \label{fig:raincloud}
\end{figure}

A practical concern for deploying AI-assisted design is robustness across different model architectures. We find that symbolic lifting substantially reduces performance variance across the three heterogeneous model families (Table~\ref{tab:combined_all}, Fig.~\ref{fig:raincloud}).

When symbolic representations are provided, distinct architectures converge to comparable performance levels. As shown in Fig.~\ref{fig:raincloud}, both the discrete-segmental and compositional representations display similar distributions across models, confirming that abstracting low-level differences via a common symbolic interface stabilises downstream results. This demonstrates that the \textbf{symbolic interface, not model scale, drives consistent design quality}, an important finding for practical deployment where model choice may be constrained by cost or availability.

Figure~\ref{fig:raincloud} reveals that the distribution of Chamfer distances narrows considerably when symbolic representations are provided: the spread between the best and worst model families shrinks, and the median values converge. This confirms the findings in Table~\ref{tab:combined_all} and supports the interpretation that the compact, semantics-preserving bundle $\mathcal{R}$ acts as a normalising interface. Practically, this means that practitioners can select models based on cost or latency constraints without substantial loss in design quality, a key consideration for deployable AI-assisted engineering workflows.

\subsection{Planner guides constraint satisfaction}
\label{sec:planner_results}

Language models often struggle with strict numerical constraints during generation. The Refinement Planning Agent mitigates this by interpreting diagnostic feedback and mapping it to canonical corrective actions (e.g., overconstraint $\mapsto$ remove a redundant link; underconstraint $\mapsto$ add a loop). As summarised in Table~\ref{tab:plan_rep_combined}, enabling the Planner and enabling \ac{sl} both consistently move Goal links closer to the ideal target (0) across all three model families. The improvements are systematic, with the strongest gains typically observed for \ac{qwen-moe}, supporting the conclusion that symbolic structural feedback and explicit planning both help the method satisfy discrete design constraints; see Table~\ref{tab:plan_rep_combined} for the exact values.

{\small
\begin{table}[!htp]
\centering
\begin{tabular}{ll rr rr}
\toprule
 & & \multicolumn{2}{c}{\textbf{Planner}} & \multicolumn{2}{c}{\textbf{\ac{sl}}} \\
\cmidrule(lr){3-4} \cmidrule(lr){5-6}
\textbf{Model} & \textbf{Config} & \textbf{Goal links} & \textbf{Imp.\,\%} & \textbf{Goal links} & \textbf{Imp.\,\%} \\
\midrule
\multirow{3}{*}{\ac{llama}}
  & Off & 2.605 & \multirow{2}{*}{\textcolor{teal}{$\uparrow$5.5\%}} & 2.706 & \multirow{2}{*}{\textcolor{teal}{$\uparrow$12.1\%}} \\
  & On  & \textbf{2.461} & & \textbf{2.380} & \\
  & $\Delta$ & \textcolor{teal}{$\uparrow$0.144} & & \textcolor{teal}{$\uparrow$0.326} & \\
\midrule
\multirow{3}{*}{\ac{qwen}}
  & Off & 2.130 & \multirow{2}{*}{\textcolor{teal}{$\uparrow$6.2\%}} & 2.124 & \multirow{2}{*}{\textcolor{teal}{$\uparrow$8.0\%}} \\
  & On  & \textbf{1.997} & & \textbf{1.955} & \\
  & $\Delta$ & \textcolor{teal}{$\uparrow$0.133} & & \textcolor{teal}{$\uparrow$0.170} & \\
\midrule
\multirow{3}{*}{\ac{qwen-moe}}
  & Off & 1.621 & \multirow{2}{*}{\textcolor{teal}{$\uparrow$13.2\%}} & 1.570 & \multirow{2}{*}{\textcolor{teal}{$\uparrow$13.9\%}} \\
  & On  & \textbf{1.407} & & \textbf{1.352} & \\
  & $\Delta$ & \textcolor{teal}{$\uparrow$0.214} & & \textcolor{teal}{$\uparrow$0.218} & \\
\bottomrule
\end{tabular}
\caption{Effect of the Refinement Planning Agent (Planner) and structural representation (\ac{sl}) on Goal links. Mean Goal-link values (closer to 0 is better) are shown for Off and On configurations; bold entries mark the best result per condition.}
\label{tab:plan_rep_combined}
\end{table}}

\subsection{Generated mechanisms approximate engineering-relevant shapes}
\label{sec:qualitative}

\begin{figure}[!htp]
    \centering
    \begin{subfigure}[t]{0.32\textwidth}
        \includegraphics[width=\linewidth]{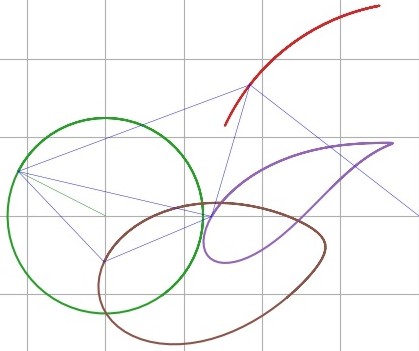}
        \caption{NACA airfoil approximation produced by a 6-bar linkage (\ac{qwen-moe}, PSO).}
        \label{fig:qual_naca}
    \end{subfigure}
    \hfill
    \begin{subfigure}[t]{0.33\textwidth}
        \includegraphics[width=\linewidth]{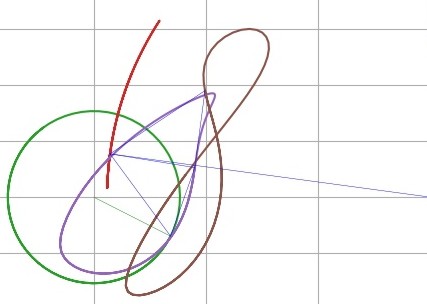}
        \caption{Lemniscate of Bernoulli produced by an 8-bar linkage (\ac{qwen-moe}, PSO).}
        \label{fig:qual_lemniscate}
    \end{subfigure}
    \hfill
    \begin{subfigure}[t]{0.33\textwidth}
        \includegraphics[width=\linewidth]{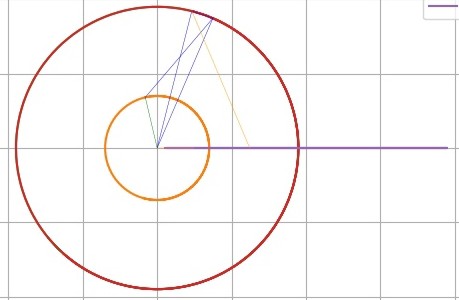}
        \caption{Straight-line mechanism produced by a 4-bar linkage (\ac{qwen}, Grid).}
        \label{fig:qual_line}
    \end{subfigure}
    \caption{Representative mechanism outputs for three target shapes. In each plot: \textbf{thin lines} are the rigid bars (links) constituting the mechanism skeleton; \textbf{thick lines} trace the complete joint trajectories over one full crank rotation; \textbf{different colours} identify different joints, with the end-effector trace shown as the most prominent coloured curve. The AI-designed linkages successfully approximate engineering-relevant curves, including a NACA 4-digit airfoil profile~(a), a lemniscate figure-eight~(b), and a straight-line motion~(c). The NACA result is particularly noteworthy given the parametric complexity of airfoil profiles. Additional qualitative results with full visual detail are shown in Appendix Figs.~\ref{figs:naca_lemniscate}--\ref{figs:ellipse}.}
    \label{fig:qualitative_main}
\end{figure}

Beyond quantitative metrics, visual inspection of the generated mechanisms confirms that our method captures both global shape structure and local geometric features. Figure~\ref{fig:qualitative_main} presents representative outputs for four target shapes, demonstrating the method's ability to produce physically realisable linkage mechanisms that approximate complex engineering curves.

The NACA airfoil results (Fig.~\ref{fig:qual_naca}) are particularly illustrative: despite the parametric complexity of airfoil profiles, characterised by leading-edge sharpness and asymmetric curvature, the generated mechanisms successfully approximate the key visual signatures used in aerospace applications. This demonstrates that the framework can handle shape targets directly relevant to engineering practice.

\section{Discussion}
\label{sec:discussion}

\subsection{Reflective reasoning and representation in AI-driven mechanism design}

The results of our experiments provide direct empirical evidence that language models engage in structured mechanical reasoning when given appropriate symbolic representations. The reasoning chains (Section~\ref{sec:learning_evidence}, Figs.~\ref{fig:learning_trajectories}--\ref{fig:reasoning_deepdive}) reveal that agents diagnose specific structural deficiencies, such as overconstrained substructures, redundant ground connections, or insufficient geometric flexibility in coupler links, and propose targeted corrections drawn from established mechanism design principles. This reflective reasoning behaviour, achieved without any parameter updates or fine-tuning, has significant implications for AI-assisted engineering design and justifies the 78.6\% monotonic improvement rate across refinement trajectories observed in the experiments.

Viewed through the lens of structured representation, the symbolic lifting operator $\mathfrak{L}$ functions as a structured encoding layer: it projects high-dimensional simulator state into a compact, interpretable feature space of qualitative descriptors (motion labels, temporal predicates, and structural diagnostics). Unlike latent embeddings produced by neural approaches~\cite{lee2024deep,kim2025data}, these representations are \emph{fully interpretable}, engineers can inspect the symbolic descriptors to understand \emph{why} a particular diagnosis was reached and \emph{what} structural modification is proposed. This transparency positions the framework as a step toward \textbf{explainable AI for engineering design}: the symbolic representations serve simultaneously as the model's ``working memory'' for reasoning and as an audit trail that domain experts can verify.

The consistency of mechanistic reasoning across three architecturally heterogeneous model families (Figs.~\ref{fig:learning_trajectories},~\ref{fig:raincloud}) supports the interpretation that it is the \emph{representation}, not the model, that enables domain-appropriate reasoning. This finding is in line with recent work on structured prompting~\cite{yao2022react} and tool-augmented language models~\cite{schick2023toolformer}, but extends it to a domain where correctness is physically verifiable: every proposed design is validated by kinematic simulation before acceptance.

Prior work on \ac{llm}-based agents for structured decision-making~\cite{yao2022react,nakano2021webgpt} has shown that interleaving reasoning with actions improves task performance. Our multi-agent decomposition extends this paradigm to engineering: by binding high-level reasoning to mechanistic interfaces and compact symbolic artefacts, we ensure that proposals remain verifiable and physically actionable. This differs from tool-augmented \ac{llm} approaches~\cite{schick2023toolformer,shen2023hugginggpt} in that our agents receive not raw tool outputs but symbolically lifted representations specifically designed for mechanical reasoning.

\subsection{Model-specific behaviours reveal architectural insights}

The three model families exhibit distinct design behaviours that illuminate how architectural choices interact with the symbolic interface:

\textbf{\ac{llama}} acts as a robust generalist, maintaining high semantic correctness ($\approx 0.96\!-\!0.98$) and stable performance across all tasks. However, its optimisation profile is flat: while it rarely fails, it also rarely reaches the extreme minima achieved by \ac{qwen} in favourable conditions.

\textbf{\ac{qwen}} exhibits a high-variance, ``deep-valley'' profile. In the unconstrained baseline it struggles significantly with complex constraints, but once the search space is structured by the modular method, it frequently surpasses \ac{llama} in raw minimisation depth. For example, \ac{qwen} achieves a Chamfer distance of 513.72 on Parabola, substantially lower than \ac{llama}'s 692.30.

\textbf{\ac{qwen-moe}} benefits most strongly from symbolic representations. We hypothesise that symbolic lifting functions as a variance-reduction filter for the mixture-of-experts routing: by normalising diverse agent outputs into a consistent symbolic structure, it stabilises the input distribution seen by the gating network, increasing the likelihood that geometric sub-problems are directed to relevant expert modules. This hypothesis is consistent with the empirical observation that \ac{qwen-moe} achieves near-perfect constraint adherence ($-0.0021$) when guided by the planner.

\subsection{Engineering impact and broader applicability}

While we demonstrated the framework on planar linkage synthesis, the underlying architecture, factorised search, symbolic lifting, and closed-loop refinement, is domain-agnostic and directly applicable to other mechanical design problems where combinatorial and continuous decisions must be jointly optimised:

\begin{itemize}
    \item \textbf{Topology optimisation}: the factorisation principle directly maps to separating structural topology decisions from material distribution and sizing, a long-standing challenge in structural engineering~\cite{bendsoe2013topology}.
    \item \textbf{Compliant mechanism design}: these mechanisms blend rigid-body kinematics with elastic deformation, requiring the same interplay of discrete topology and continuous compliance that our framework addresses.
    \item \textbf{Gear train and cam design}: the combinatorial selection of gear ratios, tooth profiles, or cam profiles paired with continuous dimensional optimisation follows the same factorised structure.
    \item \textbf{Robotic mechanism synthesis}: the design of grippers, prosthetic hands, and deployable structures involves linkage-like topology choices that could benefit from \ac{llm}-guided exploration.
\end{itemize}

These application domains connect directly to the United Nations Sustainable Development Goals that motivate this work. By enabling more efficient exploration of design spaces, AI-assisted mechanism synthesis supports \textbf{SDG~9} (Industry, Innovation, and Infrastructure) through accelerated prototyping and reduced engineering iteration cycles. The framework's ability to evaluate and reject infeasible designs early, before physical prototyping, contributes to \textbf{SDG~12} (Responsible Consumption and Production) by reducing material waste in the design process.

\subsection{Relationship to prior work}

Our approach bridges two gaps in the mechanism synthesis landscape. On one hand, classical mathematical optimisation~\cite{uicker1964iterative,sandor1984advanced,mabie1991mechanisms} and recent ML-driven approaches~\cite{lee2024deep,lee2023deep,soriano2024mathematical,kim2025data} perform effective continuous fitting but lack analogy-based proposal mechanisms for navigating combinatorial topology spaces. On the other hand, neuro-symbolic methods~\cite{garcez2023neurosymbolic,chaudhuri2021neurosymbolic} have demonstrated the value of combining neural and symbolic reasoning, but have not been applied to engineering synthesis with verified physical constraints.

Our work also relates to grounded language model systems in robotics. SayCan~\cite{brohan2023can} grounds \ac{llm} prescriptions through skill affordances, but operates on action selection rather than design synthesis. HuggingGPT~\cite{shen2023hugginggpt} orchestrates specialist models via a controller pattern, an architectural precedent for our coordinator, but targets open-ended tasks rather than constrained engineering problems. The distinctive contribution of our framework is the combination of symbolic lifting with validated kinematic simulation in a closed loop, ensuring that every design iteration is physically verifiable.

\subsection{Limitations and future directions}

Some limitations should be noted. First, our evaluation is restricted to planar linkages; extending to spatial (3D) mechanisms will require richer symbolic vocabularies and more complex kinematic solvers. Second, while we demonstrated reflective reasoning through symbolic feedback, we did not fine-tune the language models on mechanism design corpora; doing so could further improve performance, especially for smaller models.

The reliance on open-source models, as recommended for reproducibility~\cite{llama3modelcard,qwen3}, means our findings are fully reproducible.

Future work should explore: (i) extension to spatial mechanisms and multi-body dynamics; (ii) integration with physics-informed neural networks for surrogate simulation~\cite{raissi2019physics}; (iii) human-in-the-loop workflows where the symbolic representations serve as an interpretable communication layer between AI agents and engineers; and (iv) systematic evaluation of the framework's explainability properties, using the symbolic reasoning traces as design rationales that can be formally audited against engineering standards.

\section{Methods}
\label{sec:method}

\subsection{Problem formulation}

The design task is interactive and language-driven: given a motion goal expressed in natural language, a team of language-model-based agents must propose, evaluate, and iteratively refine a planar linkage so that its end-effector follows the desired path. The problem decomposes into a discrete combinatorial subproblem (selecting a topology: which links and joints to use, and how to connect them) and a continuous subproblem (fitting link lengths, joint offsets, and crank angles to minimise the distance between the realised and target trajectories).

\subsection{Architecture overview}

The framework is organised into three layers (Fig.~\ref{fig:symb_lift}). At the top, a natural-language design intent drives the \ac{llm} agents. At the bottom, a validated kinematic simulator~\cite{Farajallah2024pylinkage} executes candidate linkages and returns sampled end-effector trajectories. Between these layers sits the symbolic lifting and optimisation layer, which (i)~translates dense numerical outputs into compact symbolic descriptions that \ac{llm} agents can reason about, and (ii)~performs continuous optimisation to fit a candidate topology to the target motion.

This architecture enforces a clear division of responsibilities: the \ac{llm} handles combinatorial search and analogical reasoning (what topology might work?), while the optimiser and simulator handle numerical precision and fitting of the performed to the targeted end-effector trajectory. The symbolic representation bundle $\mathcal{R}$ is the contract between these two worlds (formal definitions are given in Supplementary Section~\ref{sec:supp_definitions}).

\subsection{Symbolic lifting}
\label{sec:symbolic_lifting_methods}

The symbolic lifting operator $\mathfrak{L}$ compresses dense, noisy trajectory evidence and low-level simulator diagnostics into a small set of stable, semantically meaningful symbols. Let $\gamma:[0,1]\to\mathbb{R}^2$ denote the end-effector trajectory, $\{\mathbf{p}_i\}_{i=0}^N$ its sampled proxy, and \texttt{sim\_msg} the simulator diagnostics. The operator maps these to a representation bundle:
\[
\mathfrak{L}:\bigl(\gamma,\ \{\mathbf{p}_i\},\ \texttt{sim\_msg}\bigr)\longmapsto\mathcal{R}.
\]

The representation bundle $\mathcal{R} = (\mathcal{T},\ \mathcal{S},\ \mathcal{C})$ is factored into three components:
\begin{itemize}
    \item $\mathcal{T}$: a bounded vocabulary of atomic tokens (quantised headings, curvature signs, monotonicity flags, event tags).
    \item $\mathcal{S}$: composite sketches formed by composing tokens into ordered spatial-temporal summaries.
    \item $\mathcal{C}$: structural and mechanical information (degrees of freedom, bar counts, joint counts, feasibility diagnostics).
\end{itemize}

Two design principles govern $\mathcal{R}$. \textbf{Hysteresis} ensures that small perturbations in simulation output do not flip token values, so descriptions reflect persistent geometric structure rather than noise. \textbf{Compositionality} ensures that atomic tokens combine into sketches and temporal predicates, allowing complex motions to be described as structured combinations of simpler parts.

We implement two complementary lifting approaches:
\begin{enumerate}
    \item \textbf{Discrete Segmental Representation}: segments the trace into a concise sequence of high-level motion motifs (runs, turns, pauses), prioritising temporal organisation and data reduction. From sampled positions, local kinematic proxies (velocities, angular changes) are computed and mapped to labels $L_i \in \{\text{Pause}, \text{Straight}, \text{Gentle Turn}, \text{Sharp Turn}, \text{U-turn}\}$ using interpretable thresholds.
    \item \textbf{Compositional Lifting}: detects local geometric features (curvature sign, inflection points, self-intersections) and composes them into temporal predicates, prioritising qualitative geometric structure. The lifting also generates bounded temporal logic formulas that express qualitative intent as formal constraints.
\end{enumerate}

Full mathematical details of both approaches, including the qualitative signature, event alphabet, feature primitives, and temporal logic operators, are provided in Supplementary Sections~\ref{sec:supp_dr}--\ref{sec:supp_temporal}.

\subsection{Multi-agent pipeline}
\label{sec:agents_methods}

The synthesis pipeline comprises four cooperating agents that communicate through the shared representation bundle $\mathcal{R}$ (Algorithm~\ref{alg:topology_refinement}).

\begin{enumerate}
    \item \textbf{Topology Agent}: proposes discrete linkage hypotheses and supplies initial parameterisations, drawing on exemplar memory for analogical transfer and API contracts for structured output.
    \item \textbf{Simulation Critic}: fuses numerical quantities, symbolic surrogates from $\mathcal{R}$, and simulator diagnostics to identify trajectory error modes and report mobility anomalies, producing a structured report with evidence-backed blocks (Kinematic Accuracy, Mobility/DOF, Compositionality, Recommendation).
    \item \textbf{Refinement Planning Agent}: interprets the critic's diagnosis and maps it to corrective actions using a failure-mode reference table (e.g., underconstraint $\mapsto$ add a loop; overconstraint $\mapsto$ remove a redundant link), applying a minimal-change heuristic.
    \item \textbf{Refinement Agent}: operationalises the planner's instructions into a new mechanism, enforcing connectivity checks, DOF parity verification, and API compliance before re-optimisation.
\end{enumerate}

\begin{algorithm}[H]
\caption{Iterative Topology Refinement Loop} \label{alg:topology_refinement}
\begin{algorithmic}[1]
\FOR{episode $= 1$ to $N$}
    \STATE $T, p \gets \textsc{TopologyAgent}(I, \text{memory})$
    \STATE $\text{trace}, \text{diagnostics}, \text{distance} \gets \textsc{SimulatorOptimiser.run}(T, p)$
    \IF{$\text{distance} \le \epsilon$}
        \STATE \textbf{break}
    \ENDIF
    \STATE $R \gets \textsc{SymbolicLifting}(\text{trace}, \text{diagnostics})$
    \STATE $\text{report} \gets \textsc{Critic.report}(T, p, R, \text{diagnostics})$
    \STATE $O \gets \textsc{Planner.plan}(\text{report}, \text{failure\_table})$
    \STATE $T, p \gets \textsc{Refiner.apply}(T, O)$
    \STATE $\text{trace}, \text{diagnostics}, \text{distance} \gets \textsc{SimulatorOptimiser.run}(T, p)$
    \IF{$\text{distance} \le \epsilon$}
        \STATE \textbf{break}
    \ENDIF
\ENDFOR
\end{algorithmic}
\end{algorithm}

\subsection{Experimental setup}

We evaluated \ac{llama}~\cite{llama3modelcard}, \ac{qwen}, and \ac{qwen-moe}~\cite{qwen3} on six target shapes: Parabola, NACA airfoil, Line, Ellipse, Circle, and Lemniscate of Bernoulli (\ac{lb}). For every model--shape pair, we ran both optimisers (Grid and PSO) and toggled the \ac{llm} planner, discrete segmental representation, and compositional lifting to cover every combination. Models were sampled with temperature 0.8 using the \texttt{pylinkage} simulator~\cite{Farajallah2024pylinkage}.

The models were chosen to ensure architectural diversity: \ac{llama} serves as a widely-used single-expert baseline; \ac{qwen} provides a contemporary instruction-tuned model with different pretraining characteristics; and \ac{qwen-moe} probes mixture-of-experts routing. All models are open-source, enabling full reproducibility. The monolithic baseline is the method described in~\cite{gandarela2025controlledagenticplanning}.

At each optimisation iteration, three candidate mechanisms were drawn; the search used $R_{\max}=10$ refinement rounds and convergence threshold $\epsilon=0.005$. For \ac{qwen} (4B parameters), $R_{\max}$ was increased to 20 to compensate for its smaller capacity; despite the doubled iteration count, total computational cost remains a fraction of that for \ac{llama} (70B).

\subsection{Evaluation metrics}

For each shape, 5 independent samples were drawn and Chamfer distance was computed for each simulated/target pair after alignment via the \ac{icp} algorithm~\cite{richardos_icp,lu1997robot}. Reported scores are mean $\pm$ standard error across the five samples.

Semantic success is a binary indicator: a generated mechanism counts as successful if it both parses and executes in the simulator without parse, compile, or runtime errors. All reported results are averages over runs for each model--shape configuration.

Specific kinematic thresholds for the symbolic representations are detailed in Supplementary Section~\ref{sec:supp_thresholds}.

\printbibliography

\clearpage
\appendix 
\section{Supplementary Information}
\label{sec:supplementary}

\subsection{Formal Definitions}
\label{sec:supp_definitions}

\begin{defbox}{Intent \(I\)}{intent}
An \emph{intent} \(I\) is a short natural-language description (optionally with example traces) specifying the desired motion goal.
\end{defbox}

\begin{defbox}{End-effector trace \(\mathbf p\)}{endeff}
The end-effector trace is the sampled planar trajectory
\(\mathbf p=(\mathbf p_0,\dots,\mathbf p_N)\) with \(\mathbf p_i\in\mathbb R^2\), produced by running the candidate mechanism in the simulator.
\end{defbox}

\begin{defbox}{Optimiser output \(p^\star(T)\)}{opt}
For a fixed topology \(T\), the optimiser returns
\[
p^\star(T)\in\arg\min_{p\in\mathcal P(T)}\mathcal L\bigl(\tau(T,p), \tau^\star\bigr),
\]
where \(\tau(T,p)\) is the realised trajectory, \(\tau^\star\) the target, \(\mathcal P(T)\) the feasible parameter domain, and \(\mathcal L\) the task loss (Chamfer distance).
\end{defbox}

\begin{propbox}{Existence of optimiser solution}{prop:opt_existence}
Under standard regularity conditions ($\mathcal P(T)$ non-empty and compact, $\mathcal L$ continuous in $p$), the argmin is non-empty. Hence $p^\star(T)$ exists.
\end{propbox}

\begin{defbox}{Representation bundle \(\mathcal R\)}{repr}
The representation bundle \(\mathcal R\) is the set of symbolic descriptions consumed by the \ac{llm} agents. It collects motion labels, detected events, structural diagnostics, and temporal predicates derived from the simulator output:
\[
\mathcal R=\bigl(\{q(t)\}_t,\ \mathrm{events}\subseteq\mathcal E,\ \{L_i\}_i,\ \{\mathfrak p_j\}_j,\ \{\Phi_k\}_k\bigr),
\]
where \(\{L_i\}_i\) denotes discrete motion labels and segmentation summaries.
\end{defbox}

\begin{defbox}{Symbolic lifting operator \(\mathfrak L\)}{symbolic_lifting}
\(\mathfrak L\) is the operator that maps an end-effector trajectory (or its sample proxy) together with simulator diagnostics to the compact representation bundle \(\mathcal R\) used by downstream agents:
\[
\mathfrak L:\bigl(\gamma\, \ \{\mathbf p_i\},\ \texttt{sim\_msg}\bigr)\longmapsto\mathcal R.
\]
\end{defbox}

\begin{propbox}{Soundness of symbolic lifting}{prop:lifting_soundness}
If a simulator trace satisfies a high-confidence property (e.g., ``entirely in region $R$'' or ``contains an inflection at time $t$'') with sufficient margin above detection thresholds, then the corresponding lifted predicate appears in $\mathcal R$. That is, strong geometric facts are faithfully preserved by the lifting operator $\mathfrak L$.
\end{propbox}

\subsection{Discrete Segmental Representation}
\label{sec:supp_dr}

From the sampled positions $\{\mathbf p_i\}$ we form local kinematic proxies
\[
\mathbf v_i=\mathbf p_{i+1}-\mathbf p_i,\qquad s_i=\|\mathbf v_i\|_2,
\]
and capture local direction and angular change by
\[
\theta_i=\operatorname{atan2}(v_{i,y},v_{i,x}),\qquad
c_i=\bigl|\operatorname{unwrap}(\theta_{i+1}-\theta_i)\bigr|.
\]
Headings are quantised $d_i=[\theta_i]_\Pi$ and the instantaneous motion is mapped to labels
$L_i\in\{\mathrm{Pause},\mathrm{Straight},\mathrm{Gentle\ Turn},\mathrm{Sharp\ Turn},\mathrm{U\!-\!turn}\}$
using interpretable thresholds $(\tau_{\mathrm{pause}},\tau_{\mathrm{straight}},\tau_{\mathrm{turn}},\dots)$. Runs of identical labels collapse the raw trace into segments, and for each segment the lifting records dominant heading, dominant curvature sign, label frequency, mean run length, transition counts and label-sequence entropy.

\subsection{Compositional Lifting: Qualitative Observables}
\label{sec:supp_cl}

From the sampled trajectory points $\{\mathbf p_i\}$, discrete velocity $\mathbf v_i$ and acceleration $\mathbf a_i$ vectors are computed using centred finite difference approximations. The instantaneous observables, speed $v_i$ and geometric curvature $\kappa_i$, are:
\[
v_i=\|\mathbf v_i\|_2,\qquad
\kappa_i=\frac{\dot x_i\,\ddot y_i-\dot y_i\,\ddot x_i}{(\dot x_i^2+\dot y_i^2)^{3/2}}.
\]

Velocities are estimated by first-order finite differences: forward/backward at boundaries, centred for interior samples:
\[
\mathbf{v}_i \approx
\begin{cases}
(\mathbf{p}_1-\mathbf{p}_0)/\Delta t, & i=0,\\[4pt]
(\mathbf{p}_{i+1}-\mathbf{p}_{i-1})/(2\Delta t), & 0<i<n-1,\\[4pt]
(\mathbf{p}_{n-1}-\mathbf{p}_{n-2})/\Delta t, & i=n-1.
\end{cases}
\]

Accelerations are obtained from second-order finite differences:
\[
\mathbf{a}_i \approx \frac{\mathbf{p}_{i+1}-2\mathbf{p}_i+\mathbf{p}_{i-1}}{\Delta t^2}.
\]

Applying hysteresis produces a noise-robust signature: the hysteretic sign function yields curvature polarity $s_{\kappa,i}=\operatorname{sgn}_\epsilon(\kappa_i)$ and velocity-monotonicity components $m_{x,i},m_{y,i}$.

\begin{defbox}{Qualitative signature $q(t)$}{qual_sig}
The qualitative signature $q(t)$ is the tuple of pointwise tokens
\[
q(t)=\bigl(\,s_\kappa(t),\,m_x(t),\,m_y(t)\bigr),
\]
where $s_\kappa$ is the hysteretic curvature polarity, and $m_x,m_y$ are velocity monotonicity components.
\end{defbox}

\begin{defbox}{Event alphabet $\mathcal E$}{events}
$\mathcal E$ is the finite set of atomic events: $\mathrm{INF}$ (inflection, hysteretic zero-crossing of $\kappa$), $\mathrm{EX}_x,\mathrm{EX}_y$ (coordinate extrema), $\mathrm{SINT}$ (self-intersection), region relations $\mathrm{in/out/cross}$, and guard crossings. Each is time-stamped.
\end{defbox}

\begin{defbox}{Feature primitive $\mathfrak p$}{primitive}
A feature primitive is a compact interval descriptor:
\[
\mathfrak p=\big\langle\texttt{curv}\in\{-1,0,+1\},\ \texttt{mono}\in\{-1,+1\}^2,\ \texttt{len}\in\mathbb R_{\ge0}\cup\{\ast\},\ \texttt{ev}\subseteq\mathcal E\big\rangle.
\]
\end{defbox}

\begin{propbox}{Compositionality and robustness of sketches}{prop:sketch_composition}
Let sketches be formed by concatenating primitives under majority statistics and hysteresis. Then:
\begin{enumerate}
  \item The concatenation of primitives yields sketches whose qualitative event ordering and dominant curvature signs coincide with the concatenation of the primitives' qualitative signatures (compositionality).
  \item If the original numeric trajectory is perturbed by $\Delta$ with $\|\Delta\|_\infty\le\epsilon$ and the hysteresis margins exceed $\epsilon$, then the resulting sketch is unchanged (robustness).
\end{enumerate}
\end{propbox}

\subsection{Temporal Logic Operators}
\label{sec:supp_temporal}

Events and primitives are mapped into bounded temporal logic formulas:

\begin{defbox}{Bounded temporal operators}{temporal_ops}
$\mathsf F_{[a,b]}\varphi$ (eventually in $[a,b]$), $\mathsf G_{[a,b]}\varphi$ (always in $[a,b]$), and $\varphi\ \mathsf U_{[a,b]}\ \psi$ (until within $[a,b]$):
\[
\begin{aligned}
F_{[a,b]} \, \varphi &\quad\text{: } \exists t' \in [a,b] \text{ with } \varphi \text{ true.} \\[6pt]
G_{[a,b]} \, \varphi &\quad\text{: } \forall t' \in [a,b], \; \varphi \text{ true.} \\[6pt]
\varphi \, U_{[a,b]} \, \psi &\quad\text{: } \exists t' \in [a,b] \text{ with } \psi \text{ true and } 
  \forall t'' \in [0,t'], \; \varphi \text{ true.}
\end{aligned}
\]
\end{defbox}

Example formula:
\[
\Phi \equiv \mathsf F_{[0,0.4]}(\mathrm{in}(R_A))\ \wedge\ \mathsf F_{[0.4,0.7]}(\texttt{curv}=0)\ \wedge\ (\neg\mathrm{cross}(L_x)\ \mathsf U_{[0.7,1.0]}\ \mathrm{in}(R_B)).
\]

\subsection{Agent Formal Definitions}
\label{sec:supp_agents}

\begin{defbox}{Topology Agent mapping \(f_{\mathrm{topology}}\)}{topology}
\[
f_{\mathrm{topology}}\bigl(M, n_{\mathrm{bars}}, \{\mathbf p_i\}, \mathrm{mem}, \mathrm{exs}\bigr)
\mapsto (T,p_{\mathrm{init}},\texttt{rationale}).
\]
\end{defbox}

\begin{defbox}{Simulation Critic mapping \(f_{\mathrm{critic}}\)}{critic}
\[
f_{\mathrm{critic}}:\bigl((T,p),\ \texttt{sim\_mes},\ \mathcal{L},\ \mathrm{DOF},\ \mathrm{TA\_response},\ \mathcal R\bigr)
\longmapsto r_{\mathrm{text}}.
\]
\end{defbox}

\begin{defbox}{Refinement Planner mapping \(f_{\mathrm{plan}}\)}{planner}
\[
f_{\mathrm{plan}}:\bigl(\mathrm{TA\_response}, (T,p),\ r_{\mathrm{text}},\ \mathrm{failure\_table},\ \mathrm{mem}\bigr)
\longmapsto O,
\]
where \(O=\{\ \texttt{failure\_mode},\ \texttt{structural\_cause},\ \texttt{suggested\_action}\ \}\).
\end{defbox}

\begin{defbox}{Refinement operator \(\mathcal F_{\mathrm{refine}}\)}{refine}
\[
\mathcal{F}_{\mathrm{refine}}\!\Bigl(\; (T,p),\,O,\,\texttt{sim\_mes},\,\mathcal{L},\,\mathrm{DOF},\,\mathcal R,\,\mathrm{mem},\,\mathrm{api\_doc}\Bigr)
\to \bigl((T',p'),\ \texttt{chg\_rationale},\ \texttt{reuse\_notes}\bigr).
\]
\end{defbox}

\begin{propbox}{Bounded improvement of the closed-loop}{prop:closed_loop_improvement}
Assume each refinement plan \(O\) selected by the planner enforces either (a) a change that resolves a certificate-flagged hard constraint, or (b) a minimal parametric/topological edit which the optimiser re-fits to locally minimise \(\mathcal L\).
Under a minimal-change heuristic and assuming the optimiser attains \(p^\star(T)\) for each proposed topology,
the loop yields a sequence of candidate designs whose task loss is non-increasing across accepted iterations, and any accepted edit resolving a hard violation strictly reduces the set of outstanding certificate failures.
\end{propbox}

\subsection{Experimental Thresholds}
\label{sec:supp_thresholds}

For the discrete segmental representation, trajectory segments were classified based on heading deviations: $2^{\circ}$ for straight paths, $30^{\circ}$ for gentle turns, and $45^{\circ}$ for sharp turns. Velocities below $1.5 \times 10^{-4}$ were categorised as pauses.

For the compositional lifting, the simulation time step was $\Delta t = 1.0$. A tolerance of $10^{-3}$ was used for identifying zero curvature and sign changes. Guard crossings were validated using a minimum normal velocity of $10^{-4}$ and a sample separation of 5. Self-intersection (SINT) events used a stochastic sampling strategy retaining 10\% of detected intersections to avoid pathologically verbose descriptions. The temporal logic synthesis employed half-sample padding for event windows and an interval merging tolerance of $10^{-12}$.

At each optimisation iteration, three candidate mechanisms were drawn; search used $R_{\max}=10$ and $\epsilon=0.005$. For \ac{qwen} (4B parameters), $R_{\max}$ was increased to 20.

\section{Qualitative Analysis}
\label{app:Qualitative_Analysis}

\begin{figure}[!htbp]
    \centering
    \begin{subfigure}[t]{0.38\textwidth}
        \includegraphics[width=\linewidth]{imgs/shapes/_naca_qwen3_30b_pso_8_half_frame6952337_1.jpg}
        \caption{\ac{qwen-moe} with PSO.}
        \label{fig:naca_qwen3_30b_pso}
    \end{subfigure}
    \begin{subfigure}[t]{0.38\textwidth}
        \includegraphics[width=\linewidth]{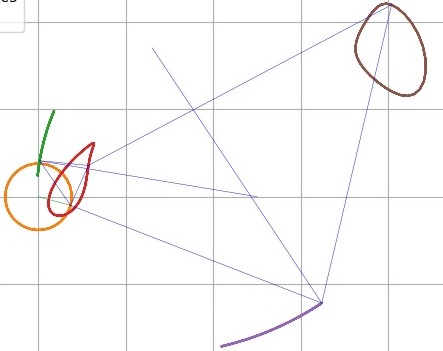}
        \caption{\ac{llama} with Grid search.}
        \label{fig:naca_llama3.3_70b_grid}
    \end{subfigure}
    \begin{subfigure}[t]{0.22\textwidth}
        \includegraphics[width=\linewidth]{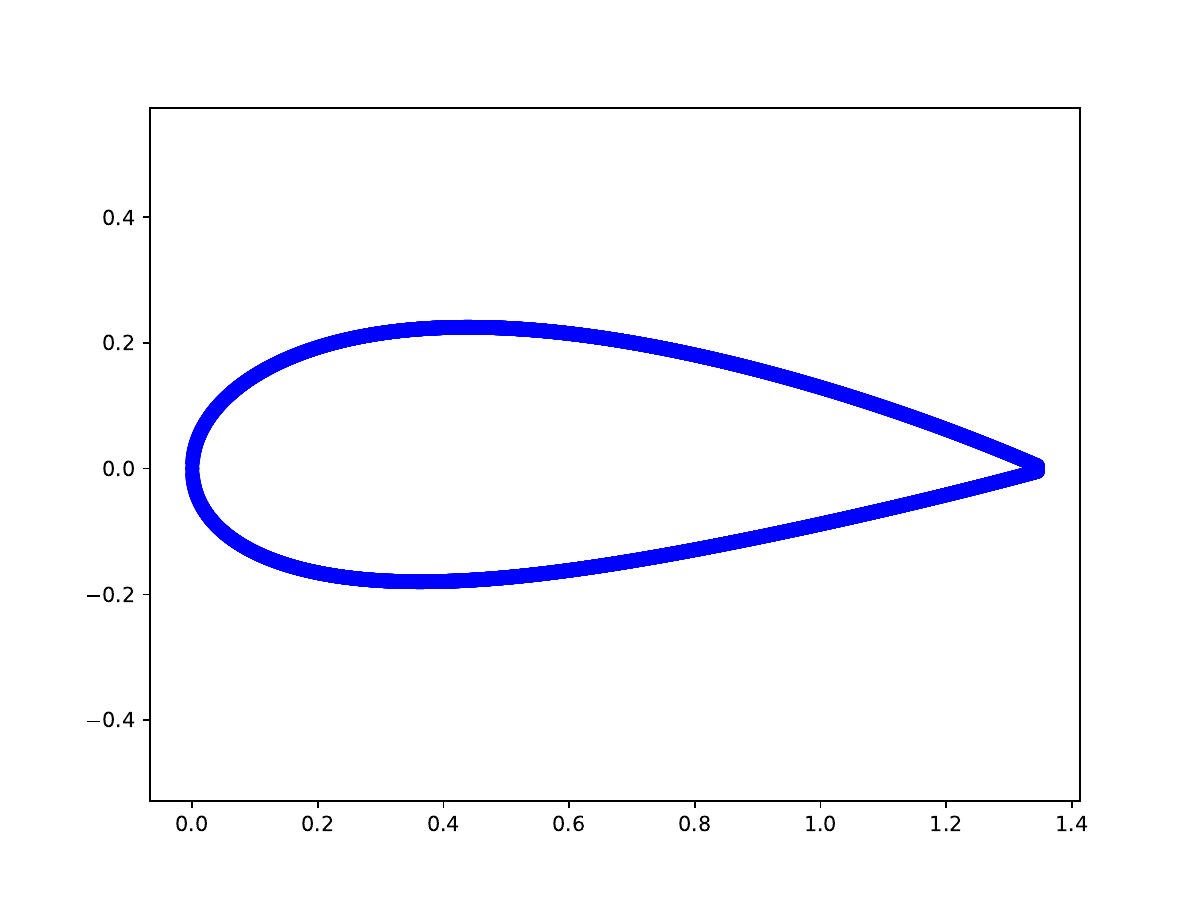}
        \caption{\textbf{Target NACA profile (ground truth).} This is the desired curve the mechanism should trace, it is not a synthesised output.}
        \label{fig:naca_target}
    \end{subfigure}
    \\
    \begin{subfigure}[t]{0.38\textwidth}
        \includegraphics[width=\linewidth]{imgs/shapes/_8_equation_qwen3_30b_pso_8_half_frame9549290_1.jpg}
        \caption{\ac{qwen-moe} with PSO.}
        \label{fig:8_equation_qwen3_30b_pso}
    \end{subfigure}
    \begin{subfigure}[t]{0.38\textwidth}
        \includegraphics[width=\linewidth]{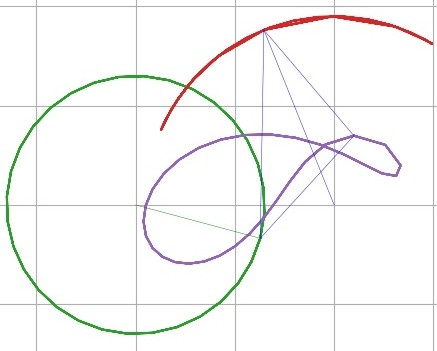}
        \caption{\ac{llama} with Grid search.}
        \label{fig:8_equation_llama3.3_70b_grid}
    \end{subfigure}
    \begin{subfigure}[t]{0.22\textwidth}
        \includegraphics[width=\linewidth]{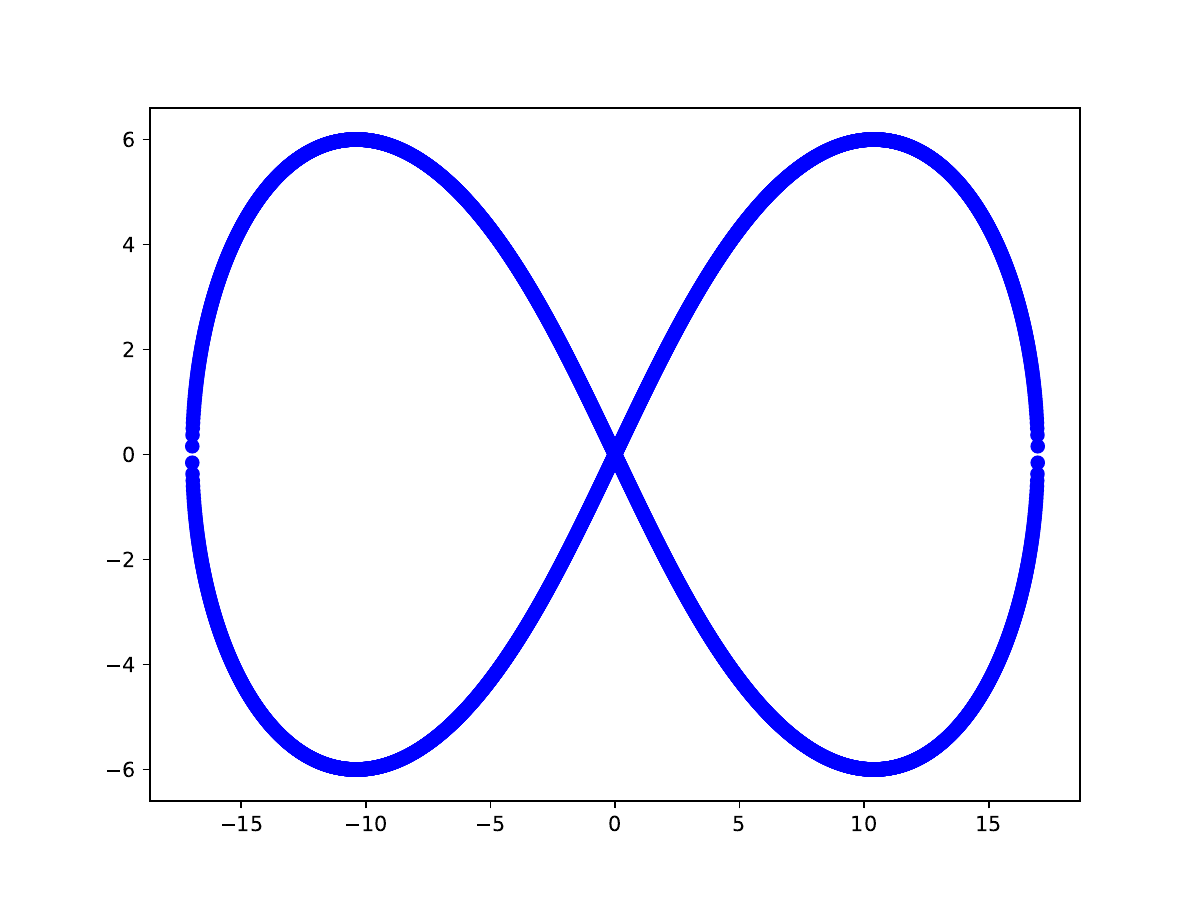}
        \caption{\textbf{Target \ac{lb} curve (ground truth).} This is the desired curve, not a synthesised mechanism.}
        \label{fig:lemniscate_target}
    \end{subfigure}
    \caption{NACA airfoil and Lemniscate of Bernoulli synthesis results. \textbf{Thin lines}: rigid bars of the mechanism. \textbf{Thick coloured lines}: full-cycle joint trajectories; different colours correspond to different joints, with the end-effector trace being the most prominent. Panel~\ref{fig:naca_target} is the \textbf{ground-truth target curve} that the synthesised mechanisms in~\ref{fig:naca_qwen3_30b_pso} and~\ref{fig:naca_llama3.3_70b_grid} are approximating, it shows the ideal NACA profile, not a generated mechanism. The generated trajectories (\ref{fig:naca_qwen3_30b_pso}, \ref{fig:naca_llama3.3_70b_grid}) capture the asymmetric camber and trailing-edge curvature of the target profile \ref{fig:naca_target}. Both model--optimiser combinations reproduce the global airfoil silhouette, with the leading-edge curvature and overall loop closure closely matching the target. Minor deviations appear in local curvature near inflection regions, but the overall geometric fidelity is high, confirming that the method can synthesise mechanisms that approximate complex parametric curves. Panel~\ref{fig:lemniscate_target} is the \textbf{ground-truth target curve}, the ideal Lemniscate of Bernoulli that the mechanisms in~\ref{fig:8_equation_qwen3_30b_pso} and~\ref{fig:8_equation_llama3.3_70b_grid} are reproducing, not an output of the method. This target is a self-intersecting figure-eight curve, which requires the mechanism to produce a trajectory that crosses itself. Both generated outputs (\ref{fig:8_equation_qwen3_30b_pso}, \ref{fig:8_equation_llama3.3_70b_grid}) reproduce the characteristic double-loop topology and a crossing point, demonstrating that the pipeline can synthesise linkages for self-intersecting targets.}
    \label{figs:naca_lemniscate}
\end{figure}

\begin{figure}[!htbp]
    \centering
    \begin{subfigure}[t]{0.55\textwidth}
        \includegraphics[width=\linewidth]{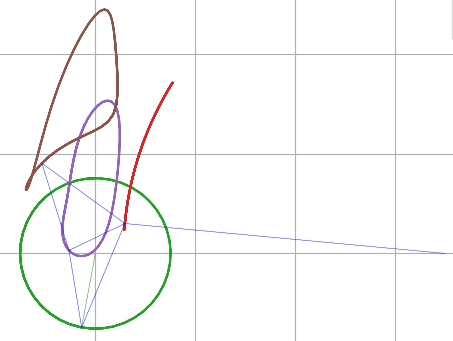}
        \caption{\ac{qwen-moe} with PSO.}
        \label{fig:parabola_qwen3_30b_pso}
    \end{subfigure}
    \begin{subfigure}[t]{0.35\textwidth}
        \includegraphics[width=\linewidth]{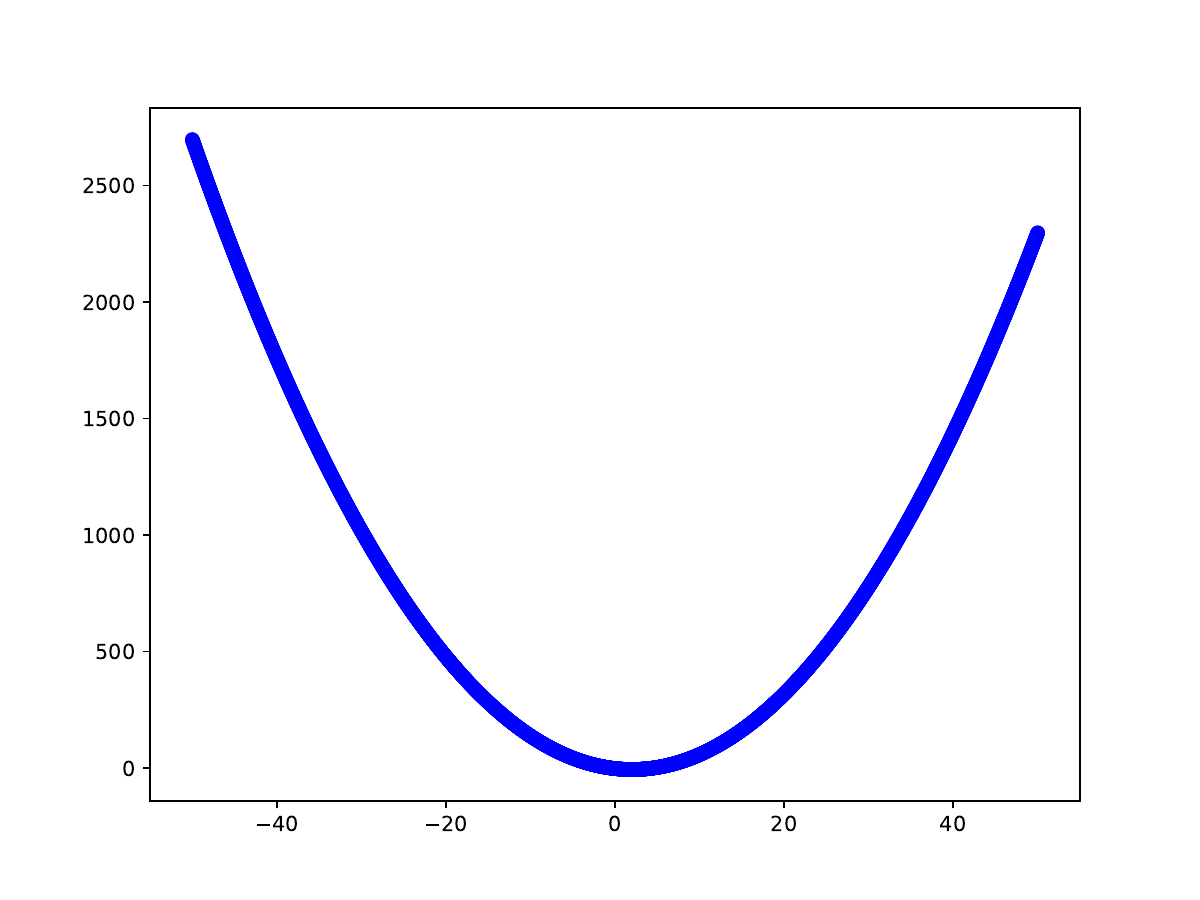}
        \caption{\textbf{Target parabolic arc (ground truth).} This is the desired curve, not a synthesised mechanism.}
        \label{fig:parabola_target}
    \end{subfigure}
    \caption{Parabola synthesis results. \textbf{Thin lines}: rigid bars of the linkage. \textbf{Thick coloured lines}: full-cycle joint trajectories; each colour identifies a distinct joint, and the end-effector trace is the most prominent. Panel~(b) is the \textbf{ground-truth target curve}, the ideal parabolic arc the mechanism in~(a) should trace, not an output of the method. The generated trajectory (a) approximates this parabolic arc with correct global curvature sign and vertex placement. Open curves present a specific challenge for linkage synthesis because the end-effector of a rotary-driven mechanism inherently produces closed paths.}
    \label{figs:parabola}
\end{figure}

\begin{figure}[!htbp]
    \centering
    \begin{subfigure}[t]{0.38\textwidth}
        \includegraphics[width=\linewidth]{imgs/shapes/_line_qwen3_4b_grid_6_half_frame3895452_1.jpg}
        \caption{\ac{qwen} with Grid search.}
        \label{fig:line_qwen3_4b_grid}
    \end{subfigure}
    \begin{subfigure}[t]{0.38\textwidth}
        \includegraphics[width=\linewidth]{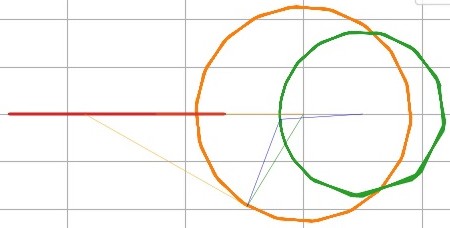}
        \caption{\ac{llama} with Grid search.}
        \label{fig:line_llama3.3_70b_grid}
    \end{subfigure}
    \begin{subfigure}[t]{0.22\textwidth}
        \includegraphics[width=\linewidth]{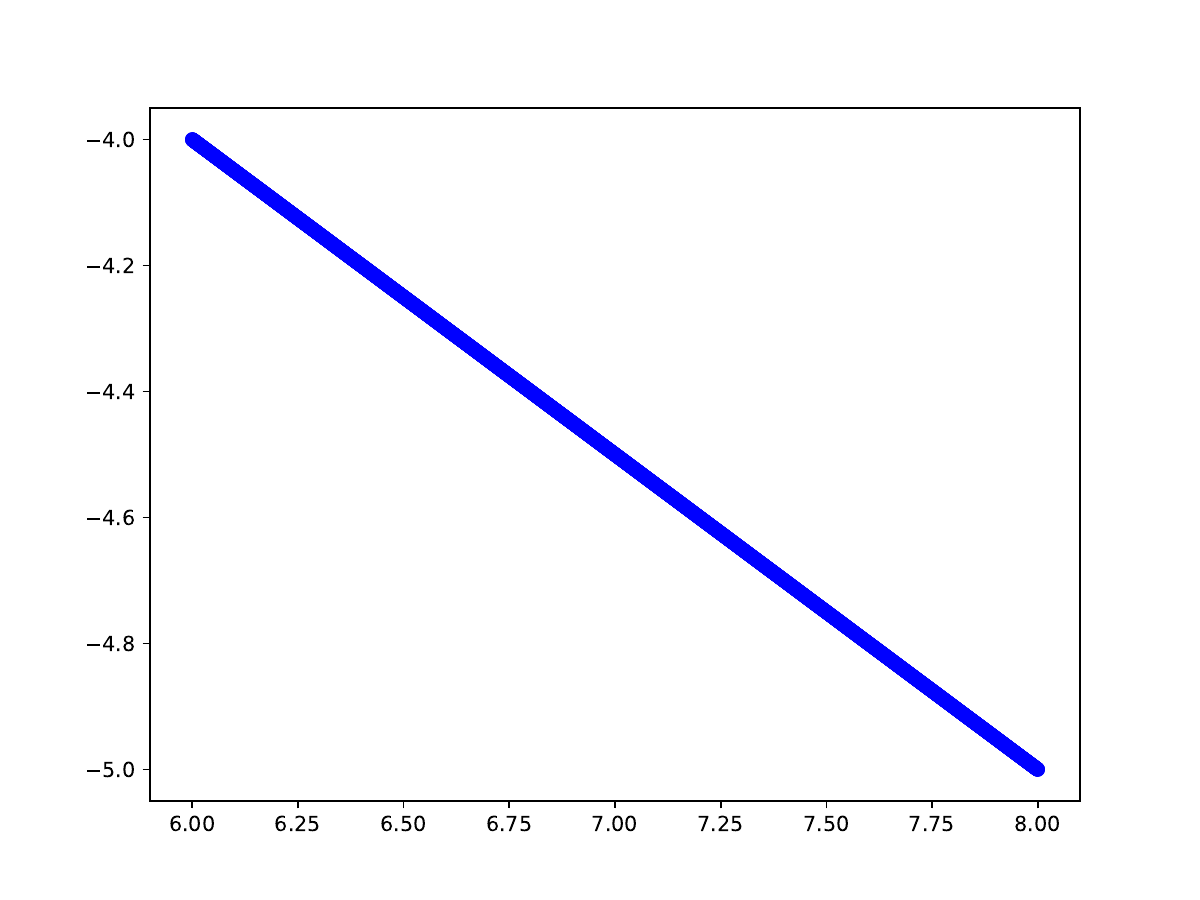}
        \caption{\textbf{Target straight line (ground truth).} This is the desired curve, not a synthesised mechanism.}
        \label{fig:line_target}
    \end{subfigure}
    \\
    \begin{subfigure}[t]{0.38\textwidth}
        \includegraphics[width=\linewidth]{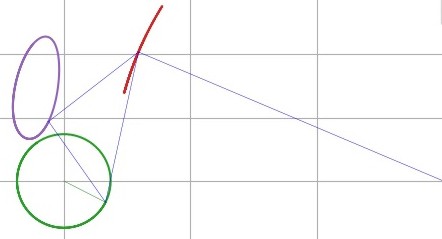}
        \caption{\ac{qwen-moe} with PSO (sample~1).}
        \label{fig:ellipse_qwen3_30b_pso}
    \end{subfigure}
    \begin{subfigure}[t]{0.38\textwidth}
        \includegraphics[width=\linewidth]{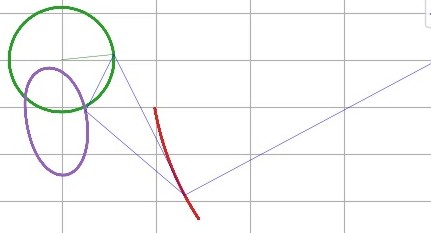}
        \caption{\ac{qwen-moe} with PSO (sample~2).}
        \label{fig:ellipse_qwen3_30b_pso_2}
    \end{subfigure}
    \begin{subfigure}[t]{0.22\textwidth}
        \includegraphics[width=\linewidth]{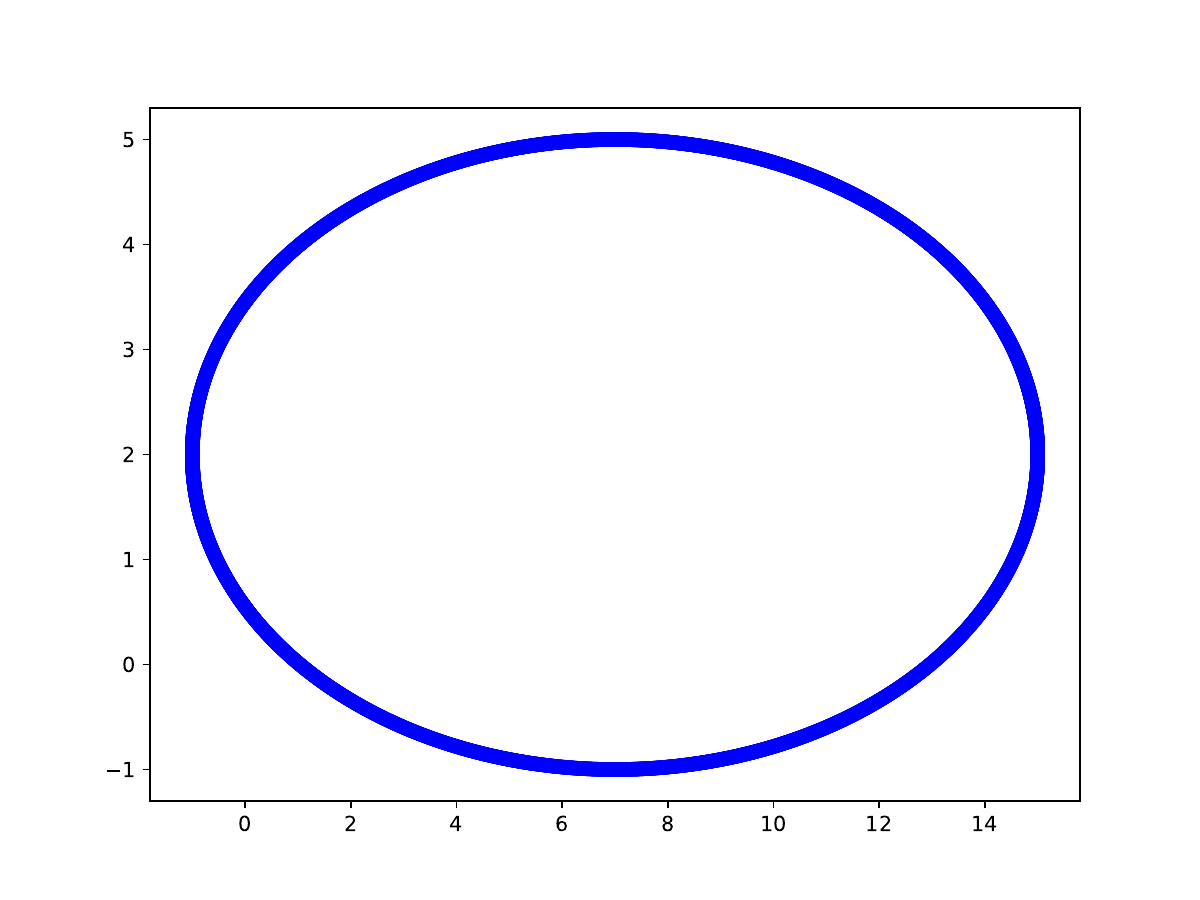}
        \caption{\textbf{Target ellipse (ground truth).} This is the desired curve, not a synthesised mechanism.}
        \label{fig:ellipse_target}
    \end{subfigure}
    \caption{Straight-line and ellipse synthesis results. \textbf{Thin lines}: rigid bars of the linkage. \textbf{Thick coloured lines}: full-cycle joint trajectories; colours distinguish joints, with the end-effector trace being the most prominent. Panels~(a)-(c) show straight-line synthesis: panel~(c) is the \textbf{ground-truth target curve}, the ideal straight-line segment that panels~(a) and~(b) approximate. The generated trajectories achieve linear paths, and the 4B-parameter \ac{qwen} result is comparable to the 70B-parameter \ac{llama} result, supporting the finding that the symbolic interface, rather than model scale, is the primary driver of design quality; this target achieved the largest Chamfer distance reduction (67.7\%) in Table~\ref{tab:combined_all}. Panels~(d)-(f) show ellipse synthesis: panel~(f) is the \textbf{ground-truth target curve}, and panels~(d) and~(e) are two independent samples from the same model-optimiser configuration, illustrating output diversity. Both samples capture a closed smooth elliptical profile while differing in aspect ratio and orientation, consistent with stochastic topology exploration and subsequent continuous fitting.}
    \label{figs:line}
    \label{figs:ellipse}
\end{figure}

This section provides a qualitative visual assessment of the end-effector trajectories generated by our method for each of the six benchmark motion targets. Figures~\ref{figs:naca_lemniscate}--\ref{figs:ellipse} compare the synthesised trajectories, produced by different model and optimiser combinations, against the corresponding target curves. Each figure juxtaposes one or two generated outputs (left and centre panels) with the ground-truth target (right panel), enabling direct visual evaluation of geometric fidelity.

Close inspection of the generated trajectories reveals several noteworthy characteristics. First, the method captures global shape structure: topological features such as closure, symmetry, and curvature sign are correctly reproduced and align well with target specifications across all test cases. Second, local geometric features are preserved with fidelity in the majority of outputs. The NACA airfoil examples (Figure~\ref{figs:naca_lemniscate}) are particularly illustrative: despite their parametric complexity, involving asymmetric camber and trailing-edge curvature, the generated mechanisms successfully approximate key visual signatures such as leading-edge sharpness and looping periodicity. The Lemniscate of Bernoulli results (Figure~\ref{figs:naca_lemniscate}) demonstrate that the method handles self-intersecting curves, a particularly challenging class of targets for linkage synthesis. The Line (Figure~\ref{figs:line}) results confirm effective performance on open curves with low curvature, while the Ellipse (Figure~\ref{figs:ellipse}) results show that the method generalises to closed, smooth curves with varying aspect ratios.

These visual results corroborate the quantitative findings reported in Section~\ref{sec:results}: the modular architecture, combined with symbolic lifting, produces mechanisms whose end-effector paths closely follow the target curves across a diverse set of geometric profiles. The consistency across different model families and optimiser choices reinforces the conclusion that the symbolic representation interface, rather than any particular model, is the primary enabler of geometric fidelity.

\paragraph{How to read the mechanism plots.}
Each visualisation shows the synthesised planar linkage at a representative configuration of its motion cycle. \textbf{Thin lines} represent the rigid links (bars) that constitute the mechanism's skeleton; their connectivity encodes the topology proposed by the language model agent. \textbf{Thick lines} trace the complete trajectories swept by each joint as the crank completes one full rotation. \textbf{Different colours} identify different joints: each joint in the mechanism traces a distinct coupler curve, and the colour distinguishes which trajectory belongs to which joint. The end-effector joint, whose trajectory is intended to approximate the target curve, is highlighted by the thickest and most prominently coloured trace. Where a target curve is shown in the right-hand panel, it is plotted in black for reference.

\clearpage

\section{Full table}
\label{app:full_table}

\begin{table*}
\centering
\resizebox{\columnwidth}{!}{\begin{tabular}{@{}lllllllllllll@{}}
  \hline
  Model & Shape & Opt & Planner & \acs{dr} & \acs{cl} & Best chamf. & Steps & \% Imp. & \% Semantic & Links & Goal links \\
  \hline
\ac{llama} & \ac{lb} & Grid & No & No & No & $6.529 \pm 0.450$ & $5.400 \pm 1.631$ & $17.875$ & $0.885 \pm 0.010$ & $6.520 \pm 0.138$ & $2.200 \pm 0.183$ \\
\ac{llama} & \ac{lb} & Grid & No & No & Yes & $6.145 \pm 0.494$ & $6.000 \pm 1.817$ & $16.207$ & $0.883 \pm 0.052$ & $6.880 \pm 0.142$ & $2.360 \pm 0.166$ \\
\ac{llama} & \ac{lb} & Grid & No & Yes & No & $6.792 \pm 0.167$ & $5.200 \pm 1.319$ & $14.742$ & $1.013 \pm 0.057$ & $6.727 \pm 0.160$ & $2.341 \pm 0.203$ \\
\ac{llama} & \ac{lb} & Grid & No & Yes & Yes & $5.966 \pm 0.366$ & $6.000 \pm 1.517$ & $26.488$ & $0.983 \pm 0.019$ & $7.040 \pm 0.143$ & $2.800 \pm 0.137$ \\
\ac{llama} & \ac{lb} & Grid & Yes & No & No & $6.322 \pm 0.619$ & $4.600 \pm 1.208$ & $15.399$ & $0.846 \pm 0.078$ & $8.280 \pm 0.323$ & $3.790 \pm 0.776$ \\
\ac{llama} & \ac{lb} & Grid & Yes & No & Yes & $6.836 \pm 0.540$ & $3.400 \pm 1.288$ & $17.190$ & $1.019 \pm 0.123$ & $7.200 \pm 0.181$ & $2.330 \pm 0.669$ \\
\ac{llama} & \ac{lb} & Grid & Yes & Yes & No & $6.581 \pm 0.600$ & $5.600 \pm 1.691$ & $23.985$ & $1.097 \pm 0.135$ & $6.800 \pm 0.171$ & $2.050 \pm 0.708$ \\
\ac{llama} & \ac{lb} & Grid & Yes & Yes & Yes & $6.022 \pm 0.643$ & $4.800 \pm 1.562$ & $27.508$ & $0.983 \pm 0.078$ & $7.320 \pm 0.211$ & $2.510 \pm 0.701$ \\
\ac{llama} & \ac{lb} & Pso & No & No & No & $8.653 \pm 0.571$ & $5.000 \pm 0.837$ & $7.852$ & $0.852 \pm 0.025$ & $6.500 \pm 0.167$ & $2.040 \pm 0.140$ \\
\ac{llama} & \ac{lb} & Pso & No & No & Yes & $8.736 \pm 0.554$ & $4.400 \pm 1.400$ & $5.666$ & $0.915 \pm 0.054$ & $6.720 \pm 0.149$ & $2.340 \pm 0.168$ \\
\ac{llama} & \ac{lb} & Pso & No & Yes & No & $8.791 \pm 0.589$ & $7.000 \pm 1.643$ & $6.749$ & $1.001 \pm 0.108$ & $7.440 \pm 0.162$ & $3.040 \pm 0.162$ \\
\ac{llama} & \ac{lb} & Pso & No & Yes & Yes & $8.621 \pm 0.416$ & $4.000 \pm 1.304$ & $6.943$ & $0.924 \pm 0.030$ & $6.640 \pm 0.145$ & $2.160 \pm 0.163$ \\
\ac{llama} & \ac{lb} & Pso & Yes & No & No & $8.807 \pm 0.487$ & $4.800 \pm 1.241$ & $6.899$ & $0.862 \pm 0.037$ & $6.720 \pm 0.179$ & $2.170 \pm 0.674$ \\
\ac{llama} & \ac{lb} & Pso & Yes & No & Yes & $8.626 \pm 0.546$ & $7.800 \pm 0.583$ & $10.785$ & $0.904 \pm 0.038$ & $7.160 \pm 0.190$ & $2.690 \pm 0.679$ \\
\ac{llama} & \ac{lb} & Pso & Yes & Yes & No & $8.689 \pm 0.419$ & $4.400 \pm 1.400$ & $8.121$ & $0.908 \pm 0.143$ & $6.600 \pm 0.237$ & $2.210 \pm 0.729$ \\
\ac{llama} & \ac{lb} & Pso & Yes & Yes & Yes & $8.687 \pm 0.441$ & $4.400 \pm 1.400$ & $8.756$ & $0.930 \pm 0.031$ & $7.380 \pm 0.235$ & $2.530 \pm 0.748$ \\
\ac{llama} & Circle & Grid & No & No & No & $1.735 \pm 0.500$ & $4.600 \pm 1.503$ & $58.708$ & $1.047 \pm 0.093$ & $6.640 \pm 0.133$ & $2.180 \pm 0.178$ \\
\ac{llama} & Circle & Grid & No & No & Yes & $1.506 \pm 0.559$ & $8.400 \pm 0.812$ & $64.091$ & $0.892 \pm 0.021$ & $6.600 \pm 0.143$ & $2.320 \pm 0.158$ \\
\ac{llama} & Circle & Grid & No & Yes & No & $1.674 \pm 0.545$ & $7.400 \pm 1.435$ & $75.051$ & $1.057 \pm 0.150$ & $6.560 \pm 0.152$ & $2.120 \pm 0.150$ \\
\ac{llama} & Circle & Grid & No & Yes & Yes & $1.663 \pm 0.522$ & $5.800 \pm 1.114$ & $70.053$ & $0.991 \pm 0.056$ & $7.040 \pm 0.174$ & $2.560 \pm 0.194$ \\
\ac{llama} & Circle & Grid & Yes & No & No & $1.768 \pm 0.508$ & $7.200 \pm 1.200$ & $69.799$ & $0.954 \pm 0.079$ & $7.120 \pm 0.173$ & $2.910 \pm 0.669$ \\
\ac{llama} & Circle & Grid & Yes & No & Yes & $1.797 \pm 0.453$ & $8.400 \pm 0.400$ & $68.511$ & $0.981 \pm 0.087$ & $7.280 \pm 0.196$ & $2.650 \pm 0.692$ \\
\ac{llama} & Circle & Grid & Yes & Yes & No & $1.711 \pm 0.409$ & $4.800 \pm 1.594$ & $80.335$ & $0.907 \pm 0.023$ & $6.680 \pm 0.168$ & $1.750 \pm 0.664$ \\
\ac{llama} & Circle & Grid & Yes & Yes & Yes & $1.579 \pm 0.451$ & $6.000 \pm 1.183$ & $68.959$ & $1.010 \pm 0.103$ & $7.140 \pm 0.232$ & $2.550 \pm 0.714$ \\
\ac{llama} & Circle & Pso & No & No & No & $7.458 \pm 3.203$ & $8.000 \pm 1.304$ & $40.772$ & $1.321 \pm 0.125$ & $7.120 \pm 0.142$ & $2.920 \pm 0.140$ \\
\ac{llama} & Circle & Pso & No & No & Yes & $6.221 \pm 2.962$ & $8.400 \pm 0.678$ & $59.205$ & $0.988 \pm 0.177$ & $6.400 \pm 0.214$ & $2.420 \pm 0.176$ \\
\ac{llama} & Circle & Pso & No & Yes & No & $5.774 \pm 2.124$ & $4.400 \pm 1.122$ & $39.765$ & $0.743 \pm 0.125$ & $6.500 \pm 0.174$ & $2.208 \pm 0.157$ \\
\ac{llama} & Circle & Pso & No & Yes & Yes & $7.079 \pm 2.873$ & $6.000 \pm 1.265$ & $47.922$ & $1.183 \pm 0.205$ & $7.200 \pm 0.181$ & $3.060 \pm 0.199$ \\
\ac{llama} & Circle & Pso & Yes & No & No & $5.337 \pm 3.050$ & $8.200 \pm 0.860$ & $60.163$ & $0.950 \pm 0.083$ & $6.960 \pm 0.154$ & $2.750 \pm 0.662$ \\
\ac{llama} & Circle & Pso & Yes & No & Yes & $7.415 \pm 3.334$ & $8.000 \pm 0.632$ & $49.773$ & $0.845 \pm 0.136$ & $7.180 \pm 0.201$ & $2.610 \pm 0.714$ \\
\ac{llama} & Circle & Pso & Yes & Yes & No & $5.647 \pm 2.594$ & $5.000 \pm 1.342$ & $57.373$ & $1.020 \pm 0.151$ & $6.280 \pm 0.229$ & $1.970 \pm 0.692$ \\
\ac{llama} & Circle & Pso & Yes & Yes & Yes & $4.872 \pm 2.358$ & $7.000 \pm 1.643$ & $56.011$ & $0.893 \pm 0.180$ & $6.760 \pm 0.228$ & $2.110 \pm 0.704$ \\
\ac{llama} & Ellipse & Grid & No & No & No & $2.045 \pm 0.264$ & $7.200 \pm 1.463$ & $39.963$ & $1.079 \pm 0.067$ & $6.320 \pm 0.165$ & $2.040 \pm 0.169$ \\
\ac{llama} & Ellipse & Grid & No & No & Yes & $2.224 \pm 0.276$ & $7.200 \pm 0.490$ & $37.106$ & $1.005 \pm 0.032$ & $7.440 \pm 0.190$ & $3.120 \pm 0.187$ \\
\ac{llama} & Ellipse & Grid & No & Yes & No & $1.985 \pm 0.297$ & $6.600 \pm 0.980$ & $53.581$ & $0.929 \pm 0.017$ & $6.520 \pm 0.170$ & $2.460 \pm 0.162$ \\
\ac{llama} & Ellipse & Grid & No & Yes & Yes & $2.036 \pm 0.203$ & $3.600 \pm 1.030$ & $43.973$ & $1.093 \pm 0.203$ & $6.520 \pm 0.149$ & $2.160 \pm 0.165$ \\
\ac{llama} & Ellipse & Grid & Yes & No & No & $2.081 \pm 0.373$ & $6.800 \pm 1.463$ & $46.470$ & $0.959 \pm 0.012$ & $6.740 \pm 0.142$ & $2.330 \pm 0.654$ \\
\ac{llama} & Ellipse & Grid & Yes & No & Yes & $2.648 \pm 0.451$ & $4.000 \pm 1.761$ & $38.017$ & $1.007 \pm 0.117$ & $7.680 \pm 0.175$ & $2.830 \pm 0.673$ \\
\ac{llama} & Ellipse & Grid & Yes & Yes & No & $2.511 \pm 0.219$ & $5.600 \pm 1.568$ & $20.566$ & $0.870 \pm 0.049$ & $6.880 \pm 0.173$ & $1.970 \pm 0.694$ \\
\ac{llama} & Ellipse & Grid & Yes & Yes & Yes & $2.167 \pm 0.186$ & $5.800 \pm 1.356$ & $35.519$ & $0.900 \pm 0.025$ & $6.600 \pm 0.208$ & $1.930 \pm 0.701$ \\
\ac{llama} & Ellipse & Pso & No & No & No & $4.326 \pm 0.631$ & $8.400 \pm 0.678$ & $23.547$ & $0.965 \pm 0.018$ & $6.680 \pm 0.135$ & $2.440 \pm 0.128$ \\
\ac{llama} & Ellipse & Pso & No & No & Yes & $4.364 \pm 0.811$ & $4.800 \pm 1.562$ & $31.403$ & $0.938 \pm 0.028$ & $6.080 \pm 0.171$ & $1.780 \pm 0.144$ \\
\ac{llama} & Ellipse & Pso & No & Yes & No & $4.424 \pm 0.683$ & $7.000 \pm 1.049$ & $28.519$ & $0.995 \pm 0.098$ & $6.480 \pm 0.135$ & $2.060 \pm 0.135$ \\
\ac{llama} & Ellipse & Pso & No & Yes & Yes & $4.350 \pm 0.647$ & $6.000 \pm 1.304$ & $27.886$ & $1.133 \pm 0.097$ & $7.560 \pm 0.154$ & $3.200 \pm 0.148$ \\
\ac{llama} & Ellipse & Pso & Yes & No & No & $4.834 \pm 0.757$ & $4.800 \pm 1.594$ & $26.183$ & $0.692 \pm 0.062$ & $7.143 \pm 0.154$ & $2.603 \pm 0.688$ \\
\ac{llama} & Ellipse & Pso & Yes & No & Yes & $4.633 \pm 0.762$ & $7.800 \pm 0.735$ & $24.938$ & $1.218 \pm 0.215$ & $7.000 \pm 0.174$ & $2.690 \pm 0.675$ \\
\ac{llama} & Ellipse & Pso & Yes & Yes & No & $4.064 \pm 0.557$ & $4.400 \pm 2.088$ & $33.918$ & $0.938 \pm 0.053$ & $6.160 \pm 0.188$ & $1.470 \pm 0.672$ \\
\ac{llama} & Ellipse & Pso & Yes & Yes & Yes & $4.801 \pm 0.761$ & $4.200 \pm 0.917$ & $20.608$ & $0.803 \pm 0.057$ & $6.680 \pm 0.177$ & $2.090 \pm 0.661$ \\
  \hline
\end{tabular}}
\caption{Full results for \ac{llama} on \ac{lb}, circle, and ellipse targets. Columns: \textbf{Opt} = optimiser (Grid search or PSO) \textbf{Best chamf.} = lowest Chamfer distance achieved across all reflection steps (mean $\pm$ standard error); \textbf{Steps} = number of reflection iterations taken for the best result; \textbf{\% Imp.} = percentage reduction in Chamfer distance relative to the initial proposal; \textbf{\% Semantic} = fraction of valid outputs; \textbf{Links} = number of rigid bars in the synthesised mechanism; \textbf{Goal links} = difference between synthesised and specified target link count. Each row aggregates 5 independent runs.}
\label{tab:full_llama_a}
\end{table*}

\begin{table*}
\centering
\resizebox{\columnwidth}{!}{\begin{tabular}{@{}lllllllllllll@{}}
  \hline
  Model & Shape & Opt & Planner & \acs{dr} & \acs{cl} & Best chamf. & Steps & \% Imp. & \% Semantic & Links & Goal links \\
  \hline
\ac{llama} & Line & Grid & No & No & No & $0.966 \pm 0.476$ & $7.200 \pm 0.970$ & $75.784$ & $1.004 \pm 0.105$ & $6.640 \pm 0.166$ & $2.460 \pm 0.170$ \\
\ac{llama} & Line & Grid & No & No & Yes & $0.639 \pm 0.307$ & $8.800 \pm 0.970$ & $88.625$ & $0.908 \pm 0.060$ & $6.920 \pm 0.269$ & $2.840 \pm 0.240$ \\
\ac{llama} & Line & Grid & No & Yes & No & $0.565 \pm 0.148$ & $6.800 \pm 0.970$ & $85.248$ & $0.962 \pm 0.054$ & $7.000 \pm 0.183$ & $2.600 \pm 0.167$ \\
\ac{llama} & Line & Grid & No & Yes & Yes & $0.725 \pm 0.275$ & $6.200 \pm 1.497$ & $88.154$ & $0.982 \pm 0.148$ & $7.360 \pm 0.202$ & $3.080 \pm 0.198$ \\
\ac{llama} & Line & Grid & Yes & No & No & $0.804 \pm 0.522$ & $6.400 \pm 1.435$ & $82.707$ & $1.018 \pm 0.075$ & $6.640 \pm 0.185$ & $2.170 \pm 0.686$ \\
\ac{llama} & Line & Grid & Yes & No & Yes & $0.533 \pm 0.266$ & $7.600 \pm 1.470$ & $77.970$ & $0.863 \pm 0.074$ & $7.200 \pm 0.181$ & $2.610 \pm 0.662$ \\
\ac{llama} & Line & Grid & Yes & Yes & No & $0.606 \pm 0.302$ & $3.600 \pm 1.030$ & $86.842$ & $0.953 \pm 0.118$ & $7.840 \pm 0.220$ & $3.270 \pm 0.716$ \\
\ac{llama} & Line & Grid & Yes & Yes & Yes & $0.581 \pm 0.195$ & $5.200 \pm 0.860$ & $87.537$ & $1.003 \pm 0.088$ & $6.880 \pm 0.207$ & $2.170 \pm 0.696$ \\
\ac{llama} & Line & Pso & No & No & No & $4.364 \pm 1.428$ & $5.400 \pm 0.510$ & $54.639$ & $0.977 \pm 0.108$ & $6.600 \pm 0.154$ & $2.500 \pm 0.162$ \\
\ac{llama} & Line & Pso & No & No & Yes & $3.076 \pm 1.452$ & $7.600 \pm 1.030$ & $64.395$ & $0.891 \pm 0.140$ & $6.840 \pm 0.152$ & $2.640 \pm 0.142$ \\
\ac{llama} & Line & Pso & No & Yes & No & $1.946 \pm 1.270$ & $6.000 \pm 1.483$ & $76.289$ & $0.814 \pm 0.095$ & $6.440 \pm 0.200$ & $2.180 \pm 0.171$ \\
\ac{llama} & Line & Pso & No & Yes & Yes & $2.480 \pm 1.314$ & $7.400 \pm 1.778$ & $64.823$ & $0.743 \pm 0.130$ & $6.208 \pm 0.240$ & $2.271 \pm 0.212$ \\
\ac{llama} & Line & Pso & Yes & No & No & $4.712 \pm 2.004$ & $6.200 \pm 1.393$ & $51.459$ & $0.748 \pm 0.093$ & $6.640 \pm 0.176$ & $2.210 \pm 0.673$ \\
\ac{llama} & Line & Pso & Yes & No & Yes & $3.278 \pm 1.533$ & $7.200 \pm 0.800$ & $68.779$ & $0.843 \pm 0.068$ & $6.680 \pm 0.211$ & $2.310 \pm 0.692$ \\
\ac{llama} & Line & Pso & Yes & Yes & No & $3.767 \pm 1.320$ & $6.200 \pm 1.393$ & $59.639$ & $0.936 \pm 0.094$ & $6.880 \pm 0.191$ & $2.550 \pm 0.685$ \\
\ac{llama} & Line & Pso & Yes & Yes & Yes & $2.252 \pm 1.507$ & $6.600 \pm 0.980$ & $76.992$ & $0.722 \pm 0.076$ & $6.080 \pm 0.206$ & $1.690 \pm 0.705$ \\
\ac{llama} & Naca & Grid & No & No & No & $0.441 \pm 0.101$ & $5.600 \pm 1.208$ & $38.649$ & $1.007 \pm 0.050$ & $7.680 \pm 0.144$ & $3.020 \pm 0.165$ \\
\ac{llama} & Naca & Grid & No & No & Yes & $0.389 \pm 0.139$ & $4.000 \pm 1.000$ & $58.292$ & $1.013 \pm 0.120$ & $7.960 \pm 0.156$ & $3.480 \pm 0.170$ \\
\ac{llama} & Naca & Grid & No & Yes & No & $0.312 \pm 0.066$ & $3.800 \pm 1.158$ & $59.191$ & $0.906 \pm 0.035$ & $7.360 \pm 0.193$ & $3.000 \pm 0.167$ \\
\ac{llama} & Naca & Grid & No & Yes & Yes & $0.397 \pm 0.110$ & $5.400 \pm 1.364$ & $61.467$ & $0.972 \pm 0.148$ & $7.640 \pm 0.204$ & $3.280 \pm 0.192$ \\
\ac{llama} & Naca & Grid & Yes & No & No & $0.397 \pm 0.131$ & $4.600 \pm 1.503$ & $50.315$ & $0.898 \pm 0.027$ & $7.720 \pm 0.190$ & $3.090 \pm 0.723$ \\
\ac{llama} & Naca & Grid & Yes & No & Yes & $0.399 \pm 0.134$ & $6.400 \pm 1.691$ & $60.472$ & $0.933 \pm 0.039$ & $7.840 \pm 0.188$ & $3.090 \pm 0.701$ \\
\ac{llama} & Naca & Grid & Yes & Yes & No & $0.483 \pm 0.107$ & $5.400 \pm 1.208$ & $42.237$ & $0.998 \pm 0.102$ & $8.120 \pm 0.175$ & $3.430 \pm 0.688$ \\
\ac{llama} & Naca & Grid & Yes & Yes & Yes & $0.357 \pm 0.120$ & $5.800 \pm 1.530$ & $51.604$ & $0.942 \pm 0.068$ & $7.320 \pm 0.168$ & $2.590 \pm 0.692$ \\
\ac{llama} & Naca & Pso & No & No & No & $0.787 \pm 0.359$ & $6.800 \pm 1.594$ & $44.268$ & $0.867 \pm 0.036$ & $7.120 \pm 0.163$ & $2.940 \pm 0.175$ \\
\ac{llama} & Naca & Pso & No & No & Yes & $0.732 \pm 0.329$ & $7.000 \pm 1.000$ & $44.496$ & $0.828 \pm 0.024$ & $7.200 \pm 0.190$ & $2.880 \pm 0.182$ \\
\ac{llama} & Naca & Pso & No & Yes & No & $0.466 \pm 0.182$ & $4.800 \pm 1.356$ & $46.681$ & $0.896 \pm 0.127$ & $7.660 \pm 0.253$ & $3.213 \pm 0.227$ \\
\ac{llama} & Naca & Pso & No & Yes & Yes & $0.633 \pm 0.254$ & $4.800 \pm 0.860$ & $44.777$ & $1.125 \pm 0.130$ & $7.640 \pm 0.195$ & $3.440 \pm 0.204$ \\
\ac{llama} & Naca & Pso & Yes & No & No & $0.559 \pm 0.205$ & $4.000 \pm 1.342$ & $32.515$ & $1.078 \pm 0.094$ & $7.680 \pm 0.165$ & $3.250 \pm 0.679$ \\
\ac{llama} & Naca & Pso & Yes & No & Yes & $0.638 \pm 0.304$ & $4.800 \pm 1.497$ & $45.338$ & $1.089 \pm 0.175$ & $8.000 \pm 0.171$ & $3.270 \pm 0.667$ \\
\ac{llama} & Naca & Pso & Yes & Yes & No & $0.386 \pm 0.165$ & $6.800 \pm 1.530$ & $63.629$ & $1.031 \pm 0.207$ & $7.600 \pm 0.236$ & $2.930 \pm 0.707$ \\
\ac{llama} & Naca & Pso & Yes & Yes & Yes & $0.608 \pm 0.268$ & $6.200 \pm 1.200$ & $45.657$ & $0.911 \pm 0.143$ & $6.880 \pm 0.207$ & $2.150 \pm 0.683$ \\
\ac{llama} & Parabola & Grid & No & No & No & $688.191 \pm 31.323$ & $7.000 \pm 1.342$ & $21.954$ & $0.997 \pm 0.098$ & $7.120 \pm 0.199$ & $2.720 \pm 0.192$ \\
\ac{llama} & Parabola & Grid & No & No & Yes & $688.919 \pm 34.117$ & $7.200 \pm 0.800$ & $19.430$ & $1.025 \pm 0.152$ & $7.160 \pm 0.152$ & $2.920 \pm 0.151$ \\
\ac{llama} & Parabola & Grid & No & Yes & No & $797.121 \pm 47.780$ & $7.400 \pm 1.691$ & $6.953$ & $0.955 \pm 0.058$ & $7.200 \pm 0.171$ & $2.940 \pm 0.175$ \\
\ac{llama} & Parabola & Grid & No & Yes & Yes & $752.773 \pm 28.310$ & $7.600 \pm 0.510$ & $15.314$ & $0.905 \pm 0.038$ & $6.480 \pm 0.203$ & $2.460 \pm 0.174$ \\
\ac{llama} & Parabola & Grid & Yes & No & No & $696.402 \pm 76.192$ & $7.800 \pm 1.020$ & $20.990$ & $1.080 \pm 0.125$ & $6.720 \pm 0.159$ & $2.350 \pm 0.659$ \\
\ac{llama} & Parabola & Grid & Yes & No & Yes & $728.409 \pm 25.214$ & $7.200 \pm 0.860$ & $17.713$ & $0.882 \pm 0.032$ & $7.200 \pm 0.181$ & $2.530 \pm 0.657$ \\
\ac{llama} & Parabola & Grid & Yes & Yes & No & $799.059 \pm 52.367$ & $5.400 \pm 1.288$ & $10.327$ & $0.862 \pm 0.034$ & $6.640 \pm 0.166$ & $2.050 \pm 0.667$ \\
\ac{llama} & Parabola & Grid & Yes & Yes & Yes & $792.749 \pm 42.745$ & $6.800 \pm 1.428$ & $10.604$ & $0.981 \pm 0.136$ & $6.880 \pm 0.215$ & $2.350 \pm 0.714$ \\
\ac{llama} & Parabola & Pso & No & No & No & $887.541 \pm 25.010$ & $5.600 \pm 0.510$ & $0.823$ & $1.007 \pm 0.137$ & $6.800 \pm 0.151$ & $2.620 \pm 0.143$ \\
\ac{llama} & Parabola & Pso & No & No & Yes & $885.810 \pm 23.119$ & $7.000 \pm 0.707$ & $1.029$ & $0.837 \pm 0.050$ & $5.800 \pm 0.230$ & $1.960 \pm 0.169$ \\
\ac{llama} & Parabola & Pso & No & Yes & No & $889.441 \pm 24.337$ & $7.200 \pm 1.068$ & $0.628$ & $0.796 \pm 0.046$ & $6.960 \pm 0.143$ & $2.580 \pm 0.151$ \\
\ac{llama} & Parabola & Pso & No & Yes & Yes & $890.400 \pm 25.080$ & $6.000 \pm 1.612$ & $0.394$ & $0.865 \pm 0.101$ & $7.120 \pm 0.182$ & $2.700 \pm 0.184$ \\
\ac{llama} & Parabola & Pso & Yes & No & No & $890.007 \pm 24.088$ & $5.200 \pm 0.970$ & $0.470$ & $0.906 \pm 0.134$ & $7.320 \pm 0.186$ & $2.850 \pm 0.694$ \\
\ac{llama} & Parabola & Pso & Yes & No & Yes & $889.856 \pm 24.576$ & $6.200 \pm 1.463$ & $0.523$ & $0.851 \pm 0.164$ & $6.762 \pm 0.166$ & $2.083 \pm 0.662$ \\
\ac{llama} & Parabola & Pso & Yes & Yes & No & $889.138 \pm 24.168$ & $4.000 \pm 1.378$ & $0.739$ & $0.785 \pm 0.050$ & $6.520 \pm 0.212$ & $1.850 \pm 0.712$ \\
\ac{llama} & Parabola & Pso & Yes & Yes & Yes & $888.016 \pm 23.356$ & $7.600 \pm 1.364$ & $0.837$ & $0.904 \pm 0.036$ & $6.120 \pm 0.231$ & $1.810 \pm 0.707$ \\
  \hline
\end{tabular}}
\caption{Full results for \ac{llama} on straight-line, NACA airfoil, and parabola targets. Column definitions are the same as in Table~\ref{tab:full_llama_a}. Each row aggregates 5 independent runs.}
\label{tab:full_llama_b}
\end{table*}

\begin{table*}
\centering
\resizebox{\columnwidth}{!}{\begin{tabular}{@{}lllllllllllll@{}}
  \hline
  Model & Shape & Opt & Planner & \acs{dr} & \acs{cl} & Best chamf. & Steps & \% Imp. & \% Semantic & Links & Goal links \\
  \hline
\ac{qwen-moe} & \ac{lb} & Grid & No & No & No & $7.290 \pm 0.471$ & $3.400 \pm 0.980$ & $0.000$ & $0.357 \pm 0.082$ & $6.000 \pm 0.051$ & $1.554 \pm 0.105$ \\
\ac{qwen-moe} & \ac{lb} & Grid & No & No & Yes & $6.859 \pm 0.550$ & $7.200 \pm 1.241$ & $8.688$ & $0.495 \pm 0.167$ & $6.046 \pm 0.060$ & $1.462 \pm 0.112$ \\
\ac{qwen-moe} & \ac{lb} & Grid & No & Yes & No & $7.135 \pm 0.600$ & $4.800 \pm 1.020$ & $3.805$ & $0.260 \pm 0.077$ & $5.934 \pm 0.117$ & $1.787 \pm 0.117$ \\
\ac{qwen-moe} & \ac{lb} & Grid & No & Yes & Yes & $4.861 \pm 0.095$ & $6.800 \pm 1.068$ & $37.703$ & $0.675 \pm 0.179$ & $6.342 \pm 0.126$ & $1.921 \pm 0.140$ \\
\ac{qwen-moe} & \ac{lb} & Grid & Yes & No & No & $6.953 \pm 0.495$ & $6.600 \pm 1.327$ & $20.610$ & $0.458 \pm 0.162$ & $5.794 \pm 0.076$ & $1.394 \pm 0.590$ \\
\ac{qwen-moe} & \ac{lb} & Grid & Yes & No & Yes & $7.045 \pm 0.268$ & $6.000 \pm 1.581$ & $7.712$ & $0.498 \pm 0.276$ & $5.682 \pm 0.121$ & $1.098 \pm 0.615$ \\
\ac{qwen-moe} & \ac{lb} & Grid & Yes & Yes & No & $7.886 \pm 0.483$ & $4.800 \pm 1.562$ & $1.765$ & $0.395 \pm 0.125$ & $6.095 \pm 0.084$ & $1.512 \pm 0.615$ \\
\ac{qwen-moe} & \ac{lb} & Grid & Yes & Yes & Yes & $6.918 \pm 0.794$ & $5.000 \pm 1.183$ & $4.605$ & $0.435 \pm 0.148$ & $5.959 \pm 0.041$ & $1.271 \pm 0.583$ \\
\ac{qwen-moe} & \ac{lb} & Pso & No & No & No & $9.649 \pm 0.379$ & $6.200 \pm 1.241$ & $1.244$ & $0.204 \pm 0.039$ & $5.929 \pm 0.071$ & $1.768 \pm 0.057$ \\
\ac{qwen-moe} & \ac{lb} & Pso & No & No & Yes & $9.439 \pm 0.540$ & $3.600 \pm 1.208$ & $1.759$ & $0.412 \pm 0.171$ & $5.912 \pm 0.045$ & $1.579 \pm 0.079$ \\
\ac{qwen-moe} & \ac{lb} & Pso & No & Yes & No & $9.682 \pm 0.456$ & $3.400 \pm 0.812$ & $7.802$ & $0.327 \pm 0.054$ & $6.016 \pm 0.060$ & $1.705 \pm 0.089$ \\
\ac{qwen-moe} & \ac{lb} & Pso & No & Yes & Yes & $9.553 \pm 0.523$ & $5.400 \pm 1.327$ & $0.831$ & $0.296 \pm 0.076$ & $6.127 \pm 0.062$ & $1.921 \pm 0.091$ \\
\ac{qwen-moe} & \ac{lb} & Pso & Yes & No & No & $11.424 \pm 2.220$ & $6.000 \pm 1.140$ & $9.779$ & $0.333 \pm 0.087$ & $6.018 \pm 0.054$ & $1.682 \pm 0.594$ \\
\ac{qwen-moe} & \ac{lb} & Pso & Yes & No & Yes & $9.766 \pm 0.464$ & $5.800 \pm 1.356$ & $1.143$ & $0.688 \pm 0.195$ & $6.000 \pm 0.000$ & $1.368 \pm 0.573$ \\
\ac{qwen-moe} & \ac{lb} & Pso & Yes & Yes & No & $9.516 \pm 0.437$ & $5.000 \pm 1.304$ & $0.000$ & $0.153 \pm 0.026$ & $5.909 \pm 0.075$ & $1.368 \pm 0.599$ \\
\ac{qwen-moe} & \ac{lb} & Pso & Yes & Yes & Yes & $9.559 \pm 0.425$ & $5.200 \pm 1.158$ & $0.000$ & $0.162 \pm 0.044$ & $5.982 \pm 0.084$ & $1.477 \pm 0.588$ \\
\ac{qwen-moe} & Circle & Grid & No & No & No & $1.808 \pm 0.654$ & $5.400 \pm 0.678$ & $4.215$ & $0.335 \pm 0.091$ & $6.233 \pm 0.117$ & $1.717 \pm 0.119$ \\
\ac{qwen-moe} & Circle & Grid & No & No & Yes & $1.258 \pm 0.537$ & $4.200 \pm 1.393$ & $6.996$ & $0.367 \pm 0.128$ & $6.050 \pm 0.055$ & $1.500 \pm 0.110$ \\
\ac{qwen-moe} & Circle & Grid & No & Yes & No & $5.497 \pm 2.700$ & $4.800 \pm 1.625$ & $26.771$ & $0.465 \pm 0.156$ & $5.905 \pm 0.071$ & $1.476 \pm 0.084$ \\
\ac{qwen-moe} & Circle & Grid & No & Yes & Yes & $2.013 \pm 0.652$ & $6.200 \pm 2.131$ & $16.543$ & $0.300 \pm 0.057$ & $6.000 \pm 0.043$ & $1.456 \pm 0.090$ \\
\ac{qwen-moe} & Circle & Grid & Yes & No & No & $3.825 \pm 1.389$ & $4.600 \pm 1.208$ & $2.862$ & $0.204 \pm 0.051$ & $6.196 \pm 0.123$ & $1.700 \pm 0.642$ \\
\ac{qwen-moe} & Circle & Grid & Yes & No & Yes & $1.856 \pm 0.550$ & $4.400 \pm 1.913$ & $0.000$ & $0.187 \pm 0.031$ & $6.036 \pm 0.063$ & $1.241 \pm 0.616$ \\
\ac{qwen-moe} & Circle & Grid & Yes & Yes & No & $2.192 \pm 0.478$ & $4.000 \pm 1.673$ & $17.379$ & $0.297 \pm 0.040$ & $5.929 \pm 0.071$ & $1.089 \pm 0.615$ \\
\ac{qwen-moe} & Circle & Grid & Yes & Yes & Yes & $2.918 \pm 1.672$ & $6.800 \pm 1.393$ & $41.734$ & $0.509 \pm 0.190$ & $6.063 \pm 0.101$ & $1.337 \pm 0.631$ \\
\ac{qwen-moe} & Circle & Pso & No & No & No & $3.658 \pm 1.028$ & $6.200 \pm 1.497$ & $21.006$ & $0.683 \pm 0.191$ & $5.885 \pm 0.094$ & $1.574 \pm 0.100$ \\
\ac{qwen-moe} & Circle & Pso & No & No & Yes & $5.498 \pm 3.262$ & $6.400 \pm 1.833$ & $30.854$ & $0.723 \pm 0.182$ & $5.972 \pm 0.078$ & $1.549 \pm 0.102$ \\
\ac{qwen-moe} & Circle & Pso & No & Yes & No & $4.949 \pm 2.281$ & $7.600 \pm 1.364$ & $47.531$ & $0.970 \pm 0.157$ & $5.953 \pm 0.033$ & $1.506 \pm 0.078$ \\
\ac{qwen-moe} & Circle & Pso & No & Yes & Yes & $8.047 \pm 4.067$ & $6.600 \pm 1.749$ & $37.524$ & $0.517 \pm 0.164$ & $5.930 \pm 0.058$ & $1.521 \pm 0.082$ \\
\ac{qwen-moe} & Circle & Pso & Yes & No & No & $3.366 \pm 0.869$ & $6.000 \pm 1.414$ & $37.633$ & $0.746 \pm 0.252$ & $5.971 \pm 0.029$ & $1.597 \pm 0.575$ \\
\ac{qwen-moe} & Circle & Pso & Yes & No & Yes & $2.894 \pm 1.345$ & $6.600 \pm 1.691$ & $31.505$ & $0.542 \pm 0.143$ & $5.791 \pm 0.078$ & $1.347 \pm 0.588$ \\
\ac{qwen-moe} & Circle & Pso & Yes & Yes & No & $6.090 \pm 3.232$ & $4.200 \pm 1.393$ & $7.242$ & $0.437 \pm 0.099$ & $6.000 \pm 0.000$ & $1.267 \pm 0.593$ \\
\ac{qwen-moe} & Circle & Pso & Yes & Yes & Yes & $3.444 \pm 1.260$ & $5.000 \pm 1.581$ & $14.536$ & $0.363 \pm 0.067$ & $5.952 \pm 0.107$ & $1.337 \pm 0.617$ \\
\ac{qwen-moe} & Ellipse & Grid & No & No & No & $3.215 \pm 0.391$ & $4.000 \pm 1.761$ & $5.687$ & $0.437 \pm 0.063$ & $5.902 \pm 0.051$ & $1.344 \pm 0.099$ \\
\ac{qwen-moe} & Ellipse & Grid & No & No & Yes & $4.469 \pm 1.276$ & $4.800 \pm 0.860$ & $12.395$ & $0.219 \pm 0.034$ & $6.000 \pm 0.051$ & $1.536 \pm 0.105$ \\
\ac{qwen-moe} & Ellipse & Grid & No & Yes & No & $3.799 \pm 0.552$ & $2.800 \pm 0.735$ & $10.434$ & $0.380 \pm 0.158$ & $5.931 \pm 0.048$ & $1.466 \pm 0.089$ \\
\ac{qwen-moe} & Ellipse & Grid & No & Yes & Yes & $2.709 \pm 0.368$ & $6.600 \pm 1.400$ & $19.425$ & $0.661 \pm 0.136$ & $6.076 \pm 0.069$ & $1.621 \pm 0.112$ \\
\ac{qwen-moe} & Ellipse & Grid & Yes & No & No & $3.592 \pm 0.451$ & $5.200 \pm 1.158$ & $7.466$ & $0.189 \pm 0.040$ & $5.931 \pm 0.073$ & $1.381 \pm 0.613$ \\
\ac{qwen-moe} & Ellipse & Grid & Yes & No & Yes & $3.197 \pm 0.586$ & $4.200 \pm 0.970$ & $9.063$ & $0.510 \pm 0.156$ & $5.918 \pm 0.051$ & $1.229 \pm 0.587$ \\
\ac{qwen-moe} & Ellipse & Grid & Yes & Yes & No & $3.031 \pm 0.835$ & $6.600 \pm 0.678$ & $38.263$ & $0.405 \pm 0.124$ & $5.917 \pm 0.055$ & $1.236 \pm 0.593$ \\
\ac{qwen-moe} & Ellipse & Grid & Yes & Yes & Yes & $3.046 \pm 0.640$ & $5.400 \pm 1.600$ & $7.528$ & $0.346 \pm 0.094$ & $6.000 \pm 0.000$ & $1.219 \pm 0.597$ \\
\ac{qwen-moe} & Ellipse & Pso & No & No & No & $6.456 \pm 1.330$ & $5.000 \pm 1.414$ & $11.438$ & $0.730 \pm 0.112$ & $6.000 \pm 0.040$ & $1.681 \pm 0.074$ \\
\ac{qwen-moe} & Ellipse & Pso & No & No & Yes & $4.988 \pm 0.643$ & $5.600 \pm 1.435$ & $17.625$ & $0.714 \pm 0.102$ & $5.813 \pm 0.092$ & $1.427 \pm 0.101$ \\
\ac{qwen-moe} & Ellipse & Pso & No & Yes & No & $6.930 \pm 1.319$ & $5.200 \pm 1.655$ & $2.226$ & $0.470 \pm 0.153$ & $6.018 \pm 0.018$ & $1.860 \pm 0.053$ \\
\ac{qwen-moe} & Ellipse & Pso & No & Yes & Yes & $5.483 \pm 0.906$ & $4.000 \pm 1.000$ & $0.000$ & $0.420 \pm 0.161$ & $5.842 \pm 0.093$ & $1.579 \pm 0.112$ \\
\ac{qwen-moe} & Ellipse & Pso & Yes & No & No & $5.210 \pm 0.861$ & $4.000 \pm 0.949$ & $7.126$ & $0.523 \pm 0.191$ & $5.848 \pm 0.124$ & $1.617 \pm 0.611$ \\
\ac{qwen-moe} & Ellipse & Pso & Yes & No & Yes & $5.935 \pm 1.284$ & $5.200 \pm 1.356$ & $11.620$ & $0.445 \pm 0.189$ & $5.847 \pm 0.078$ & $1.264 \pm 0.595$ \\
\ac{qwen-moe} & Ellipse & Pso & Yes & Yes & No & $5.408 \pm 0.911$ & $6.000 \pm 0.837$ & $6.555$ & $0.687 \pm 0.274$ & $6.000 \pm 0.037$ & $1.303 \pm 0.578$ \\
\ac{qwen-moe} & Ellipse & Pso & Yes & Yes & Yes & $5.395 \pm 0.832$ & $3.600 \pm 1.600$ & $10.839$ & $0.398 \pm 0.108$ & $5.971 \pm 0.029$ & $1.431 \pm 0.567$ \\
  \hline
\end{tabular}}
\caption{Full results for \ac{qwen-moe} on \acl{lb}, circle, and ellipse targets. Column definitions follow Table~\ref{tab:full_llama_a}. Each row aggregates 5 independent runs.}
\label{tab:full_qwenmoe_a}
\end{table*}

\begin{table*}
\centering
\resizebox{\columnwidth}{!}{\begin{tabular}{@{}lllllllllllll@{}}
  \hline
  Model & Shape & Opt & Planner & \acs{dr} & \acs{cl} & Best chamf. & Steps & \% Imp. & \% Semantic & Links & Goal links \\
  \hline
\ac{qwen-moe} & Line & Grid & No & No & No & $2.884 \pm 0.952$ & $5.000 \pm 1.517$ & $16.588$ & $0.302 \pm 0.066$ & $5.982 \pm 0.018$ & $1.404 \pm 0.090$ \\
\ac{qwen-moe} & Line & Grid & No & No & Yes & $2.435 \pm 1.014$ & $5.800 \pm 1.594$ & $20.545$ & $0.271 \pm 0.070$ & $6.079 \pm 0.057$ & $1.746 \pm 0.088$ \\
\ac{qwen-moe} & Line & Grid & No & Yes & No & $1.956 \pm 0.756$ & $7.000 \pm 1.378$ & $62.388$ & $0.499 \pm 0.091$ & $5.681 \pm 0.129$ & $1.514 \pm 0.095$ \\
\ac{qwen-moe} & Line & Grid & No & Yes & Yes & $2.338 \pm 0.965$ & $6.600 \pm 1.122$ & $19.364$ & $0.401 \pm 0.121$ & $5.935 \pm 0.079$ & $1.597 \pm 0.107$ \\
\ac{qwen-moe} & Line & Grid & Yes & No & No & $2.191 \pm 0.463$ & $5.200 \pm 1.428$ & $0.000$ & $0.254 \pm 0.026$ & $6.000 \pm 0.051$ & $1.646 \pm 0.584$ \\
\ac{qwen-moe} & Line & Grid & Yes & No & Yes & $1.516 \pm 0.446$ & $7.000 \pm 1.483$ & $13.224$ & $0.312 \pm 0.088$ & $6.016 \pm 0.072$ & $1.439 \pm 0.606$ \\
\ac{qwen-moe} & Line & Grid & Yes & Yes & No & $2.173 \pm 0.804$ & $6.400 \pm 1.470$ & $17.818$ & $0.182 \pm 0.046$ & $6.017 \pm 0.077$ & $1.500 \pm 0.584$ \\
\ac{qwen-moe} & Line & Grid & Yes & Yes & Yes & $1.985 \pm 1.026$ & $4.400 \pm 1.327$ & $22.324$ & $0.440 \pm 0.102$ & $6.206 \pm 0.104$ & $1.588 \pm 0.618$ \\
\ac{qwen-moe} & Line & Pso & No & No & No & $5.236 \pm 1.588$ & $3.400 \pm 1.661$ & $19.718$ & $0.533 \pm 0.122$ & $5.831 \pm 0.097$ & $1.746 \pm 0.057$ \\
\ac{qwen-moe} & Line & Pso & No & No & Yes & $5.526 \pm 2.195$ & $5.000 \pm 1.643$ & $40.439$ & $0.450 \pm 0.104$ & $6.075 \pm 0.039$ & $1.672 \pm 0.089$ \\
\ac{qwen-moe} & Line & Pso & No & Yes & No & $2.502 \pm 0.743$ & $8.000 \pm 1.000$ & $30.642$ & $0.745 \pm 0.090$ & $5.882 \pm 0.065$ & $1.539 \pm 0.087$ \\
\ac{qwen-moe} & Line & Pso & No & Yes & Yes & $4.822 \pm 1.372$ & $4.000 \pm 0.949$ & $27.008$ & $0.504 \pm 0.139$ & $5.958 \pm 0.042$ & $1.493 \pm 0.080$ \\
\ac{qwen-moe} & Line & Pso & Yes & No & No & $5.088 \pm 2.062$ & $4.800 \pm 1.200$ & $37.829$ & $0.718 \pm 0.331$ & $5.973 \pm 0.027$ & $1.580 \pm 0.579$ \\
\ac{qwen-moe} & Line & Pso & Yes & No & Yes & $3.720 \pm 1.320$ & $4.800 \pm 1.356$ & $31.619$ & $0.534 \pm 0.075$ & $5.468 \pm 0.121$ & $0.851 \pm 0.604$ \\
\ac{qwen-moe} & Line & Pso & Yes & Yes & No & $5.098 \pm 1.628$ & $4.200 \pm 1.393$ & $22.057$ & $0.441 \pm 0.103$ & $6.114 \pm 0.059$ & $1.593 \pm 0.590$ \\
\ac{qwen-moe} & Line & Pso & Yes & Yes & Yes & $5.330 \pm 2.232$ & $6.000 \pm 1.265$ & $26.217$ & $0.637 \pm 0.174$ & $5.935 \pm 0.047$ & $1.269 \pm 0.575$ \\
\ac{qwen-moe} & Naca & Grid & No & No & No & $0.738 \pm 0.312$ & $4.000 \pm 1.265$ & $7.437$ & $0.478 \pm 0.160$ & $6.094 \pm 0.069$ & $1.531 \pm 0.111$ \\
\ac{qwen-moe} & Naca & Grid & No & No & Yes & $0.294 \pm 0.049$ & $4.000 \pm 0.548$ & $18.170$ & $0.417 \pm 0.102$ & $6.032 \pm 0.055$ & $1.667 \pm 0.082$ \\
\ac{qwen-moe} & Naca & Grid & No & Yes & No & $0.816 \pm 0.278$ & $6.400 \pm 1.288$ & $7.359$ & $0.406 \pm 0.163$ & $6.133 \pm 0.090$ & $1.717 \pm 0.109$ \\
\ac{qwen-moe} & Naca & Grid & No & Yes & Yes & $0.450 \pm 0.126$ & $7.800 \pm 1.497$ & $21.612$ & $0.545 \pm 0.174$ & $6.000 \pm 0.045$ & $1.562 \pm 0.086$ \\
\ac{qwen-moe} & Naca & Grid & Yes & No & No & $0.497 \pm 0.175$ & $3.800 \pm 1.200$ & $0.000$ & $0.288 \pm 0.073$ & $6.086 \pm 0.099$ & $1.605 \pm 0.624$ \\
\ac{qwen-moe} & Naca & Grid & Yes & No & Yes & $0.354 \pm 0.104$ & $5.200 \pm 0.800$ & $15.108$ & $0.378 \pm 0.057$ & $6.209 \pm 0.099$ & $1.586 \pm 0.616$ \\
\ac{qwen-moe} & Naca & Grid & Yes & Yes & No & $0.764 \pm 0.313$ & $7.600 \pm 1.030$ & $11.943$ & $0.360 \pm 0.123$ & $6.200 \pm 0.106$ & $1.504 \pm 0.616$ \\
\ac{qwen-moe} & Naca & Grid & Yes & Yes & Yes & $0.469 \pm 0.096$ & $2.600 \pm 0.927$ & $0.000$ & $0.340 \pm 0.088$ & $5.930 \pm 0.049$ & $1.627 \pm 0.556$ \\
\ac{qwen-moe} & Naca & Pso & No & No & No & $0.908 \pm 0.416$ & $6.400 \pm 0.510$ & $11.451$ & $0.633 \pm 0.066$ & $6.083 \pm 0.062$ & $1.694 \pm 0.088$ \\
\ac{qwen-moe} & Naca & Pso & No & No & Yes & $1.033 \pm 0.513$ & $3.800 \pm 0.970$ & $16.175$ & $0.639 \pm 0.132$ & $6.045 \pm 0.033$ & $1.716 \pm 0.082$ \\
\ac{qwen-moe} & Naca & Pso & No & Yes & No & $0.923 \pm 0.407$ & $6.000 \pm 1.612$ & $9.208$ & $0.776 \pm 0.102$ & $6.000 \pm 0.060$ & $1.836 \pm 0.084$ \\
\ac{qwen-moe} & Naca & Pso & No & Yes & Yes & $1.199 \pm 0.570$ & $2.800 \pm 0.917$ & $1.017$ & $0.407 \pm 0.150$ & $6.018 \pm 0.053$ & $1.719 \pm 0.086$ \\
\ac{qwen-moe} & Naca & Pso & Yes & No & No & $0.814 \pm 0.377$ & $6.800 \pm 1.393$ & $23.421$ & $0.596 \pm 0.041$ & $6.195 \pm 0.076$ & $1.729 \pm 0.599$ \\
\ac{qwen-moe} & Naca & Pso & Yes & No & Yes & $1.056 \pm 0.455$ & $3.600 \pm 1.030$ & $5.512$ & $0.400 \pm 0.049$ & $5.800 \pm 0.091$ & $1.317 \pm 0.584$ \\
\ac{qwen-moe} & Naca & Pso & Yes & Yes & No & $0.831 \pm 0.400$ & $5.000 \pm 1.378$ & $15.199$ & $0.522 \pm 0.146$ & $6.014 \pm 0.057$ & $1.449 \pm 0.582$ \\
\ac{qwen-moe} & Naca & Pso & Yes & Yes & Yes & $0.700 \pm 0.273$ & $6.400 \pm 1.288$ & $22.306$ & $0.683 \pm 0.191$ & $6.110 \pm 0.046$ & $1.558 \pm 0.575$ \\
\ac{qwen-moe} & Parabola & Grid & No & No & No & $881.244 \pm 25.417$ & $5.000 \pm 1.924$ & $0.677$ & $0.432 \pm 0.111$ & $5.965 \pm 0.025$ & $1.509 \pm 0.091$ \\
\ac{qwen-moe} & Parabola & Grid & No & No & Yes & $722.703 \pm 83.052$ & $6.400 \pm 1.288$ & $15.111$ & $0.557 \pm 0.166$ & $6.041 \pm 0.090$ & $1.671 \pm 0.096$ \\
\ac{qwen-moe} & Parabola & Grid & No & Yes & No & $830.956 \pm 55.665$ & $2.600 \pm 0.927$ & $4.635$ & $0.594 \pm 0.136$ & $6.172 \pm 0.082$ & $1.719 \pm 0.090$ \\
\ac{qwen-moe} & Parabola & Grid & No & Yes & Yes & $790.106 \pm 63.040$ & $4.000 \pm 1.673$ & $10.044$ & $0.767 \pm 0.235$ & $6.307 \pm 0.085$ & $1.747 \pm 0.122$ \\
\ac{qwen-moe} & Parabola & Grid & Yes & No & No & $772.575 \pm 57.140$ & $3.600 \pm 1.249$ & $9.106$ & $0.699 \pm 0.280$ & $5.887 \pm 0.055$ & $1.527 \pm 0.579$ \\
\ac{qwen-moe} & Parabola & Grid & Yes & No & Yes & $702.749 \pm 132.429$ & $5.600 \pm 1.913$ & $3.616$ & $0.594 \pm 0.118$ & $5.761 \pm 0.134$ & $1.412 \pm 0.598$ \\
\ac{qwen-moe} & Parabola & Grid & Yes & Yes & No & $821.719 \pm 65.612$ & $4.400 \pm 1.691$ & $7.400$ & $0.515 \pm 0.107$ & $5.938 \pm 0.099$ & $1.234 \pm 0.609$ \\
\ac{qwen-moe} & Parabola & Grid & Yes & Yes & Yes & $770.787 \pm 37.359$ & $5.800 \pm 1.356$ & $9.732$ & $0.611 \pm 0.126$ & $6.013 \pm 0.080$ & $1.568 \pm 0.582$ \\
\ac{qwen-moe} & Parabola & Pso & No & No & No & $894.803 \pm 24.222$ & $3.800 \pm 1.114$ & $0.000$ & $0.262 \pm 0.063$ & $6.073 \pm 0.073$ & $1.873 \pm 0.097$ \\
\ac{qwen-moe} & Parabola & Pso & No & No & Yes & $899.069 \pm 24.509$ & $5.000 \pm 1.095$ & $0.036$ & $0.237 \pm 0.045$ & $5.982 \pm 0.040$ & $1.446 \pm 0.105$ \\
\ac{qwen-moe} & Parabola & Pso & No & Yes & No & $897.507 \pm 24.587$ & $5.200 \pm 1.625$ & $0.180$ & $0.221 \pm 0.035$ & $5.931 \pm 0.069$ & $1.603 \pm 0.081$ \\
\ac{qwen-moe} & Parabola & Pso & No & Yes & Yes & $894.902 \pm 24.116$ & $4.200 \pm 0.970$ & $0.000$ & $0.399 \pm 0.156$ & $5.911 \pm 0.064$ & $1.571 \pm 0.084$ \\
\ac{qwen-moe} & Parabola & Pso & Yes & No & No & $892.149 \pm 24.567$ & $3.400 \pm 1.288$ & $0.000$ & $0.340 \pm 0.066$ & $5.855 \pm 0.071$ & $1.386 \pm 0.609$ \\
\ac{qwen-moe} & Parabola & Pso & Yes & No & Yes & $892.757 \pm 24.417$ & $6.000 \pm 1.225$ & $0.204$ & $0.269 \pm 0.068$ & $5.895 \pm 0.060$ & $1.329 \pm 0.590$ \\
\ac{qwen-moe} & Parabola & Pso & Yes & Yes & No & $895.986 \pm 31.681$ & $6.000 \pm 1.291$ & $0.019$ & $0.217 \pm 0.057$ & $5.764 \pm 0.086$ & $1.277 \pm 0.589$ \\
\ac{qwen-moe} & Parabola & Pso & Yes & Yes & Yes & $872.936 \pm 31.964$ & $3.600 \pm 1.030$ & $0.000$ & $0.215 \pm 0.055$ & $5.895 \pm 0.060$ & $1.189 \pm 0.594$ \\
  \hline
\end{tabular}}
\caption{Full results for \ac{qwen-moe} on straight-line, NACA airfoil, and parabola targets. Column definitions follow Table~\ref{tab:full_llama_a}. Each row aggregates 5 independent runs.}
\label{tab:full_qwenmoe_b}
\end{table*}

\begin{table*}
\centering
\resizebox{\columnwidth}{!}{\begin{tabular}{@{}lllllllllllll@{}}
  \hline
  Model & Shape & Opt & Planner & \acs{dr} & \acs{cl} & Best chamf. & Steps & \% Imp. & \% Semantic & Links & Goal links \\
  \hline
\ac{qwen} & \ac{lb} & Grid & No & No & No & $6.546 \pm 0.299$ & $13.000 \pm 1.517$ & $24.042$ & $0.795 \pm 0.031$ & $7.020 \pm 0.123$ & $2.140 \pm 0.130$ \\
\ac{qwen} & \ac{lb} & Grid & No & No & Yes & $7.411 \pm 0.541$ & $10.800 \pm 1.530$ & $10.626$ & $0.810 \pm 0.099$ & $6.632 \pm 0.144$ & $1.737 \pm 0.136$ \\
\ac{qwen} & \ac{lb} & Grid & No & Yes & No & $6.547 \pm 0.593$ & $7.800 \pm 1.068$ & $21.267$ & $1.145 \pm 0.147$ & $7.386 \pm 0.145$ & $2.571 \pm 0.143$ \\
\ac{qwen} & \ac{lb} & Grid & No & Yes & Yes & $6.475 \pm 0.610$ & $11.600 \pm 2.379$ & $7.758$ & $0.742 \pm 0.122$ & $7.484 \pm 0.186$ & $2.516 \pm 0.183$ \\
\ac{qwen} & \ac{lb} & Grid & Yes & No & No & $7.214 \pm 0.656$ & $9.200 \pm 2.332$ & $18.821$ & $0.670 \pm 0.022$ & $7.620 \pm 0.225$ & $2.480 \pm 0.674$ \\
\ac{qwen} & \ac{lb} & Grid & Yes & No & Yes & $7.676 \pm 0.474$ & $15.000 \pm 0.894$ & $10.150$ & $0.885 \pm 0.181$ & $6.870 \pm 0.152$ & $2.250 \pm 0.647$ \\
\ac{qwen} & \ac{lb} & Grid & Yes & Yes & No & $7.436 \pm 0.480$ & $11.200 \pm 2.596$ & $8.809$ & $0.907 \pm 0.127$ & $7.300 \pm 0.134$ & $2.179 \pm 0.639$ \\
\ac{qwen} & \ac{lb} & Grid & Yes & Yes & Yes & $5.111 \pm 0.808$ & $14.800 \pm 2.059$ & $39.324$ & $0.998 \pm 0.128$ & $7.700 \pm 0.177$ & $2.475 \pm 0.650$ \\
\ac{qwen} & \ac{lb} & Pso & No & No & No & $8.610 \pm 0.601$ & $11.800 \pm 4.116$ & $8.467$ & $1.131 \pm 0.259$ & $7.443 \pm 0.196$ & $2.113 \pm 0.142$ \\
\ac{qwen} & \ac{lb} & Pso & No & No & Yes & $8.340 \pm 0.432$ & $15.600 \pm 2.619$ & $9.994$ & $1.060 \pm 0.135$ & $7.170 \pm 0.126$ & $2.510 \pm 0.135$ \\
\ac{qwen} & \ac{lb} & Pso & No & Yes & No & $8.033 \pm 0.610$ & $15.600 \pm 2.542$ & $14.071$ & $1.486 \pm 0.264$ & $6.867 \pm 0.091$ & $2.450 \pm 0.097$ \\
\ac{qwen} & \ac{lb} & Pso & No & Yes & Yes & $8.363 \pm 0.578$ & $16.600 \pm 2.676$ & $13.075$ & $1.356 \pm 0.273$ & $7.283 \pm 0.100$ & $2.758 \pm 0.113$ \\
\ac{qwen} & \ac{lb} & Pso & Yes & No & No & $8.094 \pm 0.330$ & $18.800 \pm 0.583$ & $12.631$ & $1.207 \pm 0.150$ & $6.840 \pm 0.166$ & $2.200 \pm 0.659$ \\
\ac{qwen} & \ac{lb} & Pso & Yes & No & Yes & $8.274 \pm 0.612$ & $16.000 \pm 3.017$ & $12.625$ & $1.068 \pm 0.160$ & $6.880 \pm 0.137$ & $2.080 \pm 0.621$ \\
\ac{qwen} & \ac{lb} & Pso & Yes & Yes & No & $7.755 \pm 0.674$ & $15.400 \pm 2.502$ & $16.945$ & $1.091 \pm 0.120$ & $6.671 \pm 0.141$ & $2.000 \pm 0.630$ \\
\ac{qwen} & \ac{lb} & Pso & Yes & Yes & Yes & $8.214 \pm 0.595$ & $15.000 \pm 2.470$ & $9.898$ & $1.408 \pm 0.094$ & $6.936 \pm 0.136$ & $1.921 \pm 0.625$ \\
\ac{qwen} & Circle & Grid & No & No & No & $1.421 \pm 0.446$ & $16.800 \pm 1.772$ & $66.319$ & $0.841 \pm 0.039$ & $6.960 \pm 0.125$ & $2.090 \pm 0.111$ \\
\ac{qwen} & Circle & Grid & No & No & Yes & $2.165 \pm 1.184$ & $13.000 \pm 2.811$ & $67.059$ & $0.570 \pm 0.074$ & $7.320 \pm 0.151$ & $2.330 \pm 0.145$ \\
\ac{qwen} & Circle & Grid & No & Yes & No & $5.729 \pm 3.229$ & $11.600 \pm 2.482$ & $49.361$ & $1.124 \pm 0.314$ & $6.993 \pm 0.172$ & $2.158 \pm 0.142$ \\
\ac{qwen} & Circle & Grid & No & Yes & Yes & $2.801 \pm 1.048$ & $16.800 \pm 1.655$ & $64.558$ & $1.140 \pm 0.210$ & $7.514 \pm 0.143$ & $2.507 \pm 0.110$ \\
\ac{qwen} & Circle & Grid & Yes & No & No & $2.055 \pm 0.540$ & $16.400 \pm 1.965$ & $71.924$ & $0.704 \pm 0.060$ & $6.520 \pm 0.160$ & $1.880 \pm 0.622$ \\
\ac{qwen} & Circle & Grid & Yes & No & Yes & $3.016 \pm 1.513$ & $17.000 \pm 1.342$ & $65.665$ & $0.683 \pm 0.089$ & $8.298 \pm 0.224$ & $2.846 \pm 0.711$ \\
\ac{qwen} & Circle & Grid & Yes & Yes & No & $1.569 \pm 0.545$ & $15.600 \pm 1.939$ & $80.521$ & $0.816 \pm 0.115$ & $6.957 \pm 0.129$ & $1.714 \pm 0.607$ \\
\ac{qwen} & Circle & Grid & Yes & Yes & Yes & $1.426 \pm 0.320$ & $12.000 \pm 2.273$ & $80.916$ & $0.803 \pm 0.082$ & $6.567 \pm 0.148$ & $1.825 \pm 0.628$ \\
\ac{qwen} & Circle & Pso & No & No & No & $2.796 \pm 1.260$ & $9.200 \pm 2.853$ & $72.914$ & $1.072 \pm 0.152$ & $7.279 \pm 0.119$ & $2.558 \pm 0.142$ \\
\ac{qwen} & Circle & Pso & No & No & Yes & $5.096 \pm 2.619$ & $11.600 \pm 2.064$ & $54.169$ & $1.544 \pm 0.345$ & $7.553 \pm 0.102$ & $2.965 \pm 0.115$ \\
\ac{qwen} & Circle & Pso & No & Yes & No & $3.511 \pm 1.910$ & $16.800 \pm 0.970$ & $72.015$ & $1.117 \pm 0.158$ & $6.886 \pm 0.129$ & $2.521 \pm 0.103$ \\
\ac{qwen} & Circle & Pso & No & Yes & Yes & $8.548 \pm 3.112$ & $13.600 \pm 2.839$ & $19.337$ & $1.288 \pm 0.152$ & $7.267 \pm 0.091$ & $2.475 \pm 0.124$ \\
\ac{qwen} & Circle & Pso & Yes & No & No & $4.499 \pm 2.935$ & $15.400 \pm 2.015$ & $71.976$ & $0.658 \pm 0.087$ & $6.380 \pm 0.120$ & $1.800 \pm 0.606$ \\
\ac{qwen} & Circle & Pso & Yes & No & Yes & $4.987 \pm 2.249$ & $17.200 \pm 1.497$ & $63.842$ & $0.932 \pm 0.164$ & $6.340 \pm 0.142$ & $1.770 \pm 0.636$ \\
\ac{qwen} & Circle & Pso & Yes & Yes & No & $5.251 \pm 2.491$ & $13.000 \pm 2.720$ & $63.196$ & $1.098 \pm 0.183$ & $6.443 \pm 0.120$ & $1.679 \pm 0.607$ \\
\ac{qwen} & Circle & Pso & Yes & Yes & Yes & $4.308 \pm 1.784$ & $14.400 \pm 2.400$ & $62.833$ & $1.337 \pm 0.164$ & $6.414 \pm 0.140$ & $1.771 \pm 0.647$ \\
\ac{qwen} & Ellipse & Grid & No & No & No & $2.912 \pm 0.692$ & $8.400 \pm 2.522$ & $40.954$ & $0.706 \pm 0.096$ & $6.620 \pm 0.113$ & $1.480 \pm 0.107$ \\
\ac{qwen} & Ellipse & Grid & No & No & Yes & $2.470 \pm 0.285$ & $7.200 \pm 1.241$ & $36.856$ & $0.724 \pm 0.040$ & $7.280 \pm 0.144$ & $2.000 \pm 0.129$ \\
\ac{qwen} & Ellipse & Grid & No & Yes & No & $2.504 \pm 0.402$ & $9.600 \pm 2.694$ & $41.216$ & $0.906 \pm 0.041$ & $5.950 \pm 0.156$ & $1.575 \pm 0.110$ \\
\ac{qwen} & Ellipse & Grid & No & Yes & Yes & $2.434 \pm 0.327$ & $12.800 \pm 2.498$ & $36.607$ & $0.710 \pm 0.099$ & $7.206 \pm 0.129$ & $2.221 \pm 0.120$ \\
\ac{qwen} & Ellipse & Grid & Yes & No & No & $2.861 \pm 0.573$ & $6.000 \pm 0.894$ & $25.059$ & $0.719 \pm 0.151$ & $7.221 \pm 0.144$ & $1.992 \pm 0.634$ \\
\ac{qwen} & Ellipse & Grid & Yes & No & Yes & $2.824 \pm 0.531$ & $12.400 \pm 3.187$ & $42.206$ & $0.630 \pm 0.087$ & $7.200 \pm 0.178$ & $1.690 \pm 0.630$ \\
\ac{qwen} & Ellipse & Grid & Yes & Yes & No & $2.518 \pm 0.384$ & $8.600 \pm 2.462$ & $36.174$ & $0.989 \pm 0.161$ & $7.214 \pm 0.128$ & $1.829 \pm 0.617$ \\
\ac{qwen} & Ellipse & Grid & Yes & Yes & Yes & $2.512 \pm 0.418$ & $12.600 \pm 2.272$ & $31.074$ & $1.010 \pm 0.223$ & $7.744 \pm 0.169$ & $2.389 \pm 0.652$ \\
\ac{qwen} & Ellipse & Pso & No & No & No & $4.383 \pm 0.769$ & $12.400 \pm 3.311$ & $35.405$ & $1.070 \pm 0.128$ & $6.820 \pm 0.140$ & $2.090 \pm 0.117$ \\
\ac{qwen} & Ellipse & Pso & No & No & Yes & $4.664 \pm 0.770$ & $10.000 \pm 2.828$ & $23.732$ & $0.941 \pm 0.068$ & $6.100 \pm 0.077$ & $1.560 \pm 0.095$ \\
\ac{qwen} & Ellipse & Pso & No & Yes & No & $4.338 \pm 0.858$ & $14.000 \pm 1.975$ & $31.468$ & $1.252 \pm 0.167$ & $6.700 \pm 0.099$ & $1.900 \pm 0.113$ \\
\ac{qwen} & Ellipse & Pso & No & Yes & Yes & $4.476 \pm 0.851$ & $14.400 \pm 3.265$ & $27.073$ & $1.348 \pm 0.172$ & $6.683 \pm 0.115$ & $2.183 \pm 0.115$ \\
\ac{qwen} & Ellipse & Pso & Yes & No & No & $4.560 \pm 0.605$ & $14.600 \pm 2.768$ & $23.837$ & $0.808 \pm 0.123$ & $6.880 \pm 0.189$ & $2.100 \pm 0.647$ \\
\ac{qwen} & Ellipse & Pso & Yes & No & Yes & $5.038 \pm 0.948$ & $17.400 \pm 0.872$ & $21.189$ & $1.178 \pm 0.158$ & $6.280 \pm 0.114$ & $1.440 \pm 0.619$ \\
\ac{qwen} & Ellipse & Pso & Yes & Yes & No & $4.032 \pm 0.575$ & $14.600 \pm 2.040$ & $37.859$ & $1.185 \pm 0.210$ & $6.233 \pm 0.159$ & $1.758 \pm 0.619$ \\
\ac{qwen} & Ellipse & Pso & Yes & Yes & Yes & $4.787 \pm 0.687$ & $11.400 \pm 3.156$ & $23.065$ & $1.090 \pm 0.116$ & $7.036 \pm 0.114$ & $2.079 \pm 0.624$ \\
\hline
\end{tabular}}
\caption{Full results for \ac{qwen} on \acl{lb}, circle, and ellipse targets. Column definitions follow Table~\ref{tab:full_llama_a}. Each row aggregates 5 independent runs.}
\label{tab:full_qwen_a}
\end{table*}

\begin{table*}
\centering
\resizebox{\columnwidth}{!}{\begin{tabular}{@{}lllllllllllll@{}}
  \hline
  Model & Shape & Opt & Planner & \acs{dr} & \acs{cl} & Best chamf. & Steps & \% Imp. & \% Semantic & Links & Goal links \\
  \hline
\ac{qwen} & Line & Grid & No & No & No & $0.839 \pm 0.439$ & $13.000 \pm 2.665$ & $78.334$ & $0.688 \pm 0.065$ & $6.580 \pm 0.162$ & $1.640 \pm 0.117$ \\
\ac{qwen} & Line & Grid & No & No & Yes & $1.487 \pm 1.214$ & $14.200 \pm 2.596$ & $80.754$ & $0.713 \pm 0.082$ & $7.010 \pm 0.139$ & $2.283 \pm 0.128$ \\
\ac{qwen} & Line & Grid & No & Yes & No & $0.678 \pm 0.311$ & $15.200 \pm 1.934$ & $89.407$ & $0.975 \pm 0.124$ & $6.702 \pm 0.179$ & $2.092 \pm 0.132$ \\
\ac{qwen} & Line & Grid & No & Yes & Yes & $0.675 \pm 0.393$ & $11.000 \pm 2.214$ & $89.604$ & $0.838 \pm 0.093$ & $6.475 \pm 0.138$ & $1.795 \pm 0.116$ \\
\ac{qwen} & Line & Grid & Yes & No & No & $0.697 \pm 0.280$ & $15.200 \pm 2.922$ & $83.392$ & $0.719 \pm 0.022$ & $7.200 \pm 0.216$ & $2.040 \pm 0.691$ \\
\ac{qwen} & Line & Grid & Yes & No & Yes & $1.821 \pm 1.136$ & $12.600 \pm 1.860$ & $74.093$ & $0.655 \pm 0.057$ & $7.220 \pm 0.188$ & $1.960 \pm 0.637$ \\
\ac{qwen} & Line & Grid & Yes & Yes & No & $0.934 \pm 0.395$ & $11.800 \pm 2.177$ & $79.404$ & $0.860 \pm 0.105$ & $6.924 \pm 0.366$ & $2.348 \pm 0.828$ \\
\ac{qwen} & Line & Grid & Yes & Yes & Yes & $1.761 \pm 0.816$ & $13.000 \pm 2.569$ & $69.311$ & $0.790 \pm 0.286$ & $6.818 \pm 0.179$ & $1.802 \pm 0.640$ \\
\ac{qwen} & Line & Pso & No & No & No & $3.026 \pm 1.214$ & $8.600 \pm 2.159$ & $72.877$ & $0.631 \pm 0.088$ & $5.760 \pm 0.111$ & $1.440 \pm 0.110$ \\
\ac{qwen} & Line & Pso & No & No & Yes & $2.921 \pm 1.958$ & $16.400 \pm 2.462$ & $76.318$ & $0.870 \pm 0.032$ & $6.140 \pm 0.111$ & $1.730 \pm 0.100$ \\
\ac{qwen} & Line & Pso & No & Yes & No & $1.184 \pm 0.642$ & $16.000 \pm 2.915$ & $87.660$ & $1.288 \pm 0.225$ & $6.057 \pm 0.127$ & $1.688 \pm 0.113$ \\
\ac{qwen} & Line & Pso & No & Yes & Yes & $4.083 \pm 1.935$ & $13.600 \pm 1.913$ & $62.652$ & $0.872 \pm 0.149$ & $5.513 \pm 0.137$ & $1.675 \pm 0.111$ \\
\ac{qwen} & Line & Pso & Yes & No & No & $3.076 \pm 1.724$ & $17.600 \pm 1.166$ & $70.736$ & $0.806 \pm 0.020$ & $6.770 \pm 0.148$ & $2.340 \pm 0.627$ \\
\ac{qwen} & Line & Pso & Yes & No & Yes & $2.091 \pm 1.315$ & $13.400 \pm 1.288$ & $75.320$ & $0.899 \pm 0.077$ & $6.915 \pm 0.168$ & $1.899 \pm 0.652$ \\
\ac{qwen} & Line & Pso & Yes & Yes & No & $1.447 \pm 0.732$ & $10.800 \pm 2.990$ & $86.191$ & $0.820 \pm 0.225$ & $6.599 \pm 0.143$ & $1.833 \pm 0.617$ \\
\ac{qwen} & Line & Pso & Yes & Yes & Yes & $2.072 \pm 1.187$ & $17.800 \pm 1.020$ & $78.845$ & $1.194 \pm 0.449$ & $5.935 \pm 0.138$ & $1.367 \pm 0.607$ \\
\ac{qwen} & Naca & Grid & No & No & No & $0.366 \pm 0.099$ & $15.000 \pm 1.871$ & $49.809$ & $0.892 \pm 0.075$ & $6.920 \pm 0.162$ & $2.310 \pm 0.130$ \\
\ac{qwen} & Naca & Grid & No & No & Yes & $0.566 \pm 0.144$ & $11.000 \pm 2.881$ & $21.614$ & $0.710 \pm 0.113$ & $8.182 \pm 0.235$ & $2.852 \pm 0.217$ \\
\ac{qwen} & Naca & Grid & No & Yes & No & $0.484 \pm 0.115$ & $8.600 \pm 2.839$ & $32.050$ & $1.058 \pm 0.103$ & $6.357 \pm 0.110$ & $1.693 \pm 0.109$ \\
\ac{qwen} & Naca & Grid & No & Yes & Yes & $0.487 \pm 0.132$ & $12.800 \pm 3.839$ & $49.797$ & $0.813 \pm 0.161$ & $7.737 \pm 0.158$ & $2.489 \pm 0.156$ \\
\ac{qwen} & Naca & Grid & Yes & No & No & $0.400 \pm 0.124$ & $12.000 \pm 3.507$ & $52.115$ & $1.003 \pm 0.172$ & $7.480 \pm 0.210$ & $2.300 \pm 0.687$ \\
\ac{qwen} & Naca & Grid & Yes & No & Yes & $0.404 \pm 0.121$ & $9.800 \pm 2.223$ & $48.271$ & $0.673 \pm 0.052$ & $7.360 \pm 0.199$ & $2.580 \pm 0.663$ \\
\ac{qwen} & Naca & Grid & Yes & Yes & No & $0.432 \pm 0.129$ & $8.400 \pm 2.676$ & $49.648$ & $0.772 \pm 0.094$ & $7.417 \pm 0.196$ & $1.875 \pm 0.624$ \\
\ac{qwen} & Naca & Grid & Yes & Yes & Yes & $0.473 \pm 0.181$ & $16.200 \pm 1.114$ & $47.869$ & $0.985 \pm 0.143$ & $7.013 \pm 0.143$ & $2.081 \pm 0.615$ \\
\ac{qwen} & Naca & Pso & No & No & No & $0.664 \pm 0.279$ & $16.200 \pm 1.158$ & $32.861$ & $1.030 \pm 0.069$ & $6.940 \pm 0.196$ & $2.050 \pm 0.158$ \\
\ac{qwen} & Naca & Pso & No & No & Yes & $0.563 \pm 0.202$ & $11.000 \pm 2.280$ & $37.899$ & $1.042 \pm 0.133$ & $6.820 \pm 0.145$ & $2.280 \pm 0.150$ \\
\ac{qwen} & Naca & Pso & No & Yes & No & $0.556 \pm 0.214$ & $8.600 \pm 2.804$ & $34.942$ & $1.274 \pm 0.299$ & $6.217 \pm 0.100$ & $1.833 \pm 0.107$ \\
\ac{qwen} & Naca & Pso & No & Yes & Yes & $0.722 \pm 0.342$ & $12.000 \pm 2.324$ & $32.324$ & $1.227 \pm 0.185$ & $5.900 \pm 0.140$ & $1.450 \pm 0.127$ \\
\ac{qwen} & Naca & Pso & Yes & No & No & $0.417 \pm 0.173$ & $16.400 \pm 2.227$ & $49.975$ & $1.175 \pm 0.099$ & $6.800 \pm 0.142$ & $2.250 \pm 0.640$ \\
\ac{qwen} & Naca & Pso & Yes & No & Yes & $0.571 \pm 0.224$ & $16.600 \pm 1.631$ & $52.741$ & $1.285 \pm 0.210$ & $6.960 \pm 0.174$ & $2.250 \pm 0.658$ \\
\ac{qwen} & Naca & Pso & Yes & Yes & No & $0.729 \pm 0.332$ & $12.800 \pm 2.888$ & $47.248$ & $1.308 \pm 0.298$ & $6.380 \pm 0.175$ & $1.810 \pm 0.626$ \\
\ac{qwen} & Naca & Pso & Yes & Yes & Yes & $0.394 \pm 0.123$ & $12.200 \pm 3.121$ & $44.510$ & $1.221 \pm 0.177$ & $6.333 \pm 0.171$ & $1.817 \pm 0.636$ \\
\ac{qwen} & Parabola & Grid & No & No & No & $493.251 \pm 99.687$ & $14.200 \pm 2.245$ & $44.790$ & $0.941 \pm 0.146$ & $7.140 \pm 0.176$ & $2.420 \pm 0.145$ \\
\ac{qwen} & Parabola & Grid & No & No & Yes & $735.749 \pm 43.624$ & $16.200 \pm 1.530$ & $17.438$ & $0.760 \pm 0.117$ & $7.680 \pm 0.147$ & $2.970 \pm 0.152$ \\
\ac{qwen} & Parabola & Grid & No & Yes & No & $623.226 \pm 107.714$ & $17.800 \pm 1.715$ & $30.770$ & $0.893 \pm 0.125$ & $6.850 \pm 0.096$ & $2.475 \pm 0.092$ \\
\ac{qwen} & Parabola & Grid & No & Yes & Yes & $545.550 \pm 159.913$ & $14.600 \pm 1.470$ & $38.662$ & $0.890 \pm 0.082$ & $6.556 \pm 0.101$ & $1.894 \pm 0.091$ \\
\ac{qwen} & Parabola & Grid & Yes & No & No & $534.193 \pm 108.984$ & $16.600 \pm 1.887$ & $40.330$ & $0.745 \pm 0.063$ & $6.520 \pm 0.174$ & $2.250 \pm 0.657$ \\
\ac{qwen} & Parabola & Grid & Yes & No & Yes & $692.076 \pm 53.259$ & $15.000 \pm 1.483$ & $22.021$ & $0.609 \pm 0.098$ & $6.900 \pm 0.210$ & $2.270 \pm 0.658$ \\
\ac{qwen} & Parabola & Grid & Yes & Yes & No & $621.629 \pm 81.346$ & $17.400 \pm 0.927$ & $30.160$ & $0.851 \pm 0.137$ & $6.744 \pm 0.130$ & $1.794 \pm 0.600$ \\
\ac{qwen} & Parabola & Grid & Yes & Yes & Yes & $814.554 \pm 50.081$ & $14.800 \pm 1.562$ & $8.473$ & $0.700 \pm 0.144$ & $6.767 \pm 0.109$ & $1.856 \pm 0.605$ \\
\ac{qwen} & Parabola & Pso & No & No & No & $846.103 \pm 49.383$ & $15.800 \pm 2.596$ & $5.560$ & $0.918 \pm 0.100$ & $6.380 \pm 0.132$ & $1.890 \pm 0.113$ \\
\ac{qwen} & Parabola & Pso & No & No & Yes & $831.704 \pm 49.272$ & $12.800 \pm 2.746$ & $7.266$ & $1.060 \pm 0.168$ & $6.740 \pm 0.109$ & $2.200 \pm 0.126$ \\
\ac{qwen} & Parabola & Pso & No & Yes & No & $842.036 \pm 47.694$ & $15.400 \pm 1.631$ & $6.046$ & $0.888 \pm 0.227$ & $6.271 \pm 0.111$ & $1.393 \pm 0.079$ \\
\ac{qwen} & Parabola & Pso & No & Yes & Yes & $893.487 \pm 24.173$ & $15.400 \pm 1.965$ & $0.283$ & $1.146 \pm 0.119$ & $7.250 \pm 0.146$ & $2.288 \pm 0.112$ \\
\ac{qwen} & Parabola & Pso & Yes & No & No & $861.781 \pm 41.365$ & $11.000 \pm 3.592$ & $3.822$ & $0.892 \pm 0.144$ & $6.530 \pm 0.219$ & $1.860 \pm 0.665$ \\
\ac{qwen} & Parabola & Pso & Yes & No & Yes & $858.141 \pm 38.651$ & $14.000 \pm 3.464$ & $4.037$ & $0.864 \pm 0.283$ & $7.300 \pm 0.185$ & $2.190 \pm 0.674$ \\
\ac{qwen} & Parabola & Pso & Yes & Yes & No & $884.992 \pm 23.229$ & $16.600 \pm 2.227$ & $1.152$ & $1.019 \pm 0.120$ & $5.929 \pm 0.126$ & $1.514 \pm 0.606$ \\
\ac{qwen} & Parabola & Pso & Yes & Yes & Yes & $887.336 \pm 21.585$ & $12.600 \pm 2.943$ & $0.954$ & $0.938 \pm 0.137$ & $6.317 \pm 0.105$ & $1.425 \pm 0.601$ \\
  \hline
\end{tabular}}
\caption{Full results for \ac{qwen} on straight-line, NACA airfoil, and parabola targets. Column definitions follow Table~\ref{tab:full_llama_a}. Each row aggregates 5 independent runs.}
\label{tab:full_qwen_b}
\end{table*}

\subsection{High-budget Enum+GA vs. Symbolic (60$\times$300)}
\label{app:high_budget_enumga_vs_symbolic}
{\small
\begin{table}[!htp]
\centering
\begin{tabular}{ll l r r r}
\toprule
\multicolumn{6}{l}{\textit{Enum+GA vs.\ Symbolic at High Budget (60\texttimes300)}} \\
\cmidrule(r){1-6}
\textbf{Pop.\ \texttimes\ Gen.} & \textbf{Bars} & \textbf{Model} & \textbf{x/Base} & \textbf{$\Delta$ (x$-$Base)} & \textbf{Imp.\%} \\
\midrule
60\texttimes300 & 4 & \ac{llama}    & $1.130 \pm 0.086$ & \textcolor{red}{${\uparrow}0.130$} & \textcolor{red}{${\downarrow}13.0\%$} \\
60\texttimes300 & 4 & \ac{qwen}     & $1.189 \pm 0.092$ & \textcolor{red}{${\uparrow}0.189$} & \textcolor{red}{${\downarrow}18.9\%$} \\
60\texttimes300 & 4 & \ac{qwen-moe} & $1.522 \pm 0.184$ & \textcolor{red}{${\uparrow}0.522$} & \textcolor{red}{${\downarrow}52.2\%$} \\
60\texttimes300 & 6 & \ac{llama}    & $1.175 \pm 0.153$ & \textcolor{red}{${\uparrow}0.175$} & \textcolor{red}{${\downarrow}17.5\%$} \\
60\texttimes300 & 6 & \ac{qwen}     & $1.219 \pm 0.140$ & \textcolor{red}{${\uparrow}0.219$} & \textcolor{red}{${\downarrow}21.9\%$} \\
60\texttimes300 & 6 & \ac{qwen-moe} & $1.519 \pm 0.157$ & \textcolor{red}{${\uparrow}0.519$} & \textcolor{red}{${\downarrow}51.9\%$} \\
\bottomrule
\end{tabular}
\caption{High-budget subset (60$\times$300). Values are normalized index (x/Base), with mean $\pm$ standard error aggregated across shared target shapes.}
\label{tab:combined_all_high_budget_app}
\end{table}}

\end{document}